\documentclass[reqno]{amsart}
\usepackage{graphicx}
\usepackage{amssymb}
\usepackage{subcaption}
\usepackage{float}
\usepackage[]{algorithm2e}

\usepackage{algorithmic}
\usepackage{hyperref}
\hypersetup{
colorlinks=true,
linkcolor=red,
filecolor=magenta, 
urlcolor=cyan,
} 
\usepackage{color}
\usepackage{bm}

\newtheorem{theorem}{Theorem}[section]

\theoremstyle{definition}

\newtheorem{example}[theorem]{Example}

\theoremstyle{remark}

\DeclareMathOperator*{\argmin}{arg\,min}
\numberwithin{equation}{section}

\newcommand{\lprod}[2]{\Big(#1,#2\Big)_2}

\newcommand{\mbb}[1]{\mathbb{#1}}
\newcommand{\mcal}[1]{\mathcal{#1}}

\title{Instabilities in Plug-and-Play (PnP) algorithms from a learned denoiser}
\author{Abinash Nayak}
\usepackage{fancyhdr}
\begin{document}


\address{Visiting Assistant Professor, Department of Mathematics, University of Alabama at Birmingham, University Hall, Room 4005, 1402 10th Avenue South, Birmingham AL 35294-1241, (p) 205.934.2154, (f) 205.934.9025}
\email{nash101@uab.edu; avinashnike01@gmail.com}


\date{\today}


\begin{abstract}
It's well-known that inverse problems are ill-posed and to solve them meaningfully, one has to employ regularization methods. Traditionally, popular regularization methods are the penalized Variational approaches. In recent years, the classical regularization approaches have been outclassed by the so-called plug-and-play (PnP) algorithms, which copy the proximal gradient minimization processes, such as ADMM or FISTA, but with any general denoiser. However, unlike the traditional proximal gradient methods, the theoretical underpinnings, convergence, and stability results have been insufficient for these PnP-algorithms. Hence, the results obtained from these algorithms, though empirically outstanding, can't always be completely trusted, as they may contain certain instabilities or (hallucinated) features arising from the denoiser, especially when using a pre-trained learned denoiser. In fact, in this paper, we show that a PnP-algorithm can induce hallucinated features, when using a pre-trained deep-learning-based (DnCNN) denoiser. We show that such instabilities are quite different than the instabilities inherent to an ill-posed problem. We also present methods to subdue these instabilities and significantly improve the recoveries. We compare the advantages and disadvantages of a learned denoiser over a classical denoiser (here, BM3D), as well as, the effectiveness of the FISTA-PnP algorithm vs. the ADMM-PnP algorithm. In addition, we also provide an algorithm to combine these two denoisers, the learned and the classical, in a weighted fashion to produce even better results. We conclude with numerical results which validate the developed theories. 
\end{abstract}
\maketitle

\section{\textbf{Introduction}}
\subsection{Inverse Problems and Regularization:}
Mathematically, an inverse problem is often expressed as the problem of estimating a (\textit{source}) $\hat{x}$ which satisfies, for a given (\textit{effect}) $b$, the following matrix (or operator) equation
\begin{equation}\label{Inv. prob.}
A\hat{x} = b,
\end{equation}
where the matrix $A \in \mbb{R}^{n\times m}$ and the vectors $\hat{x} \in \mbb{R}^{n}$, $b \in \mbb{R}^{m}$ are the discrete approximations of an infinite dimensional model, describing the underlying physical phenomenon. The inverse problem \eqref{Inv. prob.} is usually ill-posed, in the sense of violating any one of the Hadamard's conditions for well-posedness: (i) Existence of a solution (for $b$ not in the range of $A$), (ii) uniqueness of the solution (for non-trivial null-space of $A$) and, (iii) continuous dependence on the data (for ill-conditioned A). Conditions (i) and (ii) can be circumvented by relaxing the definition of a solution for \eqref{Inv. prob.}, for example, finding the least square solution or the minimal norm solution (i.e., the pseudo-inverse solution $x^\dagger$). The most (practically) significant condition is condition-(iii), since failing of this leads to an absurd (unstable) solution. That is, for an (injective) A and an exact $b$ (noiseless), the solution of \eqref{Inv. prob.} can be approximated by the (LS) least-square solution ($x^\dagger$), i.e., $x^\dagger$ is the minimizer of the following least-square functional
\begin{equation}\label{LS functional}
    F(x) = ||Ax - b||_2^2,
\end{equation}
where as, for a noisy data $b_\delta$ (which is practically true, most of the time) such that $||b - b_\delta|| \leq \delta$, the simple least-square solution $x^\dagger_\delta$, with respect to $b_\delta$ in \eqref{LS functional}, fails to approximate the true solution, i.e., $||x^\dagger_\delta - x^\dagger|| >> \delta$, which in turn implies, $||x^\dagger_\delta - \hat{x}|| >> \delta$, due to the ill-posedness of the inverse problem \eqref{Inv. prob.}. To counter such instabilities or ill-posedness of inverse problems, regularization methods have to be employed.
\subsection{Variational (or penalized) regularization and Related works}
\mbox{ }

Such approaches, also known as Tikhonov-type regularization, are probably the most well known regularization techniques for solving linear, as well as nonlinear, inverse problems (see \cite{Engl+Hanke+Neubauer,Bakushinsky+Goncharsky, Groetsch, Baumeister,Morozov_a}), where, instead of minimizing the simple least-square functional \eqref{LS functional}, one minimizes a penalized (or constrained) functional:
\begin{equation}\label{Gen. Tik. fun.}
F(x;\mcal{D},\lambda,\mcal{R}) =  \mcal{D}(Ax,b_\delta) + \lambda \mcal{R}(x),
\end{equation}
where $\mcal{D}$ is called the data-fidelity term (imposing data-consistency), $\mcal{R}$ is the regularization term (imposing certain structures, based on some prior knowledge of the solution $\hat{x}$) and, $\lambda \geq 0$ is the regularization parameter that balances the trade-off between them, depending on the noise level $\delta$, i.e., $\lambda=\lambda(\delta)$. The formulation \eqref{Gen. Tik. fun.} also has a Bayesian interpretation, where the minimization of $F(x;\mcal{D},\lambda,\mcal{R})$ corresponds to the maximum-a-posteriori (MAP) estimate of $\hat{x}$ given $b_\delta$, where the likelihood of $b_\delta$ is proportional to $exp(-\mcal{D}(x))$ and the prior distribution on $\hat{x}$ is proportional to $exp(-\mcal{R}(x))$. Classically, $\mcal{D}(Ax,b_\delta) = ||{Ax - b_\delta}||_p^p$ and $\mcal{R}(x) = ||Lx - \bar{x}||_q^q$, where $L$ is a regularization matrix, with the null spaces of $A$ and $L$ intersecting trivially, and $p$, $q$ determine the involved norms. For large scale problems, the minimization of \eqref{Gen. Tik. fun.} is done iteratively, and for convex, differentiable functions $\mcal{D}$ and $\mcal{R}$, one can minimize \eqref{Gen. Tik. fun.} either via the simple steepest descent method or via faster Krylov subspace methods, such as Conjugate-Gradient method etc., see \cite{Hanke1991, Landweber, Hanke_Neubauer_Scherzer, Engl+Hanke+Neubauer}. Where as, for a non-differentiable $\mcal{R}$, which is proper, closed and convex, the non-differentiability issue can be circumvented by using a proximal operator, see \cite{Beck_Teboulle, Boyd_Parikh_Chu_Peleato_Eckstein, Chambolle_Pock} and references therein, which is defined as
\begin{equation}
    \mbox{Prox}_{\lambda \mcal{R}}(v) = \argmin_x \; \lambda \mcal{R}(x) + \frac{1}{2} ||x - v||_2^2.
\end{equation}
Basically, for smooth $\mcal{D}(x)$ and non-smooth $\mcal{R}(x)$, the minimization problem corresponding to \eqref{Gen. Tik. fun.} can be solved via two first-order iterative methods:
\begin{enumerate}
    \item Forward-backward splitting (FBS), also known as Iterative shrinkage/soft thresholding algorithm (ISTA) and has a faster variant Fast ISTA (FISTA), where each minimization step is divided into two sub-steps, given by
    \begin{align}\label{FBS min.}
        z_{k+1}^\delta &= x_k^\delta - \tau_k \nabla_x \mcal{D}(x_k^\delta) \;\; \longleftarrow \text{ data-consistency step}\\
        x_{k+1}^\delta &= \text{Prox}_{\lambda \tau_k \mcal{R}}(z_{k+1}^\delta) \;\; \longleftarrow \text{ data-denoising step}
    \end{align}    
    where $\tau_k \geq 0$ is the step-size at the k$^{th}$ iteration.

    \item Alternating direction method of multipliers (ADMM), where three sequences are alternatively updated as follows, 
    \begin{align}\label{ADMM min.}
        x_{k+1}^\delta &= \text{Prox}_{\frac{1}{\rho}\mcal{D}}(z_k^\delta - u_k^\delta) \;\; \longleftarrow \text{ data-consistency step}\\
        z_{k+1}^\delta &= \text{Prox}_{\frac{\lambda}{\rho}\mcal{R}}(x_{k+1}^\delta + u_{k}^\delta) \;\; \longleftarrow \text{ data-denoising step}\\
        u_{k+1}^\delta &= u_k^\delta + x_{k+1}^\delta - z_{k+1}^\delta, \;\; \longleftarrow \text{ noise update step}
    \end{align}
    where $\rho > 0$ is the Lagrangian parameter, which only effects the speed of convergence and not the solution (minimizer) of \eqref{Gen. Tik. fun.}.
\end{enumerate}
From the above two expressions, one can observe that, each method comprises of two fundamental steps: (1) data-consistency, and (2) data-denoising. This motivated, authors in \cite{Venkatakrishnan_Bouman_Wohlberg}, to replace the Prox$_{\hat{\sigma} \mcal{R}}$ operator in the denoising step of ADMM by an off-the-shelf denoiser $H_{\sigma}$, which is tuned to $H_{\hat{\sigma}=\sigma/\rho}$, where $\sigma$ is the denoising strength of the original denoiser $H_\sigma$, and termed the process as the PnP-algorithm (plug-and-play method). However, note that, once the proximal operator is replaced by any general denoiser then the Variational problem \eqref{Gen. Tik. fun.} breaks down, as not all denoisers can be expressed as a proximal operator of some function $\mcal{R}$. Hence, all the theories and results related to the classical  Variational regularization methods also break down, such as the convergence, regularization and stability analysis, and even, the meaning of the solution, i.e., how to define the obtained solution? is it a minimzer of some functional? etc. Though empirical results show the convergence of these PnP-algorithms, there is no proof of it, for any general denoisers. However, under certain assumptions and restrictions (such as boundedness, nonexpansiveness, etc.) on the denoiser, there have been some convergence proof, see \cite{Chan_Wang_Elgendy, Buzzard_Chan_Sreehari_Bouman, Ryu_Liu_Wang_Chen_Zhangyang_Yin, Teodoro_Bioucas_Figueiredo, Sun_Wohlberg_Kamilov} and references therein. There are also some other variants of such PnP-methods, such as Regularization by Denoising (RED)\cite{Romano_Elad_Milanfar}, Regularization by Artifact-Removal (RARE)  \cite{Kamilov_Liu_Sun_Eldeniz_Gan_An}, etc.

\subsection*{Contribution of this paper}
\mbox{}

\begin{itemize}
\item In this paper, we present the instabilities arising in such PnP-algorithms, due to the lack of theoretical underpinnings, especially for an off-the-shelf non-calssical denoisers, such as, a deep-learning based denoiser.
\item We also present certain regularization methods to subdue the above mentioned instabilities, which leads to much better and stable recoveries.
\item We also compare the FBS-PnP algorithm with the ADMM-PnP algorithm and show the advantages/disadvantages of one over the other, i.e., which algorithm is more appropriate for a given denoiser. Note that, in the classical scenario, both these algorithms produce the same result, which is the minimizer of the functional defined in \eqref{Gen. Tik. fun.}. However, for PnP algorithms with general denoisers, they are not the same, i.e., the architecture of the iterative process does effect the recovered solution.
\item We also provide methods to combine these two denoisers, the classical and the learned, in a weighted manner, which take advantages of both these worlds and produce better results.
\item We conclude with numerical examples, validating the developed theories. 
\end{itemize}

\section{PnP-Algorithms as structured iterations}
In this section, we interpret PnP-algorithms from a different perspective. First, let's rewrite the PnP-versions, for any general denoiser $H_\sigma$, from their respective classical proximal gradient methods, i.e.,
\begin{enumerate}
\item \textbf{FBS-PnP (Forward-backward splitting - PnP):}
In this algorithm, for a fixed denoiser $H_\sigma$ (of denoising strength corresponding to noise level $\sigma$) and starting from an initial choice $z_0^\delta$, at any iteration step $k\geq 1$, we have 
\begin{align}
z_{k-1}^\delta &\longmapsto \; x_k^\delta \; = \; z_{k-1}^\delta - \tau_k \nabla_x \mcal{D}(z_{k-1}^\delta) \; &\longleftarrow \mbox{ data-consistency step} \label{FBS-PnP data-consistency} \\
x_k^\delta &\longmapsto \; z_k^\delta \; = \; H_{\sigma_k}(x_k^\delta), \; &\longleftarrow \mbox{ data-denoising step}, \label{FBS-PnP data-denoising}
\end{align}
where $H_{\sigma_k}$ is the updated kth denoiser, with the denoising strength corresponding to $\sigma_k = \tau_k \sigma$.

\item \textbf{ADMM-PnP (Alternating direction method of multipliers - PnP):}
Here, for a fixed denoiser $H_\sigma$ (of denoising strength corresponding to noise level $\sigma$) and starting from initial choices $x_0^\delta$, $z_0^\delta$ and $u_0^\delta$, at any iteration step $k\geq 1$, we have 
    \begin{align}
        x_{k+1}^\delta &= \text{Prox}_{\frac{1}{\rho}\mcal{D}}(z_k^\delta - u_k^\delta) \;\; \longleftarrow \text{ data-consistency step}\label{ADMM-PnP data-cons.} \\
        &= \argmin_x \;\; \mcal{D}(x) + \rho ||x - (z_k^\delta - u_k^\delta)||_2^2 \notag\\
        z_{k+1}^\delta &= H_{\sigma_k}(x_{k+1}^\delta + u_{k}^\delta) \;\; \longleftarrow \text{ data-denoising step} \label{ADMM-PnP data-deno.}\\
        u_{k+1}^\delta &= u_k^\delta + x_{k+1}^\delta - z_{k+1}^\delta \;\; \longleftarrow \text{ noise update step} 
        \label{ADMM-PnP noise upd.}
    \end{align}
where $H_{\sigma_k}$ is the kth updated denoiser, with the denoising strength corresponding to $\sigma_k = \frac{\sigma}{\rho}$.
\end{enumerate}
Note that, for the classical case ($H_{\sigma_k}$ corresponding to a closed, proper and convex regularizer $\mcal{R}$ in \eqref{Gen. Tik. fun.}), both the above algorithms should produce the same result, the minimizer of \eqref{Gen. Tik. fun.}, and the parameters values, $\tau_k$ and $\rho$, only effect the convergence of the algorithms and not the final solution, $x^\delta{(\mcal{D},\lambda,\mcal{R})}$. However, this might not be true for PnP algorithms, when using any general denoiser $H_\sigma$.

Also, note that, the resulting direction at $(k-1)^{th}$ step, in the FBS-PnP algorithm, is given by
\begin{align}\label{FBS-PnP direction}
d_{k-1}^\delta &:= z_k^\delta - z_{k-1}^\delta \\ 
&= \underbrace{-\tau_k \nabla_x \mcal{D}(z_{k-1}^\delta)}_{data-consistency} \; + \; \underbrace{\left(H_{\sigma_k}(x_k^\delta) - x_k^\delta\right)}_{data-denoising} \;, \notag
\end{align}
and the direction $d_{k-1}^\delta$, as defined in \eqref{FBS-PnP direction}, will be a descent direction provided it satisfies, for $\mcal{D}(x)=||Ax - b_\delta||_2^2$,
\begin{equation}\label{LS descend prop.}
	\lprod{d_k^\delta}{-\nabla_x \mcal{D}(x_k^\delta)} = \lprod{d_k^\delta}{-A^*(Ax_k^\delta - b_\delta)} > 0,
\end{equation}
where $\lprod{.}{.}$ is the associated $\ell_2$-product. This can be achieved for $H_{\sigma_k}$ satisfying
\begin{equation}\label{H(x) - x}
||H_{\sigma_k}(x_k^\delta) - x_k^\delta|| \; < \; ||-\tau_k \nabla_x \mcal{D}(x_{k-1}^\delta)||,
\end{equation}
since then
\begin{align}
\lprod{d_{k-1}^\delta}{-\tau_k \nabla_x \mcal{D}(z_{k-1}^\delta)} \geq \tau_k ||\nabla_x \mcal{D}(z_{k-1}^\delta)||\left( \tau_k ||\nabla_x \mcal{D}(z_{k-1}^\delta)|| - ||H_{\sigma_k}(x_k^\delta) - x_k^\delta|| \right) > 0. \notag
\end{align}
Therefor, for such descent directions $d_{k-1}^\delta$, the relative errors in the recovery process will follow a semi-convergent trail, and hence, one can recover a regularized solution (via early stopping) containing certain structures in it, which are imposed by the denoiser $H_{\sigma_k}$, for further details see \cite{nayak2021PnP}. In other words, for directions $d_k^\delta$ satisfying \eqref{H(x) - x}, we obtain a family of regularized solutions given by
\begin{align}\label{LS. family I_3}
\mcal{I}_3 := &\left\{ x^\delta(\mcal{D},k,d_k^\delta;\mcal{S}) \right. \;  : \; \left. x_{k}^\delta = x_{k-1}^\delta \;+\; d_{k-1}^\delta, \;\; 1\leq k \leq k(\delta,\mcal{S}) \leq  \infty, \right.\\
& \left. \; \mbox{s.t.}\; \lprod{d_{k-1}^\delta}{- \tau_{k-1} \nabla_x \mcal{D}(x_{k-1}^\delta)}>0 \mbox{ and $\mcal{S}$ is a selection criterion.}  \right\}. \notag
\end{align}
Note that, \eqref{H(x) - x} is only a sufficient condition for $d_k^\delta$ to be a descent direction, i.e., $d_k^\delta$ violating \eqref{H(x) - x} can also be a descent direction (satisfying \eqref{LS descend prop.}). In fact, $d_k^\delta$ need not even satisfy \eqref{LS descend prop.} for all values of $k$, i.e., $d_k^\delta$ doesn't need to be a descent direction for all $k\geq 1$, in which case, we obtain a family of regularized solutions given by

\begin{align}\label{LS. family I_5}
\mcal{I}_5 := &\left\{ x^\delta(\mcal{D},k,d_k^\delta,x_0^\delta;\mcal{S}) \right. : \left. x_{k}^\delta = x_{k-1}^\delta \;+\; d_{k-1}^\delta, \; \mbox{ starting from $x_0^\delta$}, \right. \notag\\
& \left. \; \mbox{s.t. $d_{k-1}^\delta$ satisfies \eqref{descent dir. gen.} and $\mcal{S}$ is a selection criterion.}  \right\},
\end{align}
where the condition \eqref{descent dir. gen.} is a generalization of \eqref{LS descend prop.}, given by,
\begin{gather}\label{descent dir. gen.}
\lprod{d_k^\delta}{-\nabla_x \mcal{D}(x_k^\delta)} > 0, \;\;\; \mbox{ for } \; \mcal{D}(x_k^\delta) > \epsilon_2(\delta) \; \mbox{ or } \; k \leq k(\delta), \mbox{ and}\\
\mcal{D}(x_k^\delta + d_k^\delta) \geq \epsilon_1(\delta), \;\;\;\ \mbox{ for }\; \mcal{D}(x_k^\delta) \leq \epsilon_2(\delta) \; \mbox{ or } \; k > k(\delta). \notag
\end{gather}
With the above formulation for the family of regularized solutions, it can be shown that the solution of an ADMM-PnP algorithm falls in the class $\mcal{I}_5$, see \cite{nayak2021PnP} for details.

Also, in the recovery process, the dynamics of the denoising is reflected in the denoising-to-consistency ratio, which is defined as follows
\begin{equation}\label{denoising ratio}
DC(k):= \frac{||H_{\sigma_k}(x_k^\delta) - x_k^\delta||}{||-\tau_k \nabla_x \mcal{D}(x_{k-1}^\delta)||} = \frac{||H_{\sigma_k}(x_k^\delta) - x_k^\delta||}{||x_k^\delta - x_{k-1}^\delta||}.
\end{equation}
That is, if the ratio is very small $DC(k) << 1$, then the extent of denoising is very small in comparison to the amount of improvement towards the noisy data, and hence, can be lead to a noisy  recovery. Where as, if the ratio is very large $DC(k) >> 1$, then the extent of denoising is also very large, relative to the improvement in the data-consistency step, and hence, can lead to an over-smoothed solution. 
However, this doesn't always means that for $DC(k) << 1$ or $DC(k) >> 1$, the recoveries will be too noisy or over-smoothed, respectively, since, if the noise levels in $b_\delta$ is low (i.e., $||x_k^\delta - x_{k-1}^\delta||$ can be large) and the noise in $x_k^\delta$ is small (i.e., $||H_{\sigma_k}(x_k^\delta) - x_k^\delta||$ can be small), then $DC(k) << 1$, but can still produce well-denoised iterate $z_k^\delta$; on the other hand, if $H_{\sigma_k}$ is an excellent denoiser (i.e., $||H_{\sigma_k}(x_k^\delta) - \hat{x}|| << 1$, where $\hat{x}$ is the true solution), then, even for high noise levels in $x_k^\delta$ and $b_\delta$, the ratio $DC(k) >> 1$ and one can still produce excellent recovery. 
Nevertheless, inspecting the ratio $DC(k)$ provides some insights regarding the denoising dynamics in the recovery process, and thus, can help to improve the recovery in certain cases, details in \S \ref{Sec. Numerical Results}.

\subsection{Classical Denoiser vs. Learned Denoiser:}
\mbox{ }

Note that, a classical denoiser $H_\sigma$ is dependent on a denoising parameter ($\sigma \geq 0$), which controls the denoising strength of the denoiser, i.e., larger $\sigma$ implies stronger denoising and vice-verse. Therefore, the parameter $\sigma$ needs to appropriately tuned, based on the noise level $\delta$, for an effective denoising, i.e., one cannot use $H_\sigma$, for a fixed $\sigma$, universally for any noise level $\delta$. In contrast, ``an ideally learned denoiser" $H_{\theta_0}$ can be used to denoise universally, where an ideally learned denoiser implies, $H_{\theta_0}$ has learned to denoise at an universal level, i.e., noises of all levels and distributions. Here, $\theta_0 \in \mbb{R}^d$, for some $d$ (usually, high dimension), denotes the pre-trained internal parameters of the learned denoiser, such as the weights of a neural network. Of course, an ideally learned denoiser only exists in a hypothetical setting. In a typical scenario, one has a set of learning examples (training data) and a denoising architecture $H_{\theta}$ is trained on that data set, i.e., the parameters ($\theta$) of the denoiser $H_\theta$ is optimized to $\theta_0$ such that $H_{\theta_0}$ yields the ``most effective denoising" on that data set, where the ``most effective denoising" depends on the performance measuring metric (loss function) and the optimization process. Then, one hopes that for examples outside the training data set (i.e., in the testing data set) $H_{\theta_0}$ will also perform effective denoising. 
Hence, one can see that, when using a parameter-dependent classical denoiser $H_\sigma$, for solving an inverse problem, the denoising parameter $\sigma$ can also serve as a regularization parameter, which can be tuned appropriately for different noise levels $\delta$. Where as, when using a pre-trained learned denoiser $H_{\theta_0}$, whose internal parameters have been already optimized, one hopes that the bag of training examples contains noises of different levels and distributions, suited for that inverse problem. Now, even with a set of proper training examples, it is shown in \cite{Antun_Renna_Poon_Adcock_Hansen} that recovery algorithms for inverse problems based on deep learning can be very unstable, as a result of adversarial attacks. Although, the structure of the recovery processes mentioned there is of slightly different flavor than that of a PnP-algorithm with (deep) learned denoiser.

In this paper, we consider a pre-trained deep learning based denoiser $H_{\theta_0}$, more specifically a pre-trained DnCNN network for denoising, and compare the recoveries obtained using it with the recoveries obtained using a classical denoiser $H_\sigma$ (here, BM3D denoiser). To have a fairer comparison, we even fixed the denoising strength of the classical denoiser, i.e., we don't tune the parameter $\sigma$ for an effective denoising or regularizing the recovered solution of the inverse problem, rather, we use a fixed the denoiser $H_{\sigma_0}$ for the recovery process. Furthermore, we even chose a weaker denoiser (i.e., smaller $\sigma$ value) so as to compare the instabilities in the recovered solutions, arising from a weak classical denoiser $H_{\sigma_0}$ vs. a strong learned denoiser $H_{\theta_0}$, i.e., the difference between the lack of adequate (classical) denoising vs. denoising based on some prior learning. We also compare the recoveries obtained using different iterative processes, i.e., the FBS-PnP vs. the ADMM-PnP algorithm, based on the classical denoiser $H_{\sigma_0}$ and the learned denoiser $H_{\theta_0}$. In other words, we show that, not only the denoisers, but also the nature of the iterative flow (even for the same denoiser) significantly influence the efficiency of the recovery process, which is not the case for traditional proximal operators, that are based on some regularization function $\mcal{R}$ in \eqref{Gen. Tik. fun.}.

In addition, we also present techniques to subdue the instabilities arising from these denoisers, for an effective recovery. It is shown that, for stronger denoisers (be it a classical or learned denoiser), the ADMM-PnP algorithm is more effective than the FBS-PnP algorithm, where as, for a weaker denoiser, the FBS-PnP algorithm is better than the ADMM-PnP algorithm, as in the FBS-PnP method the data-consistency steps are improved gradually, and hence, a weaker denoiser can denoise the creeping noise effectively, in contrast, for a stronger denoiser, the FBS-PnP algorithm will easily over-smooth the recovery process, and in this case, the ADMM-PnP algorithm is much more effective. These statements are (empirically) validated via numerical and computational examples in the following section.

\section{Numerical Examples}\label{Sec. Numerical Results}
In this section, we present certain computational results to validate the reasoning  provide in the previous sections. Note that, the goal here is to compare the recoveries obtained using a pre-trained learned denoiser $H_{\theta_0}$ and a fixed classical denoiser $H_{\sigma_0}$, corresponding to the FBS-PnP and ADMM-PnP algorithm. Hence, we don't repeat the experiments over and over to fine tune the denoising parameters $\theta$ and/or $\sigma$, respectively, to produce the optimal results, rather, for the fixed $\theta_0$ and $\sigma_0$, we study the instabilities arising from these denoisers and suggest appropriate measures to subdue them and improve the recovery process. 

\begin{figure}[h!]
    \centering
    \begin{subfigure}{0.495\textwidth}
        \includegraphics[width=\textwidth]{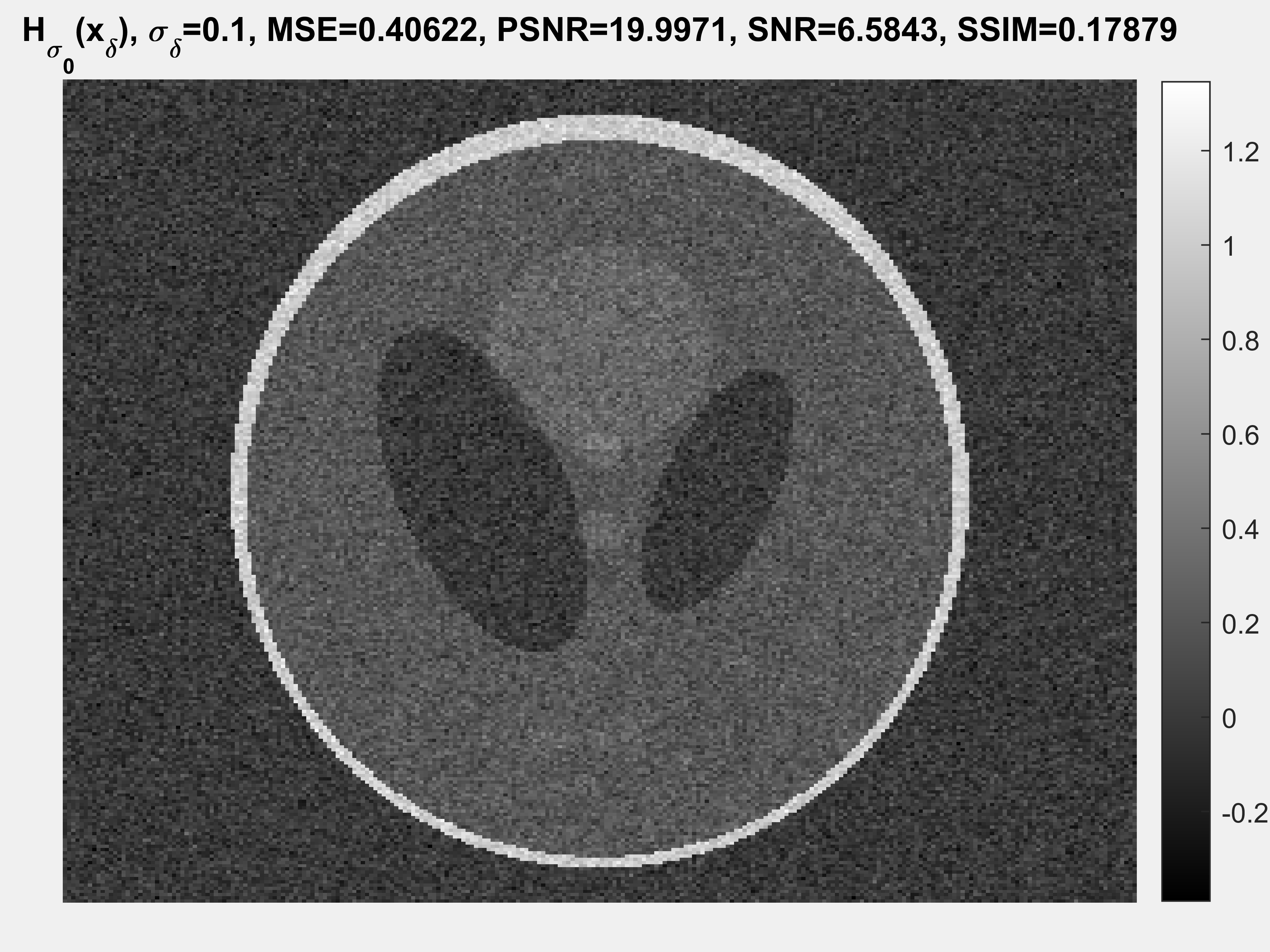}
        \caption{$H_{\sigma_0}(x_{\sigma_\delta}), \sigma_\delta=0.1$}
        \label{BM3D_pt1}
    \end{subfigure}
    \begin{subfigure}{0.495\textwidth}
        \includegraphics[width=\textwidth]{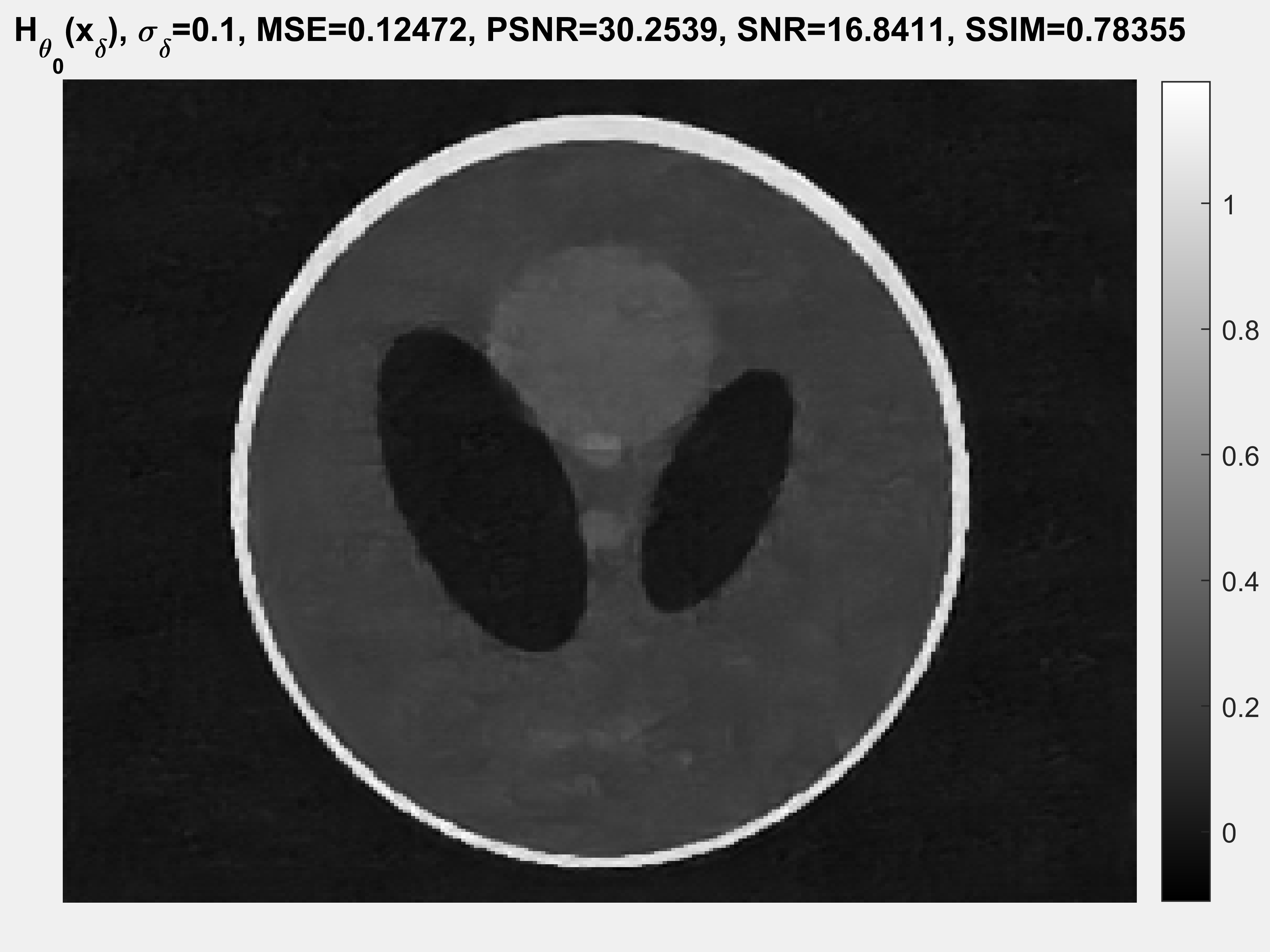}
        \caption{$H_{\theta_0}(x_{\sigma_\delta}), \sigma_\delta=0.1$}
        \label{DnCNN_pt1}
    \end{subfigure}    
    \begin{subfigure}{0.495\textwidth}
        \includegraphics[width=\textwidth]{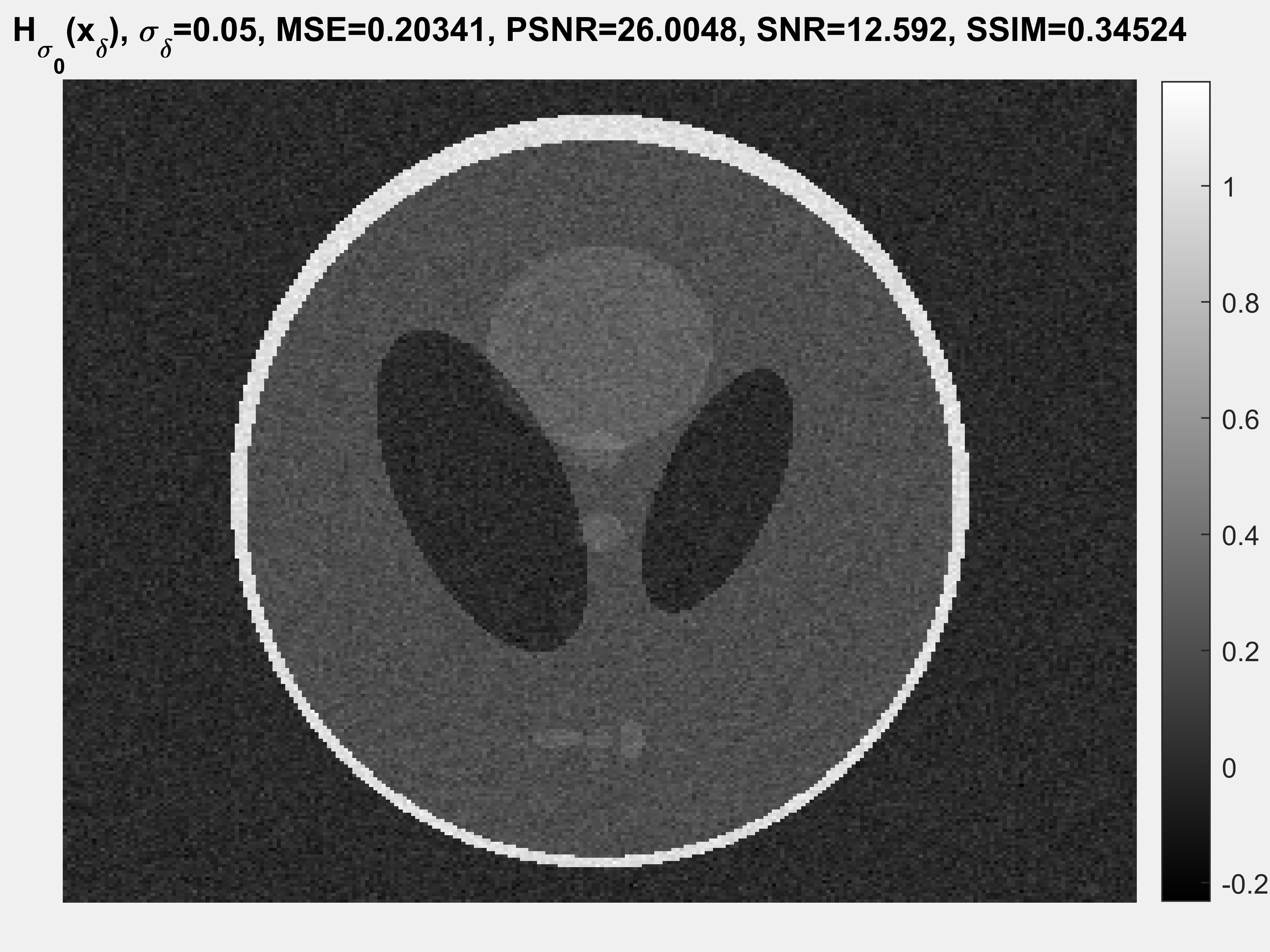}
        \caption{$H_{\sigma_0}(x_{\sigma_\delta}), \sigma_\delta=0.05$}
    \end{subfigure}
    \begin{subfigure}{0.495\textwidth}
        \includegraphics[width=\textwidth]{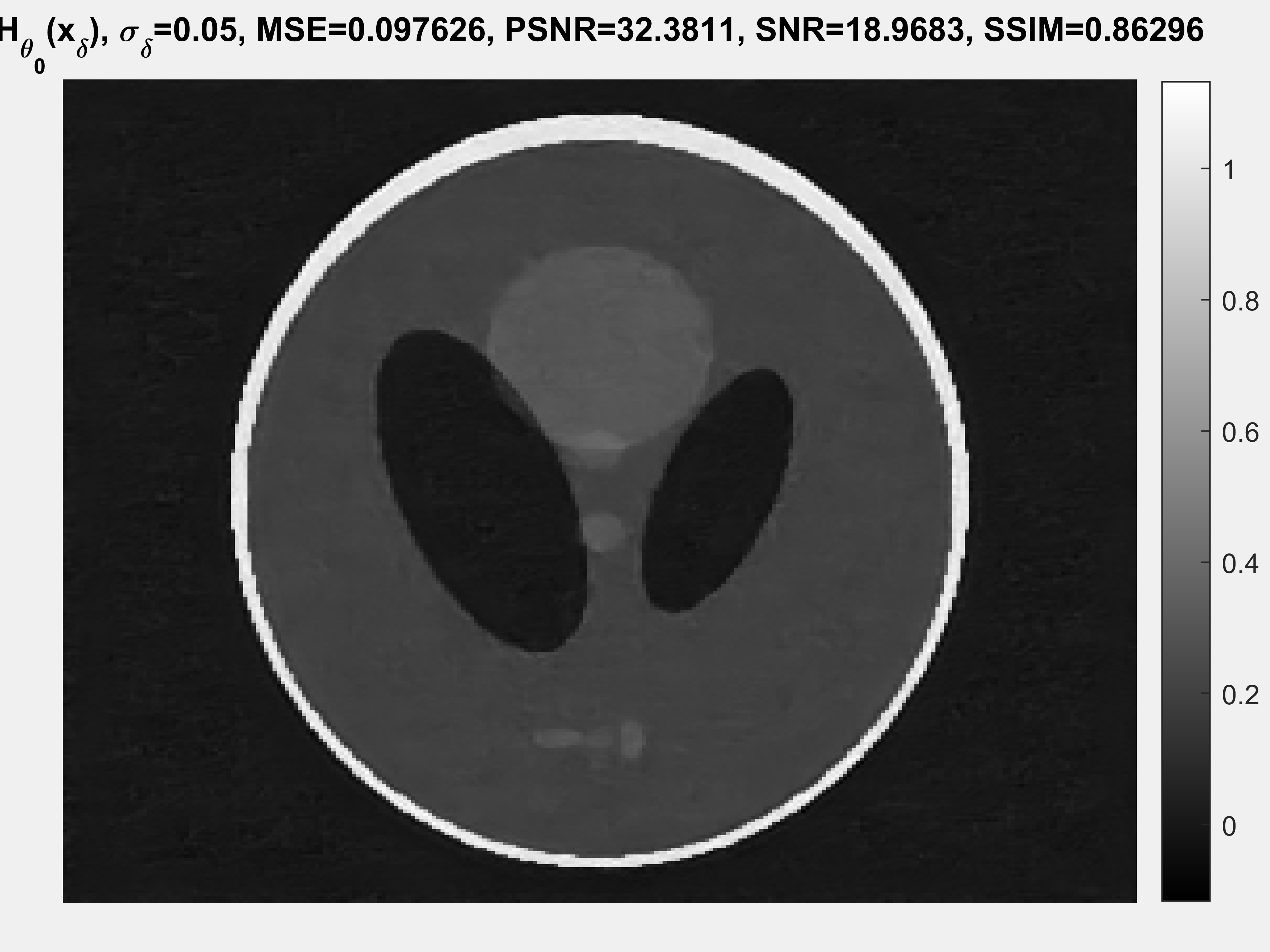}
        \caption{$H_{\theta_0}(x_{\sigma_\delta}), \sigma_\delta=0.05$}
    \end{subfigure}
    \begin{subfigure}{0.495\textwidth}
        \includegraphics[width=\textwidth]{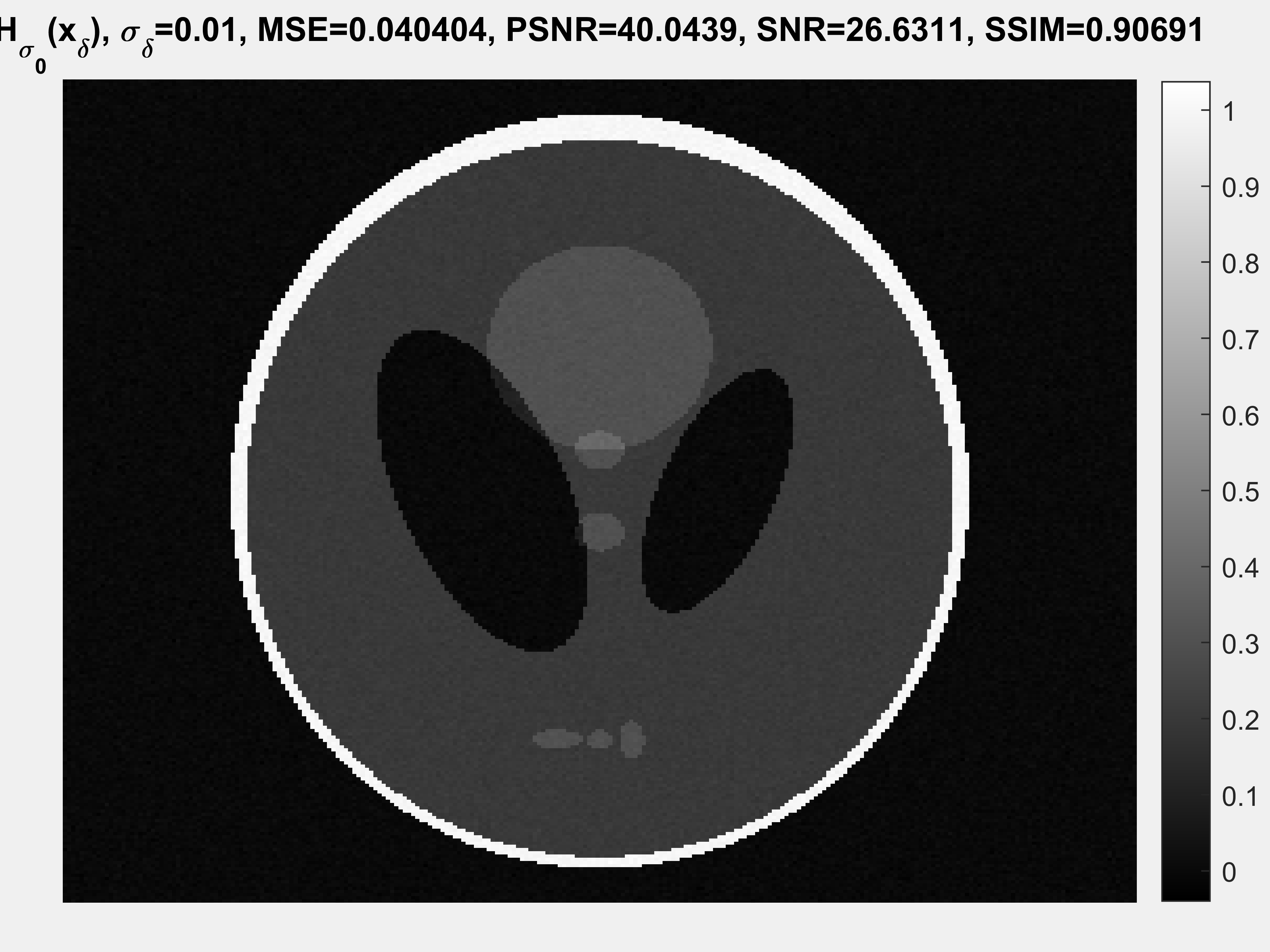}
        \caption{$H_{\sigma_0}(x_{\sigma_\delta}), \sigma_\delta=0.01$}
        \label{BM3D_pt01}
    \end{subfigure}
    \begin{subfigure}{0.495\textwidth}
        \includegraphics[width=\textwidth]{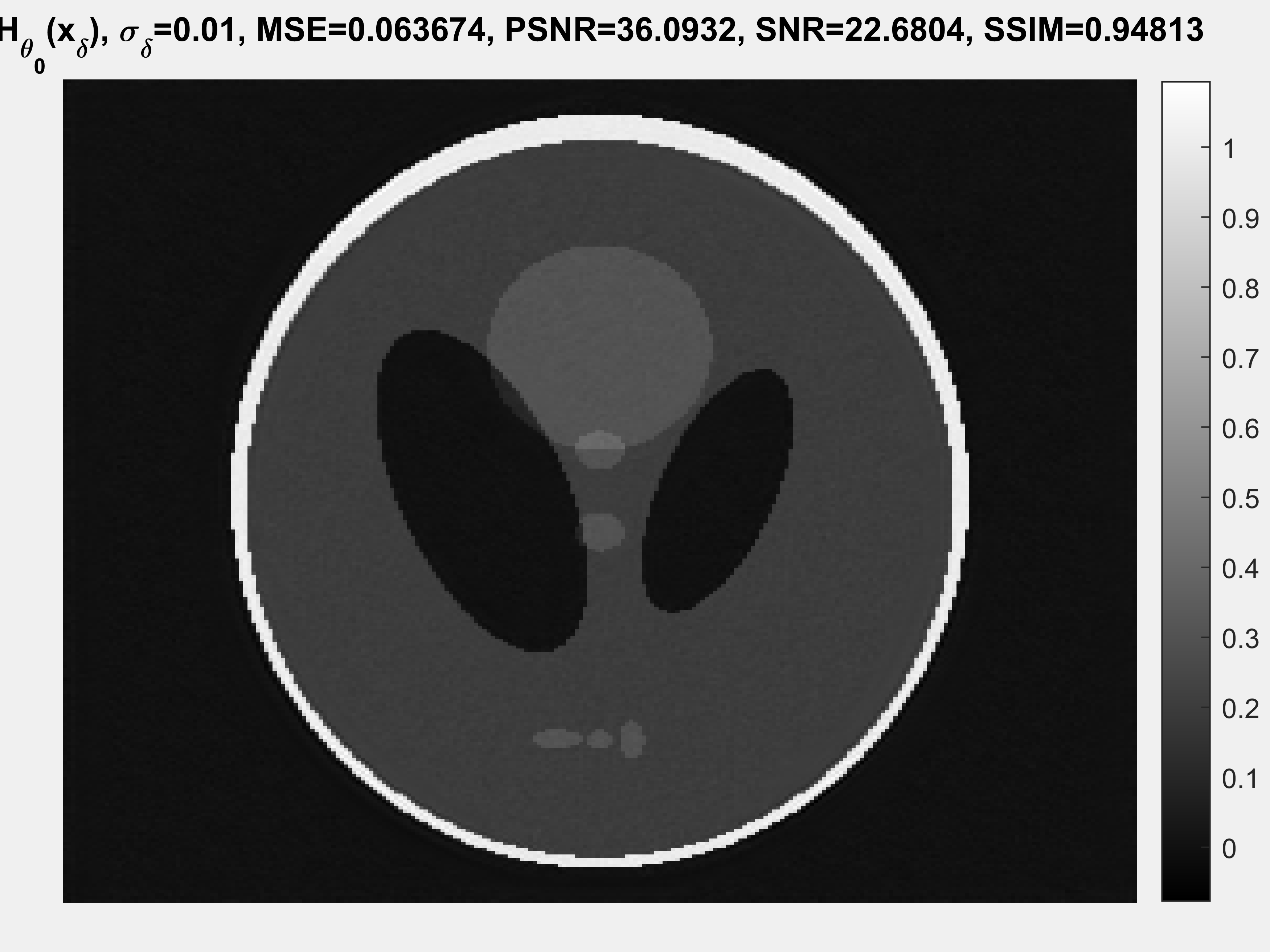}
        \caption{$H_{\theta_0}(x_{\sigma_\delta}), \sigma_\delta=0.01$}
        \label{DnCNN_pt01}
    \end{subfigure}      
    \begin{subfigure}{0.495\textwidth}
        \includegraphics[width=\textwidth]{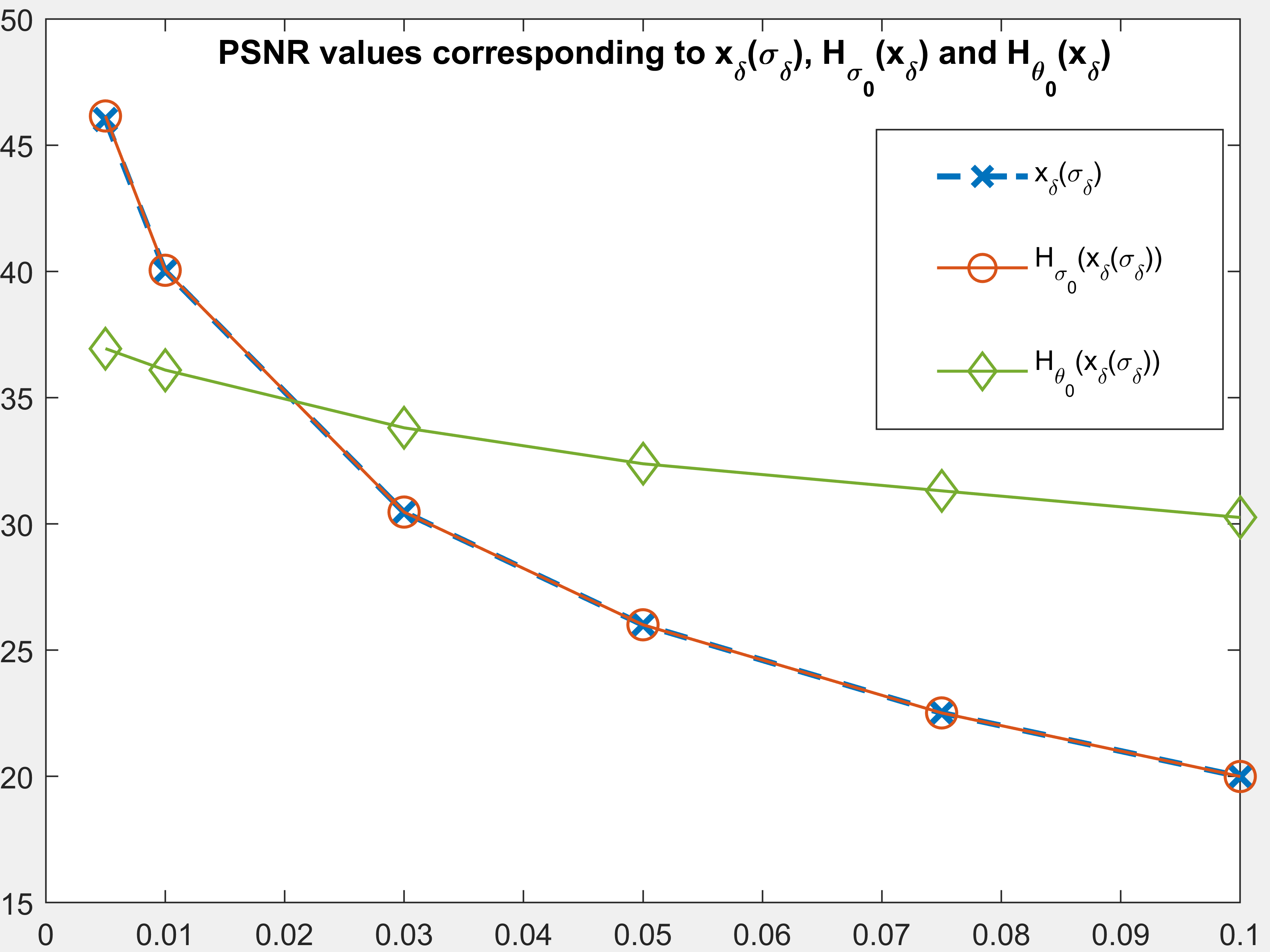}
        \caption{PSNR values vs. noise levels $\sigma_\delta$}
        \label{PSNRcomparison}
    \end{subfigure}
    \begin{subfigure}{0.495\textwidth}
        \includegraphics[width=\textwidth]{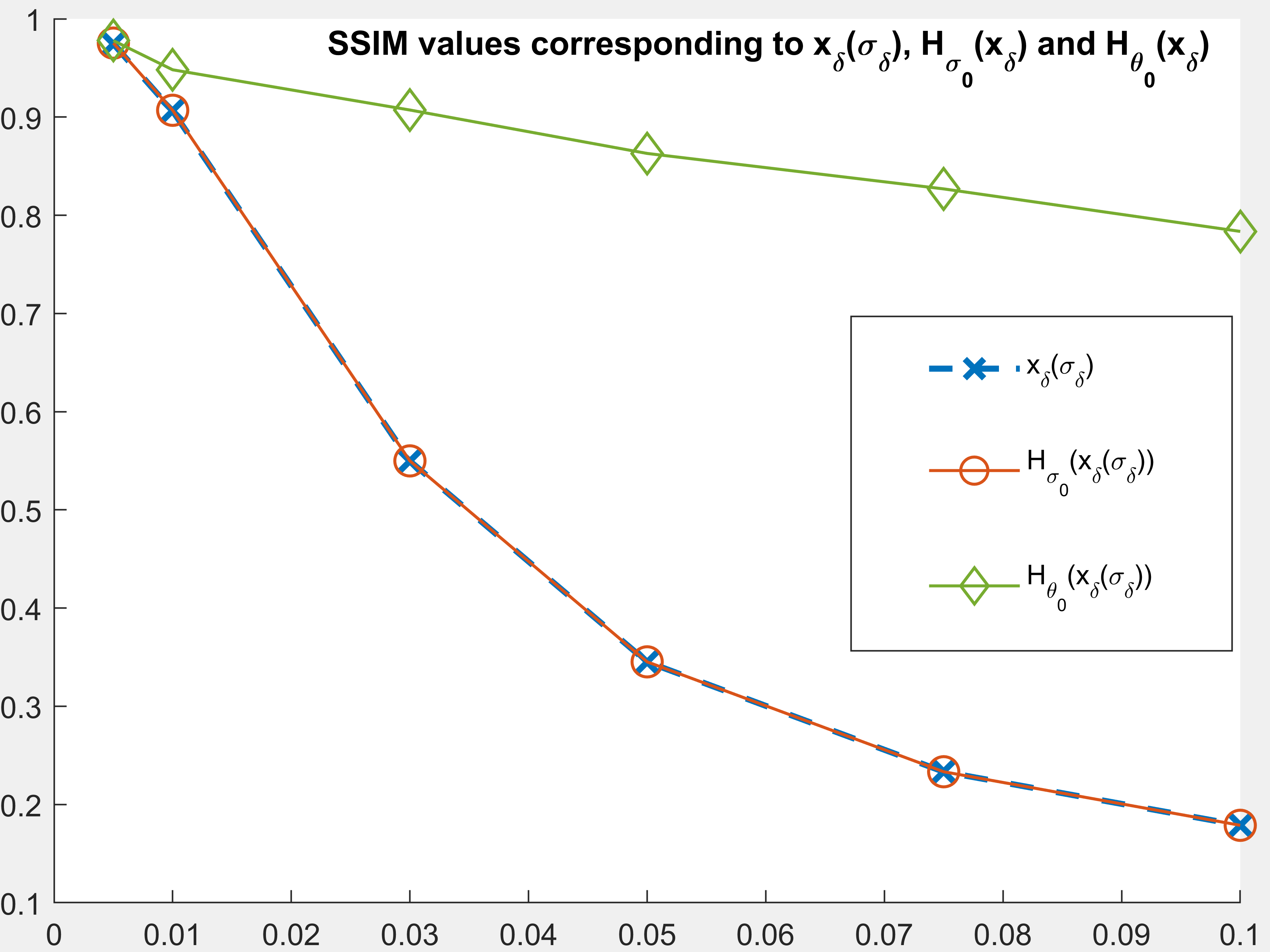}
        \caption{SSIM values vs. noise levels $\sigma_\delta$}  
        \label{SSIMcomparison}
    \end{subfigure}     
    \caption{Comparing denoising strength of $H_{\theta_0}$ vs. $H_{\sigma_0}$.} 
    \label{Figure classical vs learned denoiser ability}
\end{figure}

\begin{figure}[h!]
    \centering
    \begin{subfigure}{0.495\textwidth}
        \includegraphics[width=\textwidth]{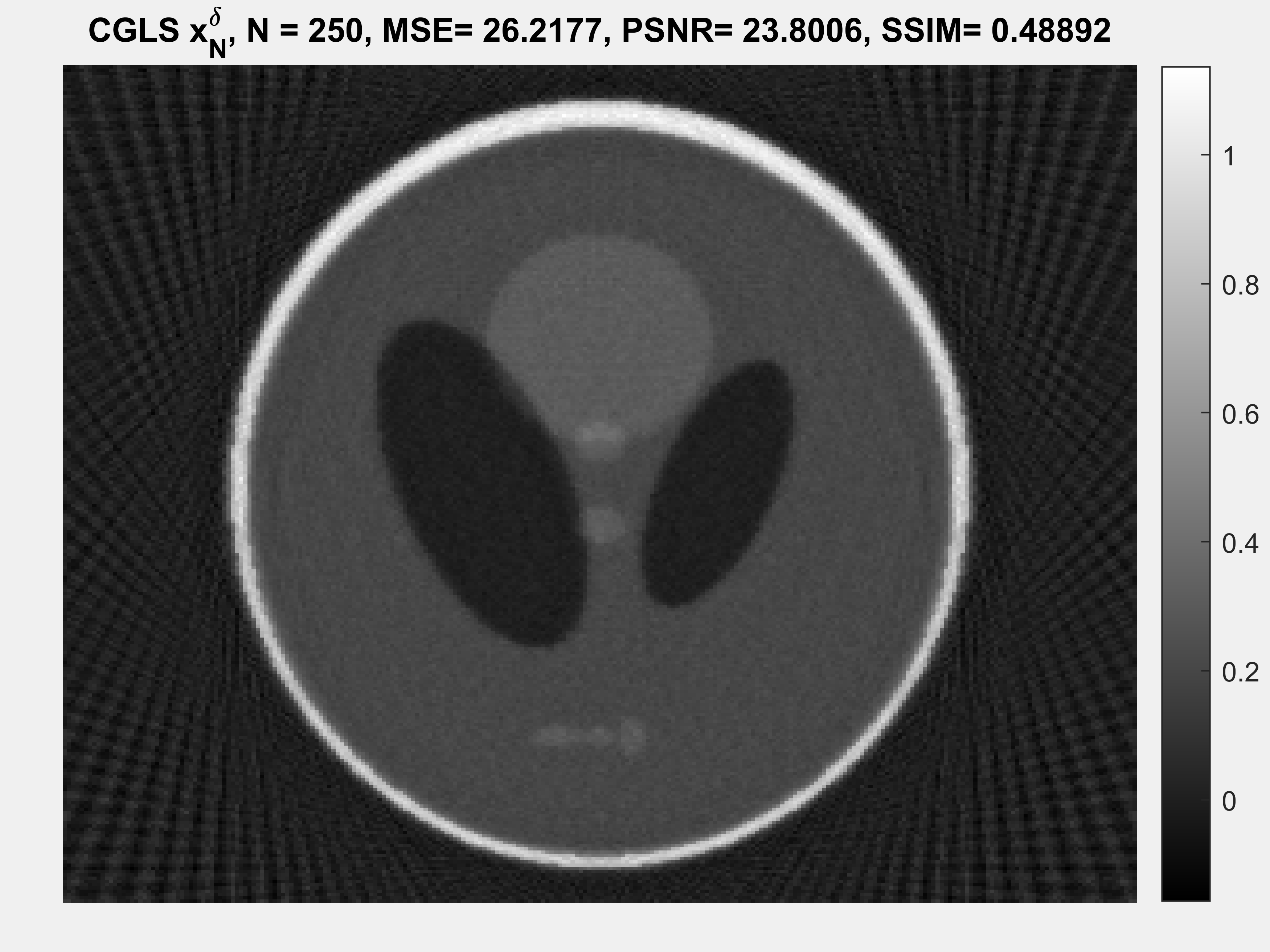}
        \caption{CGLS $x_N^\delta$ for N=250 and no denoisers.}
        \label{Fig CGLS x_N}
    \end{subfigure}       
    \begin{subfigure}{0.495\textwidth}
        \includegraphics[width=\textwidth]{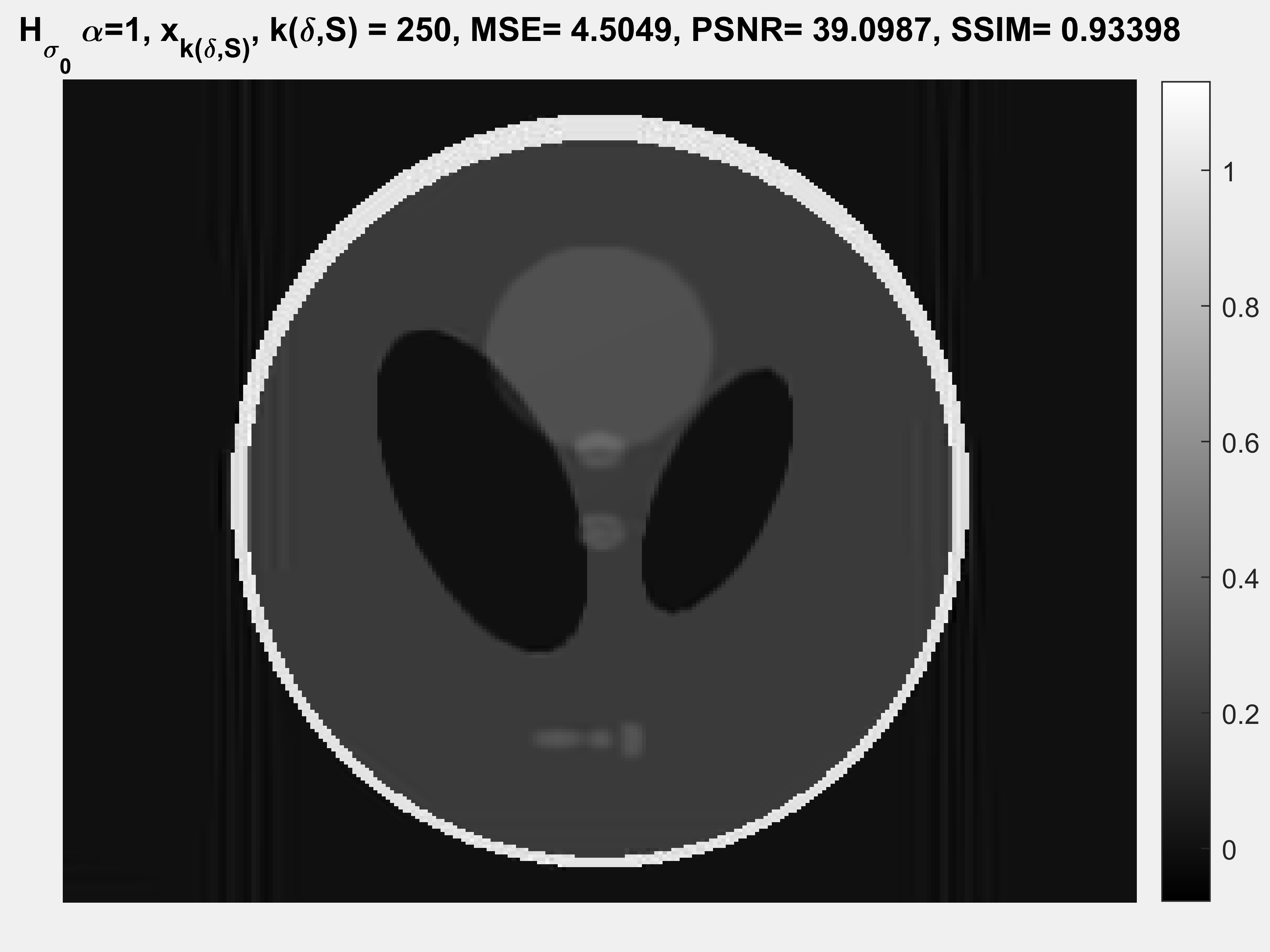}
        \caption{$x_{k(\delta,\mcal{S}_0)}$ for $k(\delta,\mcal{S}_0)=250$ and $H_{\sigma_0, \alpha=1}$}
        \label{Fig FBS-BM3D alpha 1}
    \end{subfigure}       
    \begin{subfigure}{0.495\textwidth}
        \includegraphics[width=\textwidth]{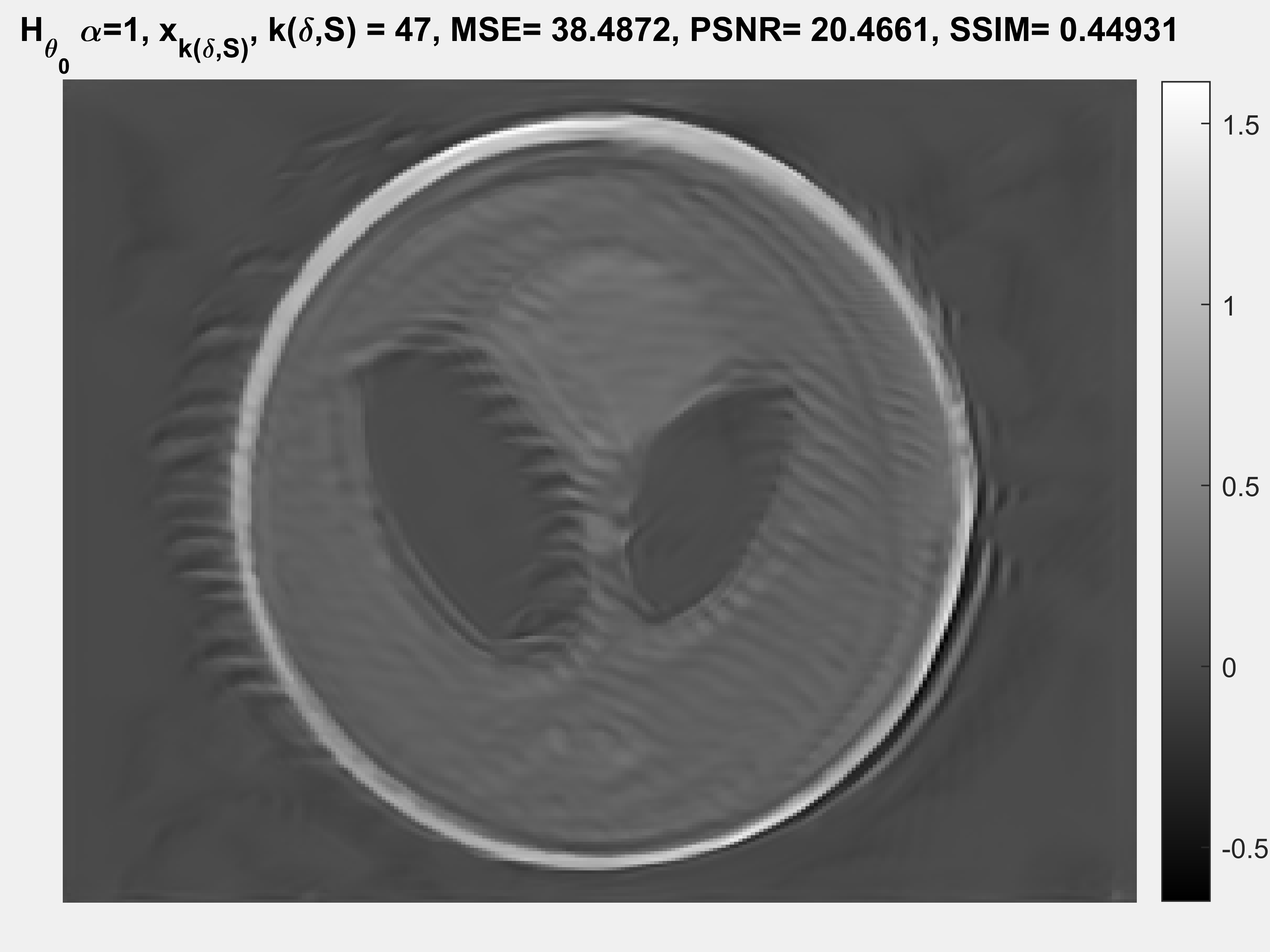}
        \caption{$x_{k(\delta,\mcal{S}_0)}$ for $k(\delta,\mcal{S}_0)=47$ and $H_{\theta_0, \alpha=1}$}
        \label{Fig FBS-DnCNN alpha 1}
    \end{subfigure} 
    \begin{subfigure}{0.495\textwidth}
        \includegraphics[width=\textwidth]
        {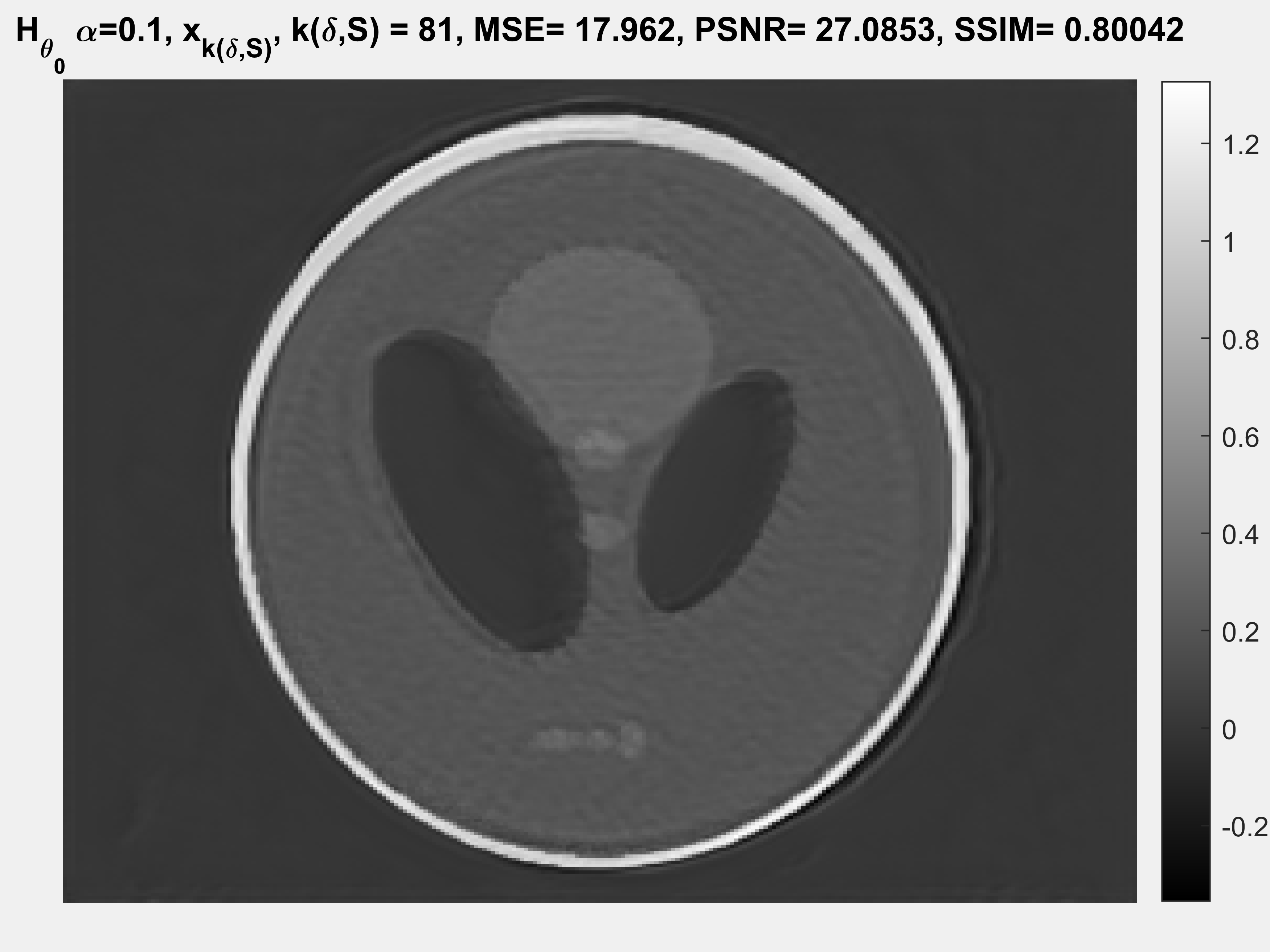}
        \caption{$x_{k(\delta,\mcal{S}_0)}$ for $k(\delta,\mcal{S}_0)=81$ and $H_{\theta_0, \alpha=0.1}$}
        \label{Fig FBS-DnCNN alpha 0.1}
    \end{subfigure}    
    \begin{subfigure}{0.495\textwidth}
        \includegraphics[width=\textwidth]
        {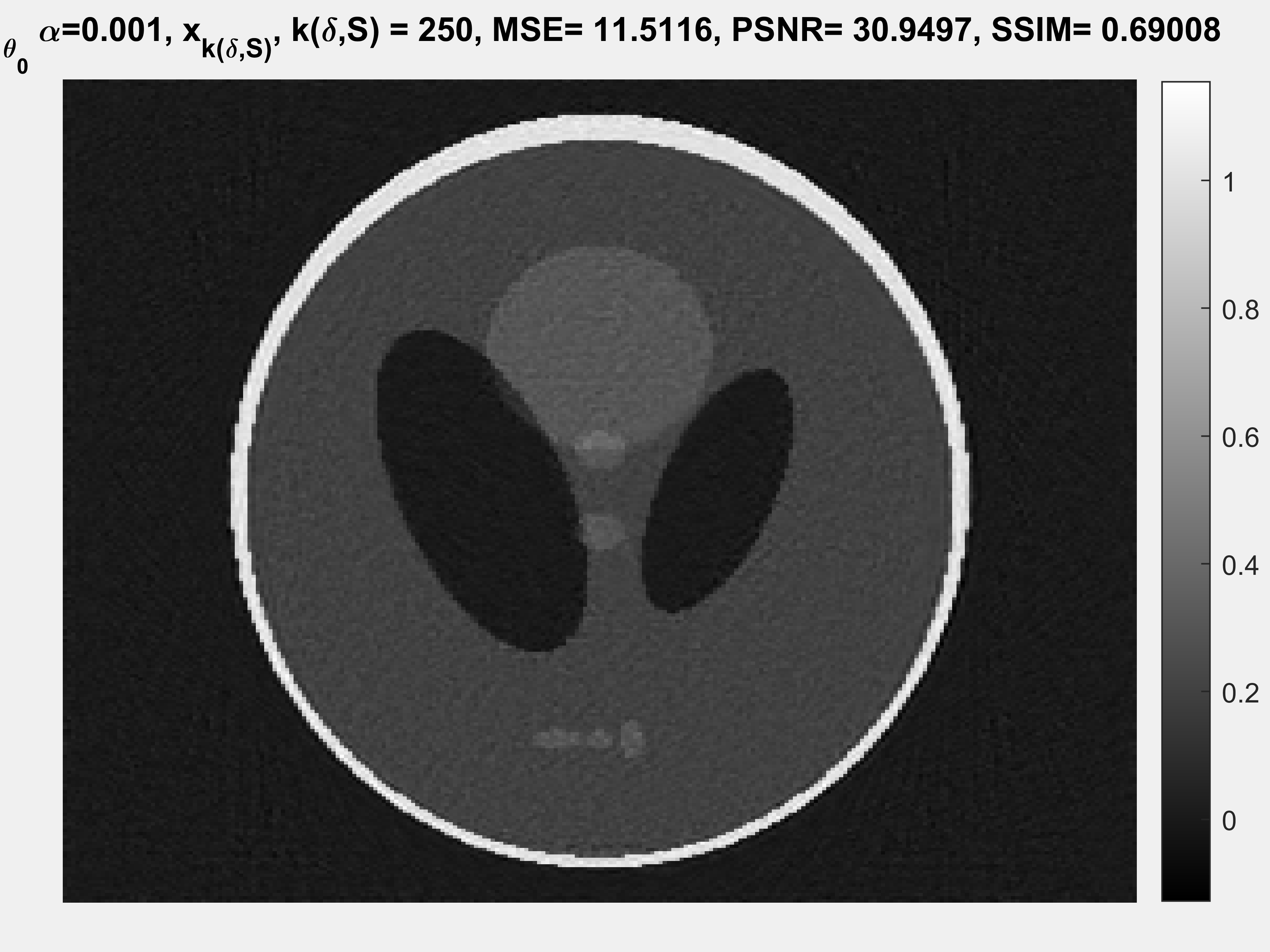}
        \caption{$x_{k(\delta,\mcal{S}_0)}$ for $k(\delta,\mcal{S}_0)=250$, $H_{\theta_0, \alpha=0.001}$}
        \label{Fig FBS-DnCNN alpha 0.001}
    \end{subfigure}
    \begin{subfigure}{0.495\textwidth}
        \includegraphics[width=\textwidth]
        {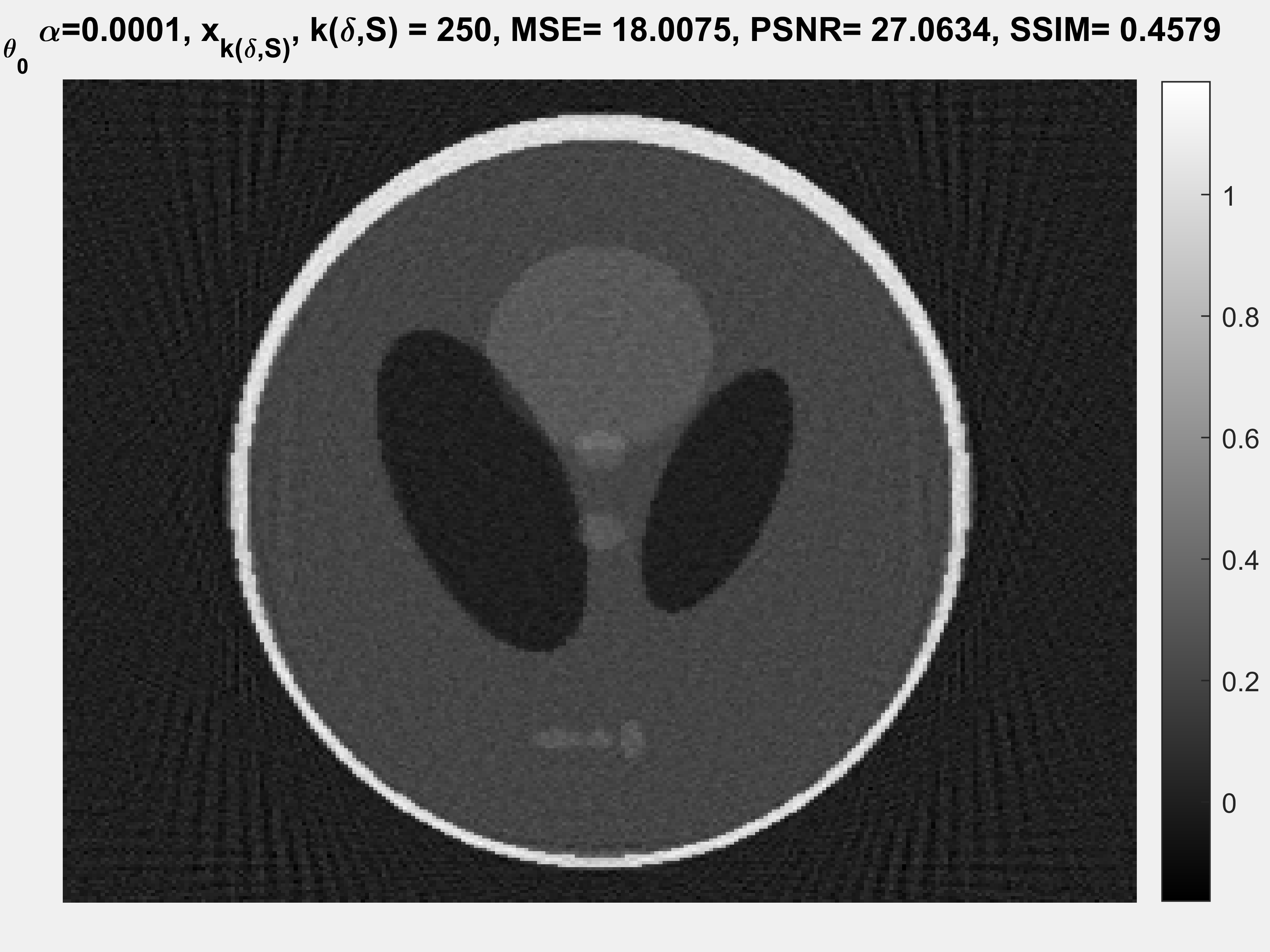}
        \caption{$x_{k(\delta,\mcal{S}_0)}$ for $k(\delta,\mcal{S}_0)=205$, $H_{\theta_0, \alpha=0.0001}$}
        \label{Fig FBS-DnCNN alpha 0.0001}
    \end{subfigure}
    \begin{subfigure}{0.495\textwidth}
        \includegraphics[width=\textwidth]
        {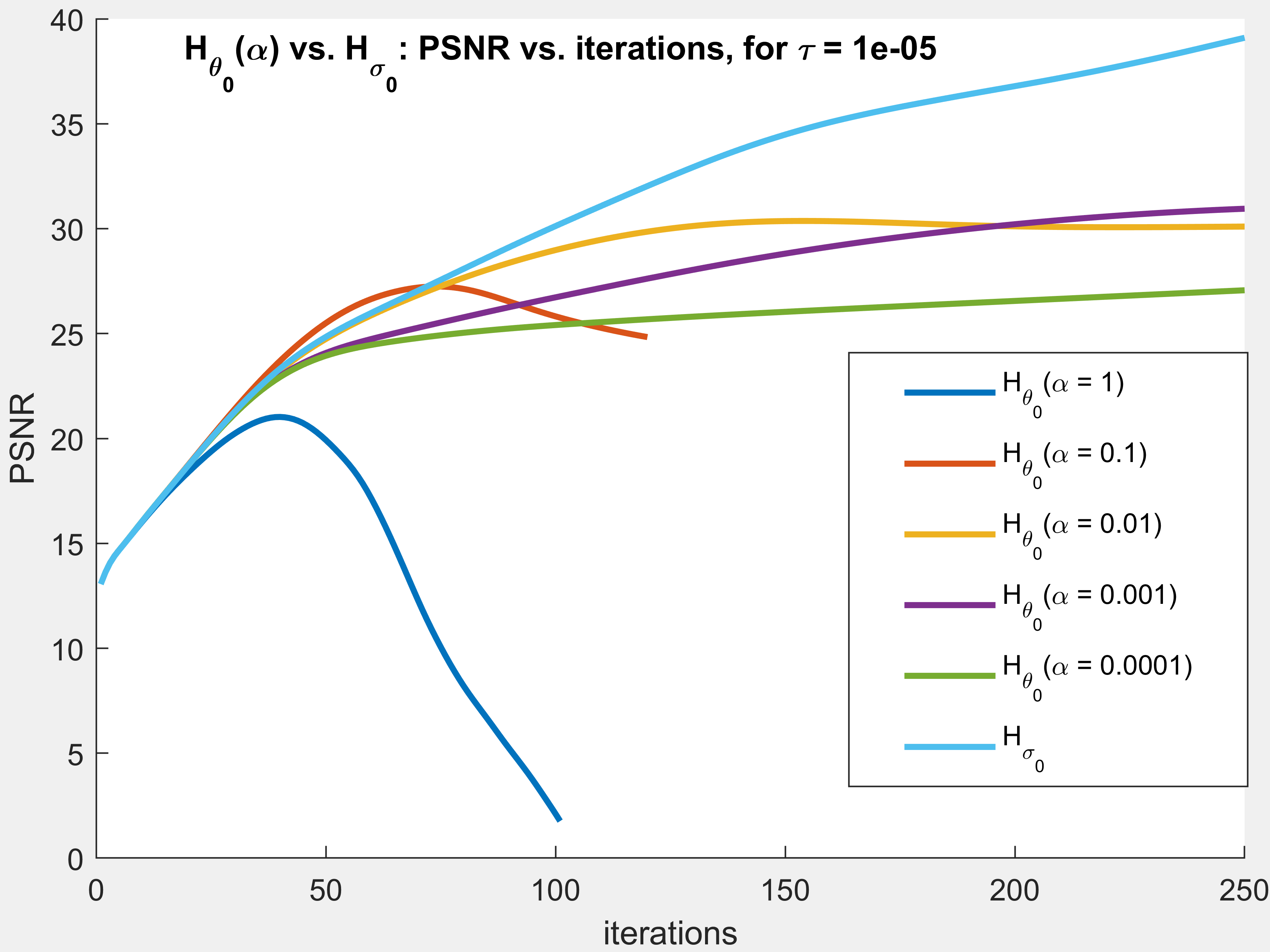}
        \caption{PSNR curves, $H_{\theta_0}(\alpha)$ vs. $H_{\sigma_0}$}
        \label{Fig FBS-PnP PSNR}
    \end{subfigure}
    \begin{subfigure}{0.495\textwidth}
        \includegraphics[width=\textwidth]
        {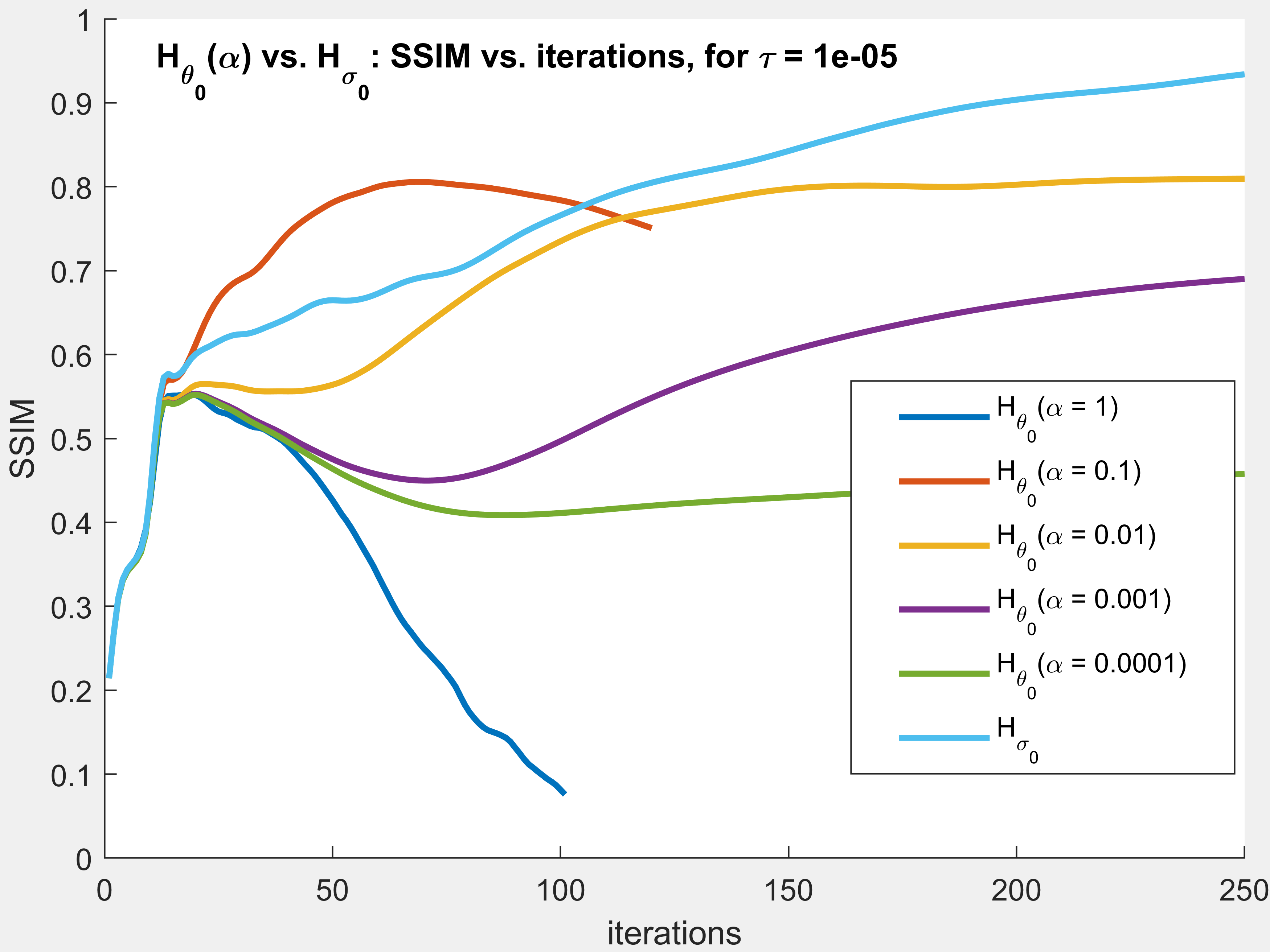}
        \caption{SSIM curves, $H_{\theta_0}(\alpha)$ vs. $H_{\sigma_0}$}
        \label{Fig FBS-PnP SSIM}
    \end{subfigure}    
    \caption{FBS-PnP: denoiser $H_{\theta_0}(\alpha)$ vs. $H_{\sigma_0}$, see Example \ref{Example FBS-PnP}.} 
    \label{Figure FBS-PnP}
\end{figure}

All the experiments are computed in MATLAB, where we consider the classical denoiser $H_{\sigma_0}$ as the BM3D denoiser with $\sigma_0 = 0.001$, and the MATLAB code for the BM3D denoiser is obtained from \href{http://www.cs.tut.fi/~foi/GCF-BM3D/}{http://www.cs.tut.fi/~foi/GCF-BM3D/}, which is based on \cite{Ymir_Azzari_Foi_1, Ymir_Azzari_Foi_2}. Here, we kept all the attributes of the code in their original (default) settings and assign the denoising strength $\sigma = 0.001$, as the standard deviation of the noise, when denoising $x_k^\delta$ iterates to $z_k^\delta$. And for the learned denoiser $H_{\theta_0}$, we used MATLAB's pre-trained DnCNN denoiser, the details of which (such as the number of layers, optimization procedures etc.) can be found in 
\\
\href{https://www.mathworks.com/help/images/ref/denoisingnetwork.html}{https://www.mathworks.com/help/images/ref/denoisingnetwork.html}.
First, to compare the effectiveness of their denoising abilities, we implement them on noisy Shepp-Logan phantom for different noise levels and 
 the results are shown in Figure \ref{Figure classical vs learned denoiser ability}, where $\sigma_\delta$ denotes the standard deviation of the additive Gaussian noise with zero-mean. 
 Observe that, the denoising ability of the learned denoiser $H_{\theta_0}$ is very impressive overall, irrespective of the noise levels, where as, the denoising from the classical denoiser $H_{\sigma_0}$ is (practically) negligible for all noise levels. Also, note that, when the noise level ($\sigma_\delta$) decreases, the performance metrics of $H_{\theta_0}$ (as well as $H_{\sigma_0}$, though insignificantly) increases, however, there is a noticeable difference in the dynamics of different evaluation metrics, such as, the PSNR values of the noisy data $x_{\sigma_\delta}$ (for smaller $\sigma_\delta$ values) are even better than that of the (learned) denoised image $H_{\theta_0}(x_{\sigma_\delta})$, where as, the SSIM values follow a completely different trail for different noise levels, it's always better than the others. Such discrepancies reflect that the denoiser $H_{\theta_0}$ has been trained (or has learned) to emphasize certain features/structures of denoising, over certain others, and hence, can be susceptible to hallucinate (or impose) those features, leading to instabilities (or generating artifacts) that are quite different in nature than the instabilities (or noises) arising from the inherent ill-posedness of the inverse problems, which are reflected in the following examples. In all of the following examples, when the FBS-PnP algorithms is implemented, we consider the step-size to be $\tau = 10^{-5}$ (a constant step-size), unless otherwise stated, and the relative noise levels in the data $\left(\frac{||b_\delta - b||}{||b||}\right)$, the number of iterations, the Lagrangian parameter ($\rho$) value, when using ADMM-PnP algorithm, etc. are specified in the example settings. The data-consistency term is consider to be $\mcal{D}(x) = ||Ax - b_\delta||_2^2$ and the selection criterion $\mcal{S}_0$, for the regularized solution, is considered to be the cross validation criterion, for some leave-out set of the noisy data $b_\delta$ (which is 1\% of $b_\delta$). The numerical values, corresponding to the recovered solutions, are shown in Tables \ref{Table FBS-PnP}, \ref{Table ADMM-PnP}, \ref{Table FBS-PnP combined denoisers} and \ref{Table ADMM-PnP combined denoisers}, where MSE denotes the mean-squared error (calculated as $\frac{||x_\delta - x||}{||x||}$), PSNR stands for the peak signal-to-noise ratio in dB (computed using MATLAB's inbuilt function $psnr(x_\delta,x)$), SSIM stands for the structure similarity index measure (which is again computed using MATLAB's inbuilt routine $ssim(x_\delta,x)$), $`\mcal{D}$-err.' stands for the discrepancy error $\left(\frac{||Ax_k^\delta - b_{D,\delta}||_2}{||b_{D,\delta}||_2}\right)$ and $`\mcal{S}$-err.' stands for the cross-validation error $\left(\frac{||Ax_k^\delta - b_{S,\delta}||_2}{||b_{S,\delta}||_2}\right)$, where $b_{S,\delta} \subset b_\delta$ is the left-out set and $b_{D,\delta}= b_\delta\backslash b_{S,\delta}$, and the recoveries are shown in Figures \ref{Figure FBS-PnP}, \ref{Figure ADMM-PnP} and \ref{Figure FBS-PnP and ADMM-PnP combined denoiser}.

In the following examples, the matrix equation \eqref{Inv. prob.} corresponds to the discretization of a radon transformation, which is associated with the X-ray computed tomography (CT) reconstruction problem, where we generate the matrix $A \in \mbb{R}^{m \times n}$, $\hat{x} \in \mbb{R}^n$ and $b \in \mbb{R}^m$ from the MATLAB codes presented in \cite{Hansen_IRtools}. The dimension $n$ corresponds to the size of a $N \times N$ image, i.e., $n = N^2$, and the dimension $m$ is related to the number of rays per projection angle and the number of projection angles, i.e., $m = M_1 \times M_2$, where $M_1$ implies the number of rays/angle and $M_2$ implies the number of angles.

\begin{example}\label{Example FBS-PnP}\textbf{[Fast FBS-PnP using $H_{\theta_0}$ vs. $H_{\sigma_0}$]}\\
In this example, we compare the recoveries obtained in the Fast FBS-PnP algorithm, when using a learned denoiser $H_{\theta_0}$ vs. when using a classical denoiser $H_{\sigma_0}$. We present the nature of instabilities arising from a learned denoiser $H_{\theta_0}$, when used in a FBS-PnP algorithm, and provide a technique to subdue them. Here, for the true phantom, we consider the standard ($256\times 256$) Shepp-Logan phantom ($\hat{x} \in \mbb{R}^{65536}$ and $\hat{x}_i \in [0,1]$). However, during the recovery process, we do not enforce the constraint $x_i \geq 0$ on the iterates, since the motive is to compare the efficiency of these two denoisers, while solving an inverse problem, independent of any constraints. The matrix $A\in \mbb{R}^{43440 \times 65536}$ is generated using the $PRtomo()$ code from \cite{Hansen_IRtools}, corresponding to a `fancurved' CT problem  with only 120 view angles (which are evenly spread over $360^o$). The noiseless data is generated by $b := A\hat{x} \in \mbb{R}^{43440 (=362 * 120)}$, which is then contaminated by additive Gaussian noise of zero-mean to produce noisy data $b_\delta$ such that the relative error is around 1\%. We leave out 1\% of the noisy data $b_\delta$ for generating the cross-validation errors (the selection criterion $\mcal{S}_0$) during the iterative process and consider a constant step-size $\tau = 10^{-5}$. The iterations are terminated if the cross-validation errors start increasing steadily and continuously, after allowing certain number of small fluctuations, or if the iterations have reached the maximum limit (250 iterations), unless otherwise stated. The numerical values corresponding to the recoveries are shown in Table \ref{Table FBS-PnP} and figures in Figure \ref{Figure FBS-PnP}.

Note that, Figure \ref{Fig CGLS x_N} shows the recovered solution without using any denoisers, i.e., the solution after 250 Conjugate-Gradient Least-Squares (CGLS) iterations, and one can notice the noisy texture in the image and certain artifacts arising from the ill-posedness of the problem. Where as, Figure \ref{Fig FBS-DnCNN alpha 1} shows the regularized solution ($x_{k(\delta,\mcal{S}_0)}^\delta$, for $k(\delta,\mcal{S}_0)=47$) when using the learned denoiser $H_{\theta_0}$, where the nature of instabilities (or artifacts) are quite different than that in Figure \ref{Fig CGLS x_N}, without any denoiser. The reason being, as is explained above, the learned denoiser $H_{\theta_0}$ has learned certain features/structures to impose on the images, which it considers as denoising, especially for images with lower noise levels. And, in a Fast FBS-PnP algorithm, the iterates $x_k^\delta$ are improved gradually to fit the (noisy) data $b_\delta$, implying that the initial iterates are less noisy, and hence, $H_{\theta_0}$ imposes certain structures to them, which are then transformed into corrupted artifacts over later iterations. Where as, in an ADMM-PnP algorithm the iterates $x_k^\delta$ approximates the noisy LS-solution $x_\delta^\dagger$ very rapidly (for smaller values of the Lagrangian parameter $\rho$), and thus, the iterates are heavily contaminated with noise arising from the noisy data ($b_\delta$), which then can be effectively denoised by $H_{\theta_0}$, see Example \ref{Example ADMM-PnP}.

In addition, note that, if the iterations were not terminated at $k(\delta,\mcal{S}_0)$, then the relative error in the recovered solution for the last iterate ($x_N^\delta$) would be enormous, i.e., the relative errors in the recovery process have a semi-convergence nature, see Figure \ref{Fig FBS-PnP PSNR} and \ref{Fig FBS-PnP SSIM}. Equivalently, $x^\delta_{k(\delta,\mcal{S}_0)}$ is the best solution during that iterative process, based on $\mcal{S}_0$. However, $x_{k(\delta,\mcal{S}_0)}^\delta$ may not be the most optimal solution during that iterative process, i.e., with the minimal MSE, but since $x^\dagger$ is not known a-priori, the most optimal solution can not be estimated without additional knowledge.

In contrast, the classical (weaker) denoiser $H_{\sigma_0}$ produce a significantly better result, as can be seen in Figure \ref{Fig FBS-BM3D alpha 1}. Again, the reasons being, (1st) it does not hallucinate features or impose structures on a learned basis and, (2nd) the (Fast) FBS-PnP algorithm updates the iterates $x_k^\delta$ gradually to fit the noisy data $b_\delta$, i.e., noise in $x_k^\delta$ appears gradually, which then can be effectively denoised by $H_{\sigma_0}$, even if it's weak, without any hallucinations.  However, $H_{\sigma_0}$ will fail in the ADMM-PnP algorithm, if used naively, as shown in Example \ref{Example ADMM-PnP}, since the noise intensities in the iterates $x_k^\delta$ (for ADMM-PnP algorithm) is very high, due to the large updates in the data-consistency steps towards the noisy data $b_\delta$, and as $H_{\sigma_0}$ is a weaker denoiser (for $\sigma_0 = 0.001$), it can not effectively denoise the noisy iterates $x_k^\delta$, of high noise levels, to produce well-denoised iterates $z_k^\delta$.

\end{example}

\subsection{Subduing the instabilities/artifacts of a learned denoiser}\label{Sec. FBS-PnP attenuation}
\mbox{ }

\begin{figure}[h!]
    \centering
    \begin{subfigure}{0.495\textwidth}
        \includegraphics[width=\textwidth]
        {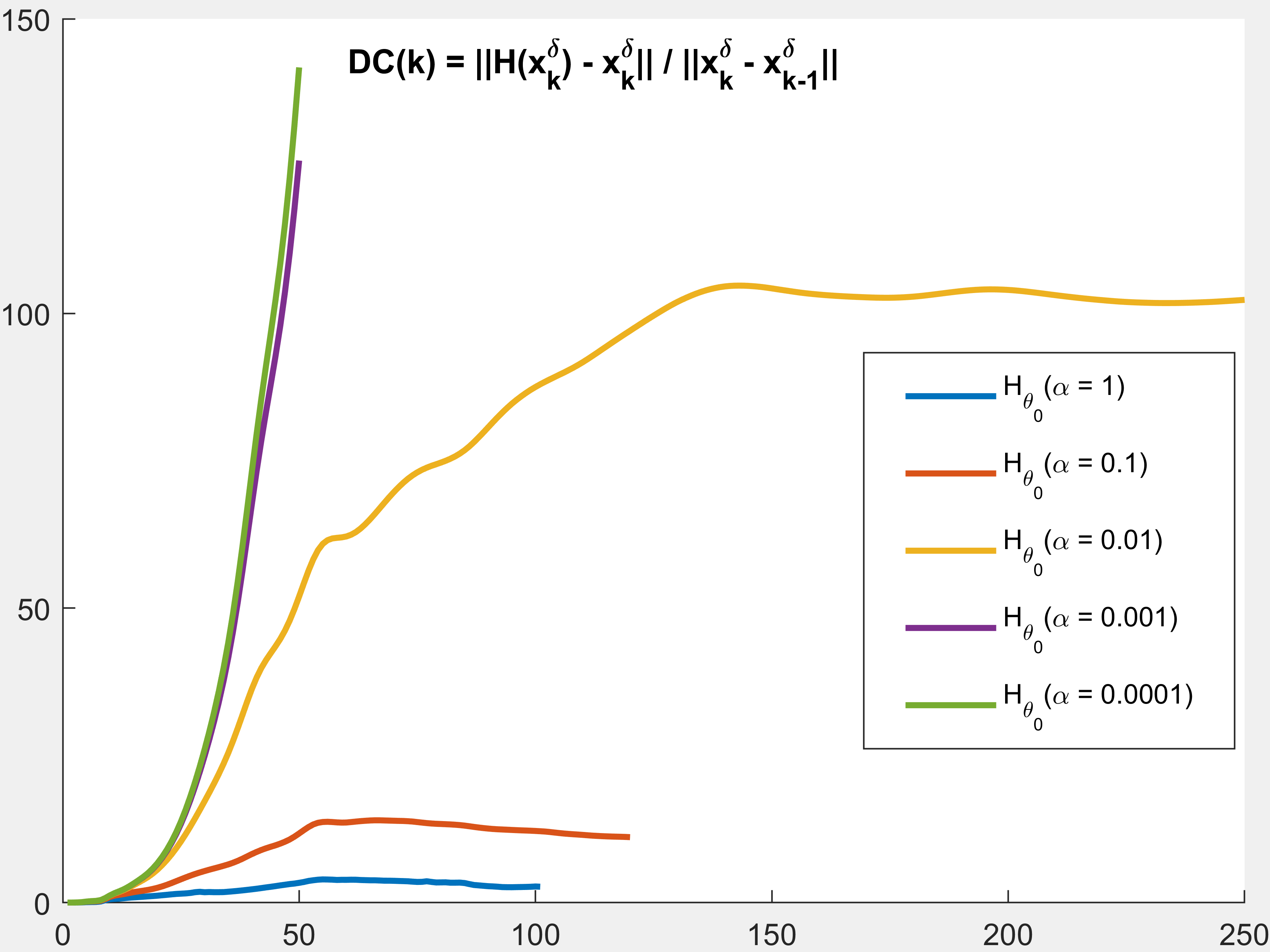}
        \caption{$DC(k,\alpha)$ before attenuating $H_{\theta_0}$}
        \label{Fig FBS-PnP before attenuation}
    \end{subfigure}
    \begin{subfigure}{0.495\textwidth}
        \includegraphics[width=\textwidth]
        {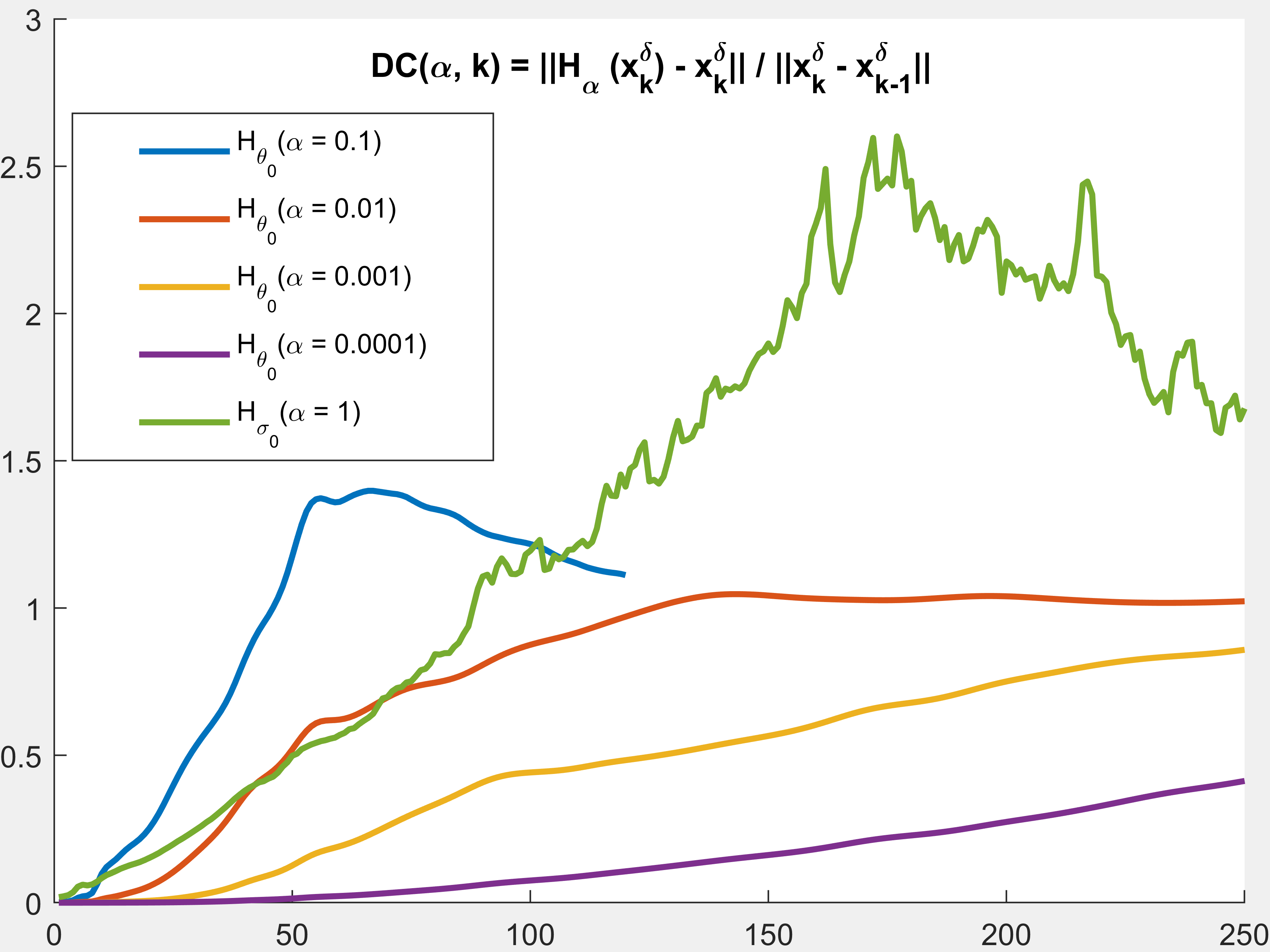}
        \caption{$DC(k,\alpha)$ after $H_{\theta_0}$ attenuated to $H_{\theta_0,\alpha}$}
        \label{Fig FBS-PnP after attenuation}
    \end{subfigure}    
    \caption{Denoising-to-consistency ratio ($DC(k)$), for Example \ref{Example FBS-PnP}.} 
    \label{Figure denoising-to-consistency FBS-PnP}
\end{figure}

As explained in \cite{nayak2021PnP}, we would like to introduce an additional attenuating-parameter $0 \leq \alpha \leq 1$ to attenuate the denoising strength of $H_{\theta_0}$. This can be achieved by (externally) parameterizing $H_{\theta_0}$ to $H_{\theta_0, \alpha}$, where the denoiser $H_{\theta_0, \alpha}$ is defined as follows
\begin{align}\label{Strong deno. alpha0}
H_{\theta_0, \alpha}(x_k^\delta) :=& \; x_k^\delta \; + \; \alpha \left( H_{\theta_0}(x_k^\delta) \; - \; x_k^\delta \right) \\
=& \; (1 - \alpha)\;x_k^\delta \; + \; \alpha \; H_{\theta_0}(x_k^\delta). \notag
\end{align}
Note that, with this transformation, the resulting (new) direction at any kth-step is given by, for the (new) denoised iterate $z_k^\delta(\alpha) := H_{\theta_0,\alpha}(x_k^\delta)$,
\begin{align}\label{Strong deno. alpha1}
d_k^\delta(\alpha) &= z_k^\delta(\alpha) - z_{k-1}^\delta \notag \\
&= (x_k^\delta - z_{k-1}^\delta) \; + \; \alpha \left( H_{\sigma_k}(x_k^\delta) - x_k^\delta \right).
\end{align} 
Hence, if the denoising-to-consistency ratio $DC(k) >> 1$ (very large), which can indicate over-denoising, then by opting a smaller $\alpha$ value ($\alpha << 1$), one can reduce the extent of denoising and can obtain a well-regularized solution. Table \ref{Table FBS-PnP} shows the performance metrics of the recoveries, obtained using different values of $\alpha$, and Figure \ref{Figure FBS-PnP} shows the corresponding recovered solutions. Figures \ref{Fig FBS-PnP before attenuation} and \ref{Fig FBS-PnP after attenuation} show the graph of denoising-to-consistency ratio ($DC(k,\alpha)$ vs. $k$), for different values of $\alpha$, before and after attenuating the denoising strength of $H_{\theta_0}$ to $H_{\theta_0,\alpha}$. One can see that, the ratios $DC(k,\alpha)$ are quite high before attenuating the denoiser and are moderate after subduing it, indicating suitable denoising. However, for very small value of $\alpha$, the ratio $DC(k,\alpha) << 1$, indicating inadequate denoising, which is also reflected in the recovery. 

\begin{figure}[h!]
    \centering
    \begin{subfigure}{0.495\textwidth}
        \includegraphics[width=\textwidth]{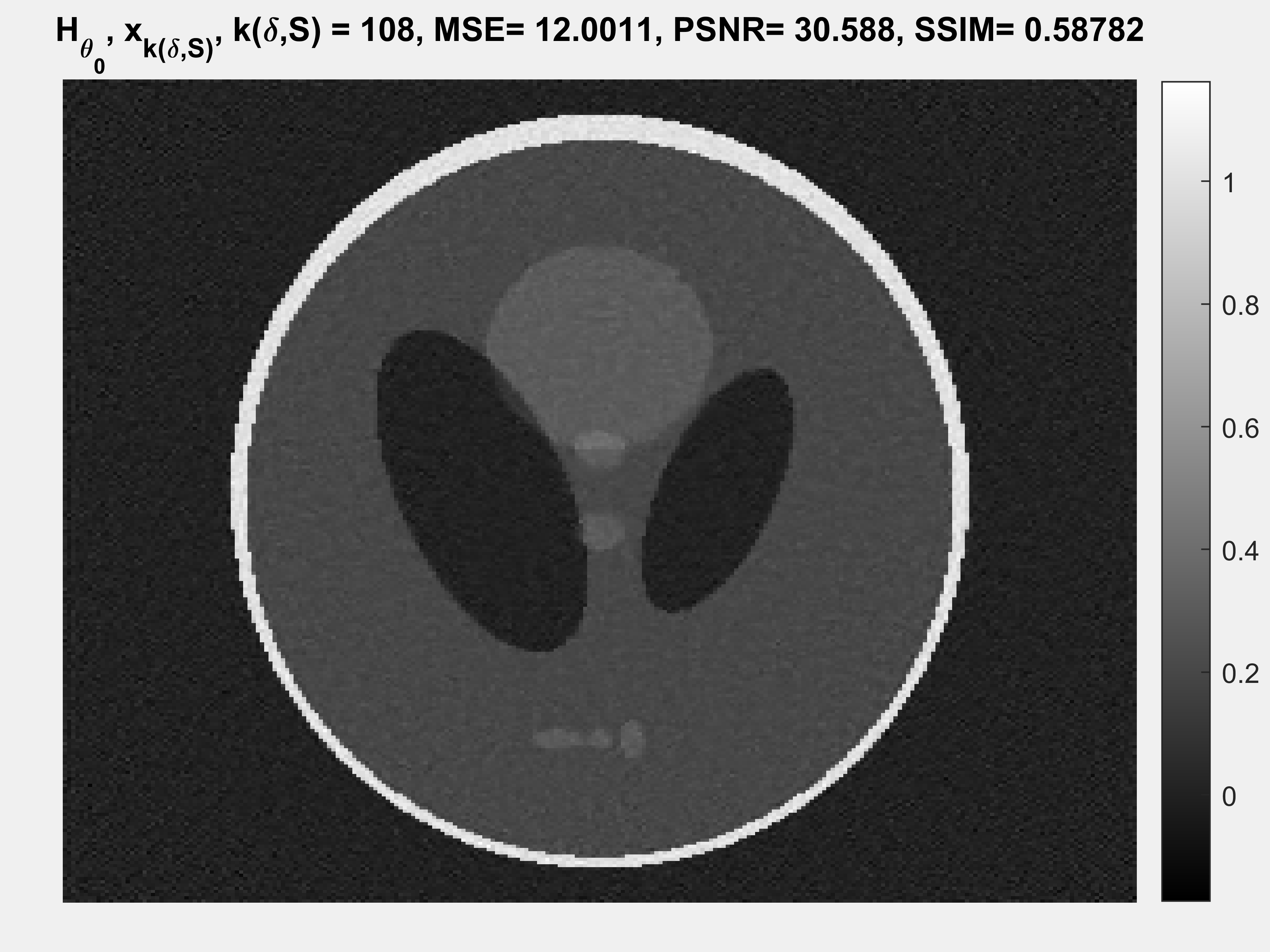}
        \caption{$x_{k(\delta,\mcal{S}_0)}$, CGLS$(100,1,1,x_k^\delta; H_{\theta_0})$}
        \label{Fig ADMM-PnP DnCNN CGLS rho1 phi1}
    \end{subfigure}       
    \begin{subfigure}{0.495\textwidth}
        \includegraphics[width=\textwidth]{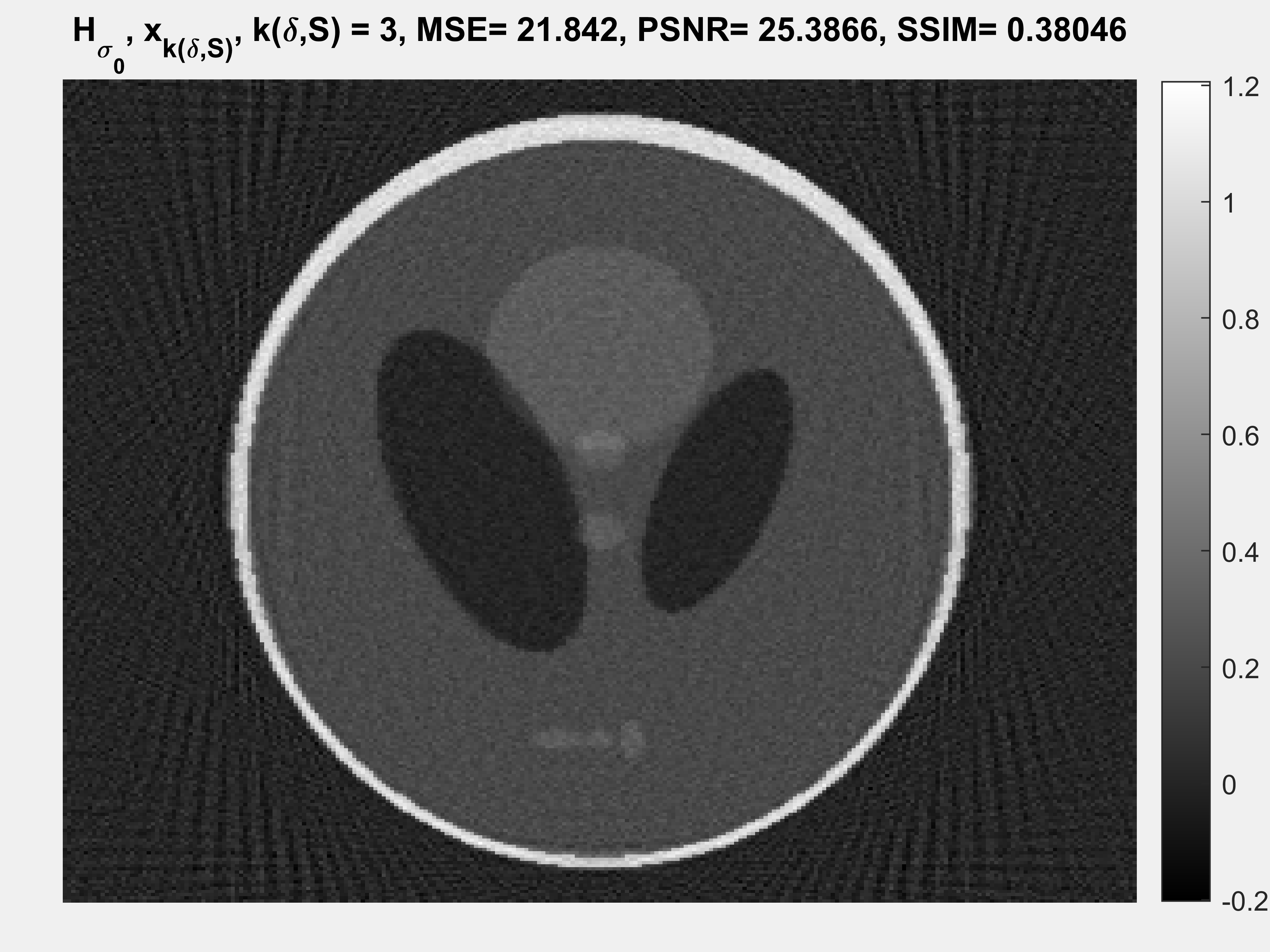}
        \caption{$x_{k(\delta,\mcal{S}_0)}$, CGLS$(100,1,1,x_k^\delta; H_{\sigma_0})$}
        \label{Fig ADMM-PnP BM3D CGLS rho1 phi1}
    \end{subfigure}       
    \begin{subfigure}{0.495\textwidth}
        \includegraphics[width=\textwidth]{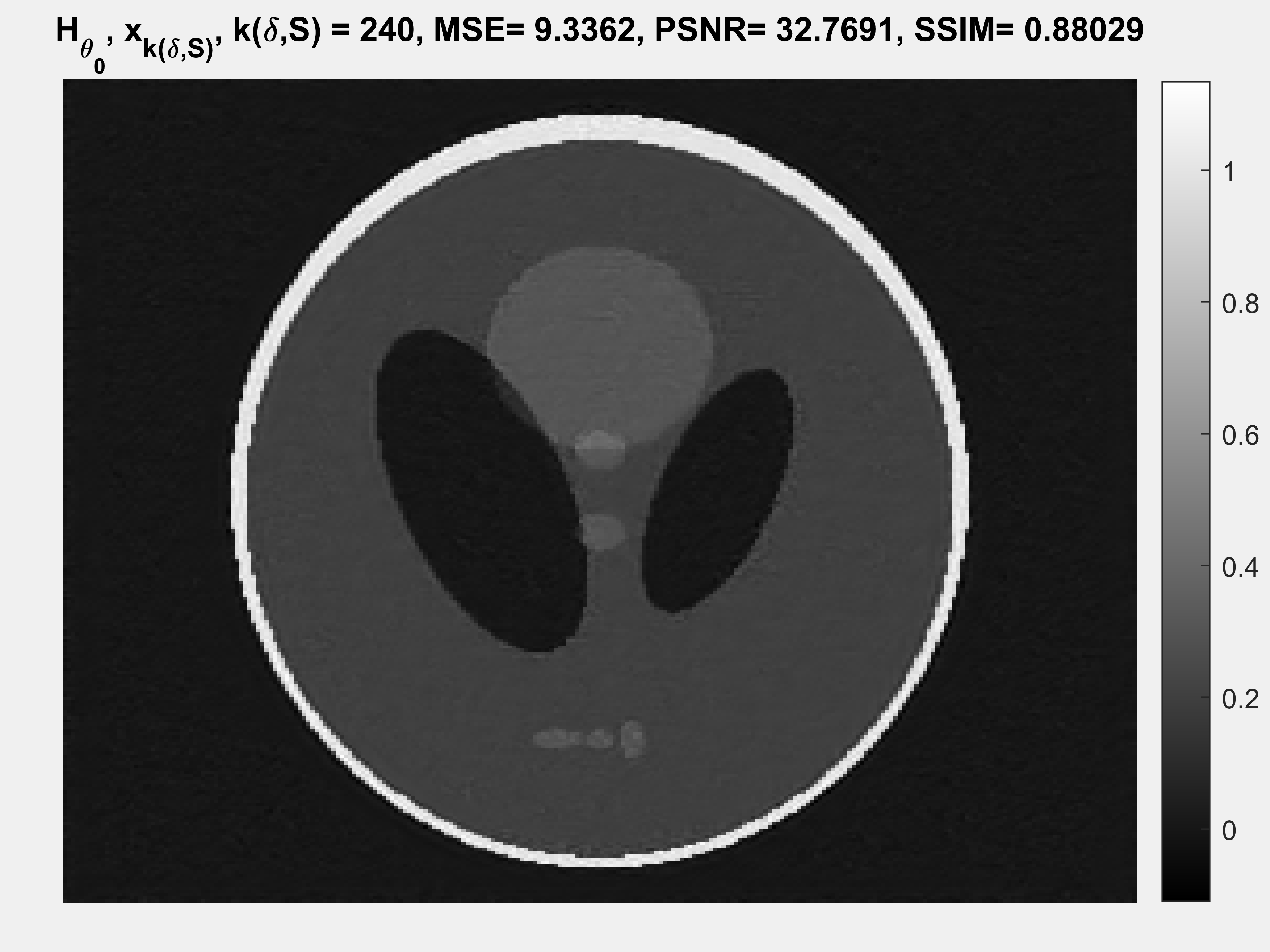}
        \caption{$x_{k(\delta,\mcal{S}_0)}$, CGLS$(10,100,1,x_k^\delta; H_{\theta_0})$}
    \end{subfigure} 
    \begin{subfigure}{0.495\textwidth}
        \includegraphics[width=\textwidth]{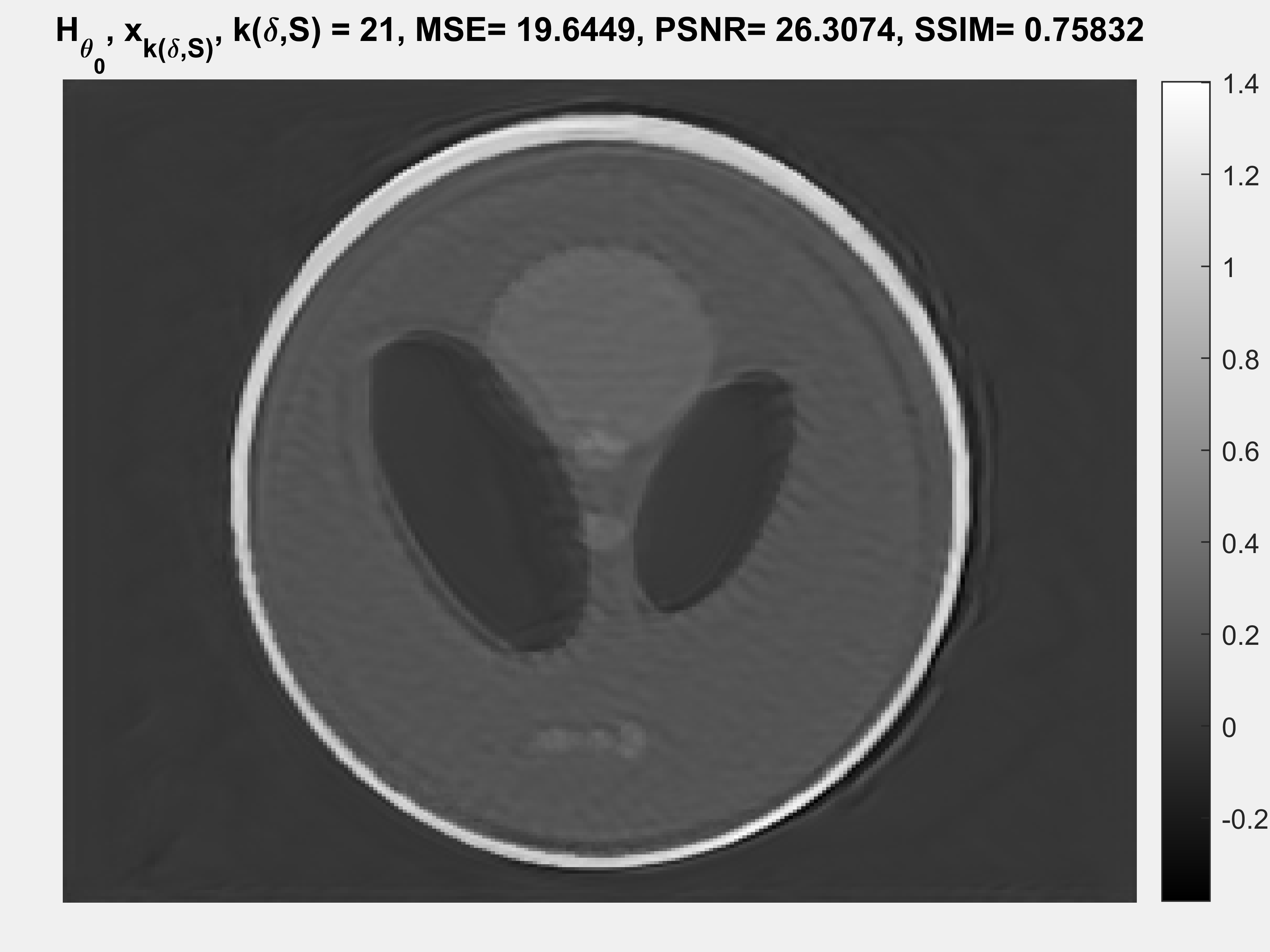}
        \caption{$x_{k(\delta,\mcal{S}_0)}$, GD$(10,10^{-5},10^{-5},y_k^\delta; H_{\theta_0})$}
    \end{subfigure}    
    \begin{subfigure}{0.495\textwidth}
        \includegraphics[width=\textwidth]{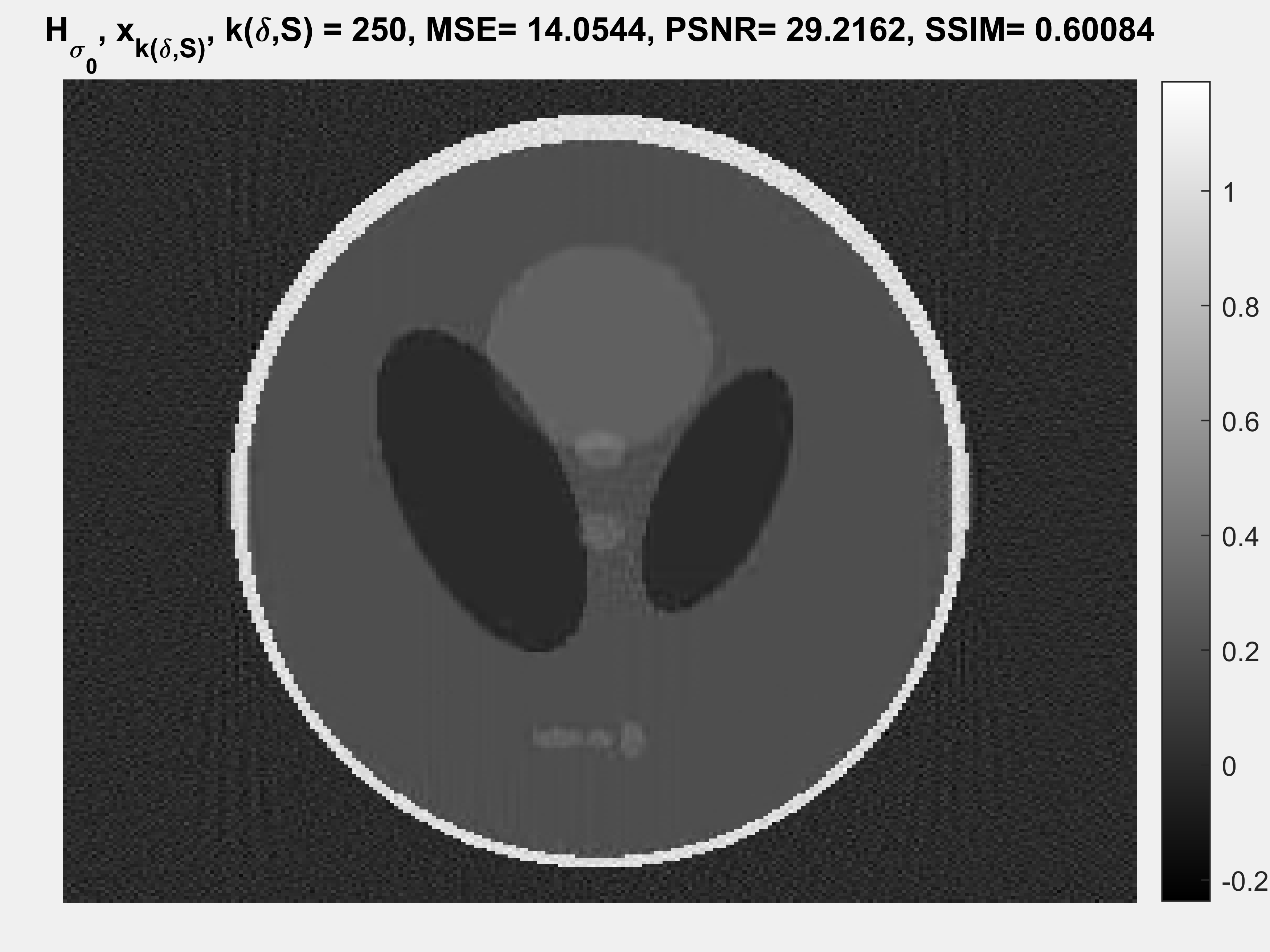}
        \caption{$x_{k(\delta,\mcal{S}_0)}$, GD$(10,10^{-5},10^{-5},y_k^\delta; H_{\sigma_0})$}
    \end{subfigure}
    \begin{subfigure}{0.495\textwidth}
        \includegraphics[width=\textwidth]{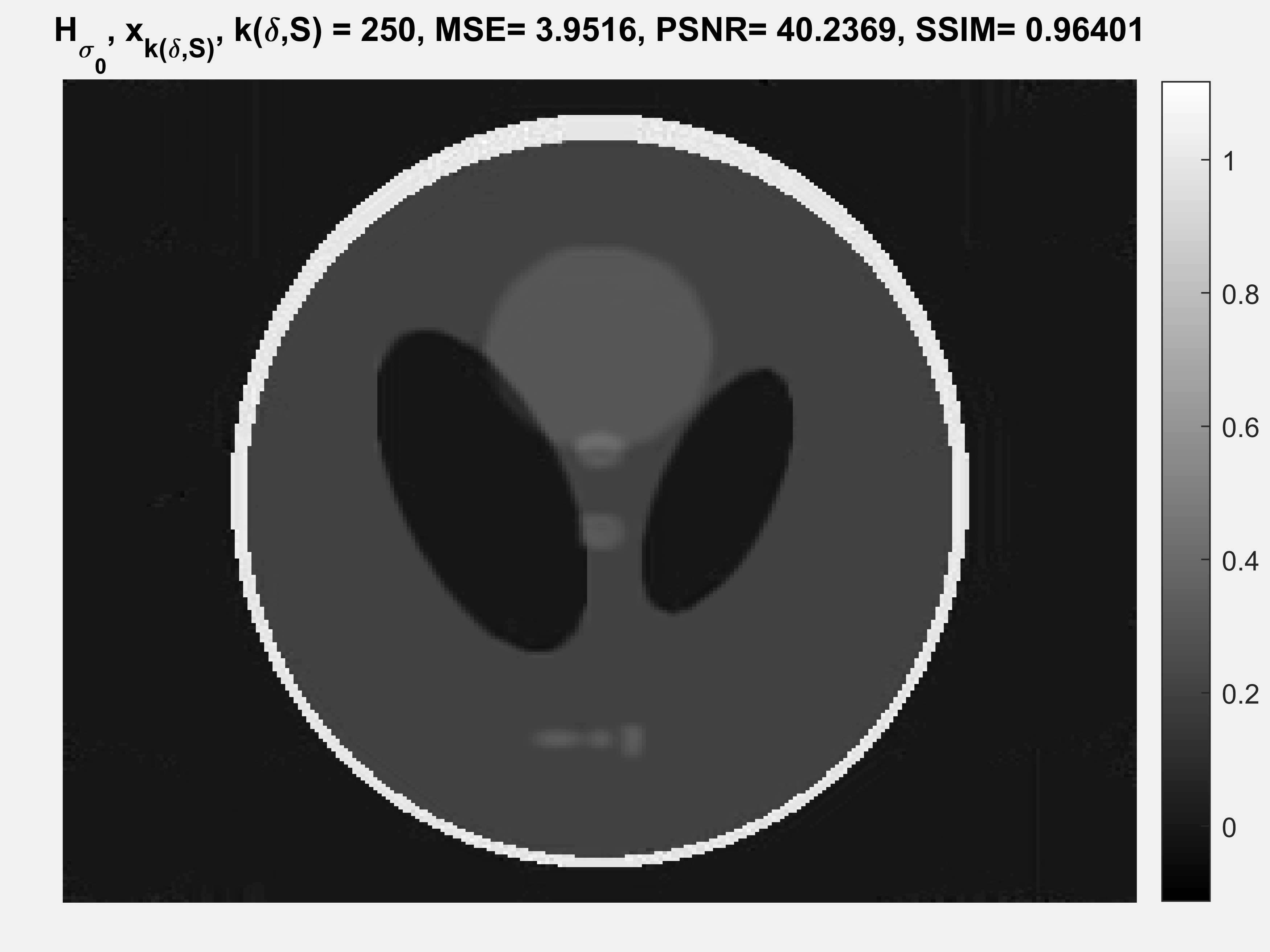}
        \caption{$x_{k(\delta,\mcal{S}_0)}$, GD$(1,10^{-5},10^{-5},y_k^\delta; H_{\sigma_0})$}
    \end{subfigure}
    \begin{subfigure}{0.495\textwidth}
        \includegraphics[width=\textwidth]{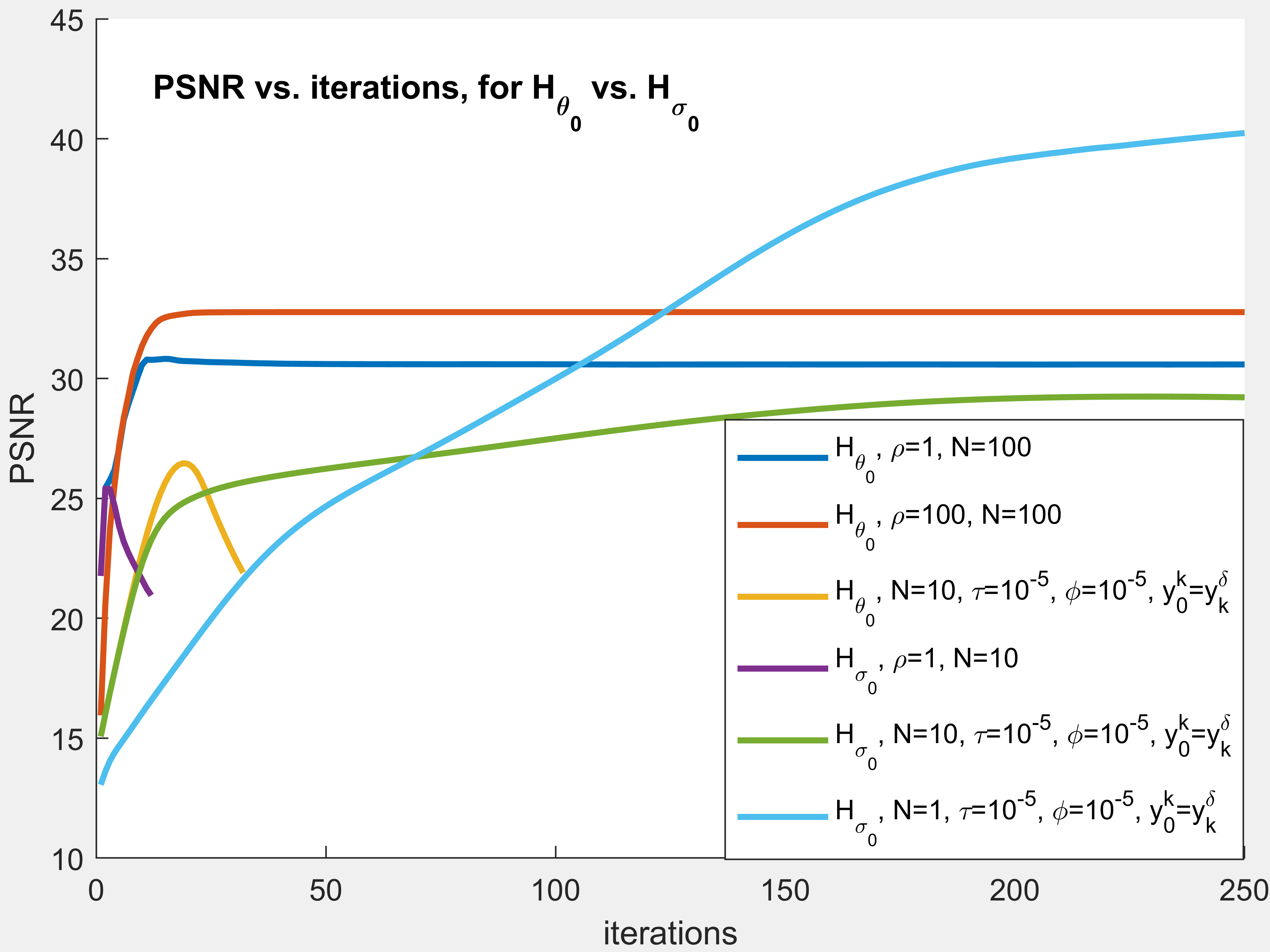}
        \caption{PSNR curves, $H_{\theta_0}(\alpha)$ vs. $H_{\sigma_0}$}
        \label{Fig ADMM-PnP PSNR}
    \end{subfigure}
    \begin{subfigure}{0.495\textwidth}
        \includegraphics[width=\textwidth]{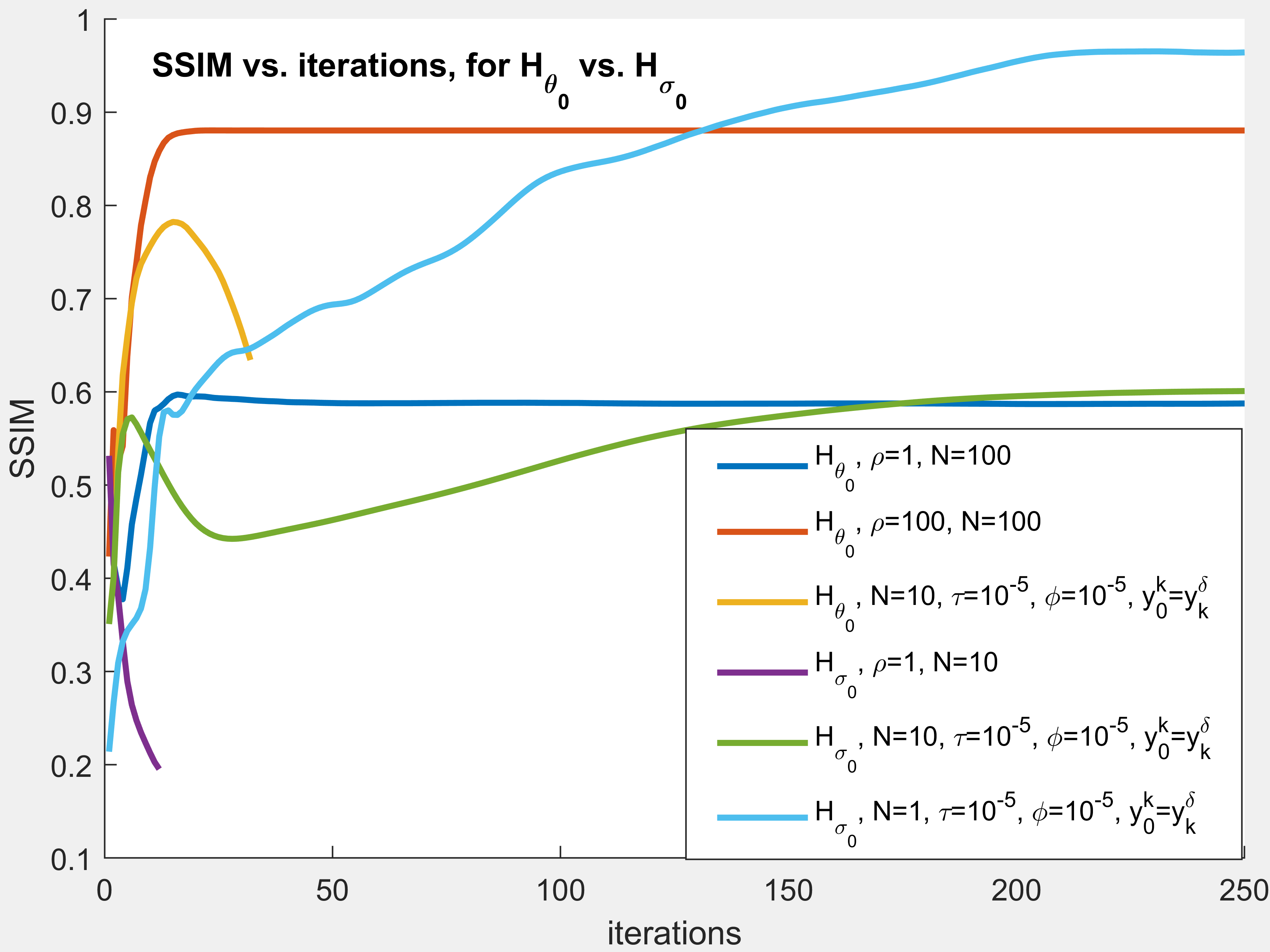}
        \caption{SSIM curves, $H_{\theta_0}(\alpha)$ vs. $H_{\sigma_0}$}
        \label{Fig ADMM-PnP SSIM}
    \end{subfigure}    
    \caption{ADMM-PnP: denoiser $H_{\theta_0}(\alpha)$ vs. $H_{\sigma_0}$, see Example \ref{Example ADMM-PnP}.} 
    \label{Figure ADMM-PnP}
\end{figure}

\begin{example}\label{Example ADMM-PnP}\textbf{[ADMM-PnP using $H_{\theta_0}$ vs. $H_{\sigma_0}$]}
\mbox{ }

In this example, we compare the recoveries obtained in the ADMM-PnP algorithm, when using $H_{\theta_0}$ vs. $H_{\sigma_0}$. Here, we show that, unlike the previous example, using the (strong) denoiser $H_{\theta_0}$ leads to a much better recovery than using the classical (weak) denoiser $H_{\sigma_0}$, naively. The reason being, in an ADMM-PnP algorithm the data-consistency step \eqref{ADMM-PnP data-cons.} can be large, for smaller $\rho$ values, resulting in $x_k^\delta$ having high noise intensities, and hence, can be appropriately denoised by a stronger denoiser $H_{\theta_0}$. 
In contrast, here, the weaker denoiser $H_{\sigma_0}$, for small $\sigma_0$, will yield a noisy reconstruction, since the data-denoising step is not strong enough to compensate the high noise levels arising in the iterates $x_k^\delta$ (from the large data-consistency step towards the noisy data $b_\delta$), resulting in under-denoised iterates $z_k^\delta$. Now, similar to attenuating a strong denoiser (via parameterizing it with an external attenuating parameter $\alpha$), one can attempt to boost or augment the denoising strength of a weaker denoiser (via some form of parameterization) but, this is relatively much harder than the previous scenario, for reasons explained below.

Again, we keep the experimental settings of Example \ref{Example FBS-PnP} unchanged, except, we implement the ADMM-PnP algorithm, instead of the FBS-PnP algorithm, using the learned denoiser $H_{\theta_0}$ and the classical denoiser $H_{\sigma_0}$. The numerical values of the results are shown in Table \ref{Table ADMM-PnP} and the figures in Figure \ref{Figure ADMM-PnP}. Note that, Figure \ref{Fig CGLS x_N} shows the recovered solution without any denoisers ($x_N^\delta$ after 250 CGLS iterations), Figure \ref{Fig ADMM-PnP DnCNN CGLS rho1 phi1} shows $x_{k(\delta,\mcal{S}_0)}^\delta$, for $k(\delta,\mcal{S}_0) = 108$, using the learned denoiser $H_{\theta_0}$ and Figure \ref{Fig ADMM-PnP BM3D CGLS rho1 phi1} shows $x_{k(\delta,\mcal{S}_0)}^\delta$, for $k(\delta,\mcal{S}_0) = 3$, using the classical denoiser $H_{\sigma_0}$, where for inner optimization problem \eqref{ADMM-PnP data-cons.} we consider $\rho=1$ and used 100 CGLS-iterations (with $x_k^\delta$ as the starting point). Here, one can see the improvements in the recovered solution using $H_{\theta_0}$ over $H_{\sigma_0}$, for reasons explained above. 
\end{example}

\subsection{Boosting a weak denoiser in ADMM-PnP}
\mbox{ }

\begin{figure}[h!]
    \centering
    \begin{subfigure}{0.495\textwidth}
        \includegraphics[width=\textwidth]
        {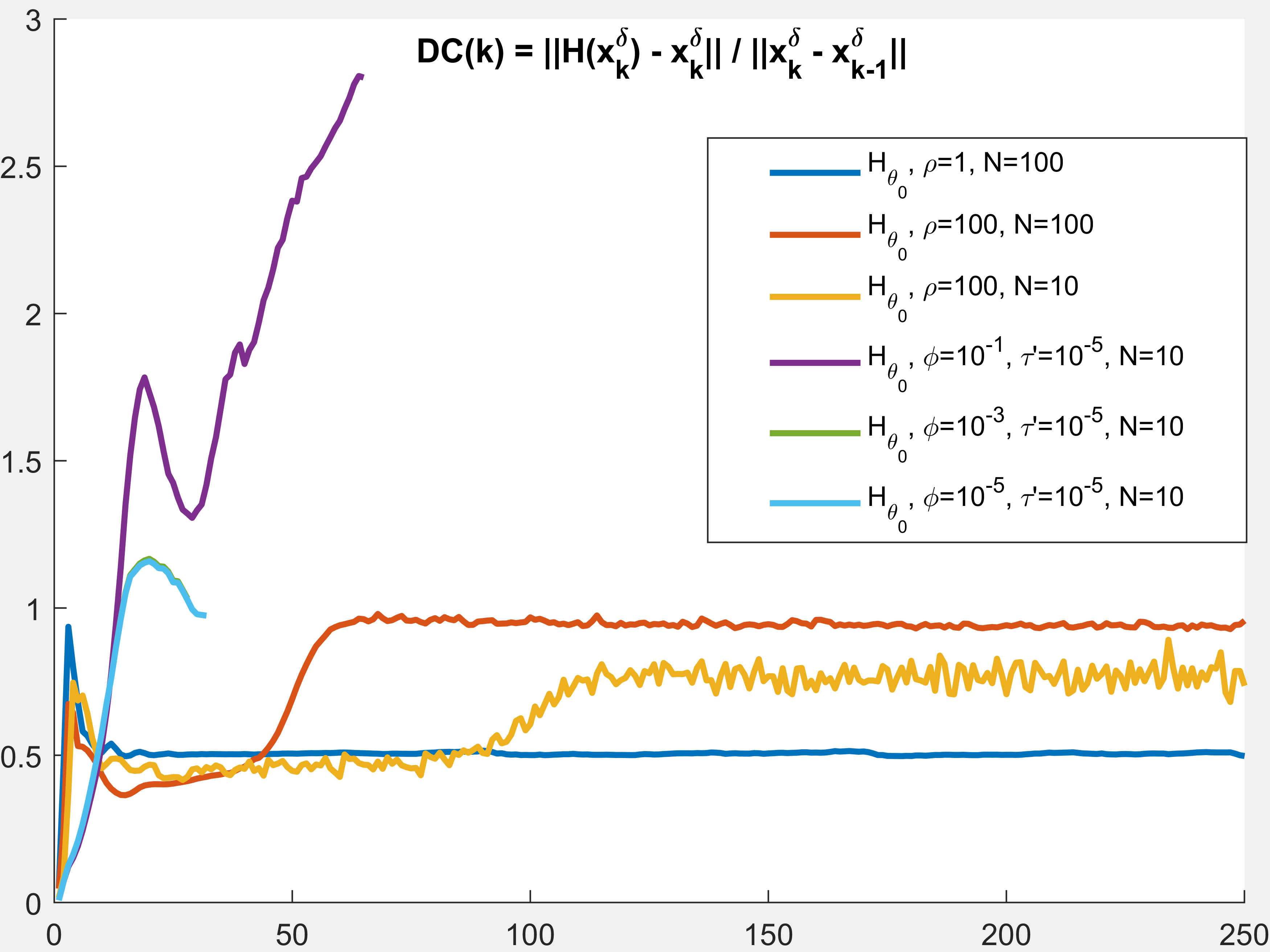}
        \caption{$DC(k)$ for $H_{\theta_0}$ in ADMM-PnP}
    \end{subfigure}  
    \begin{subfigure}{0.495\textwidth}
        \includegraphics[width=\textwidth]
        {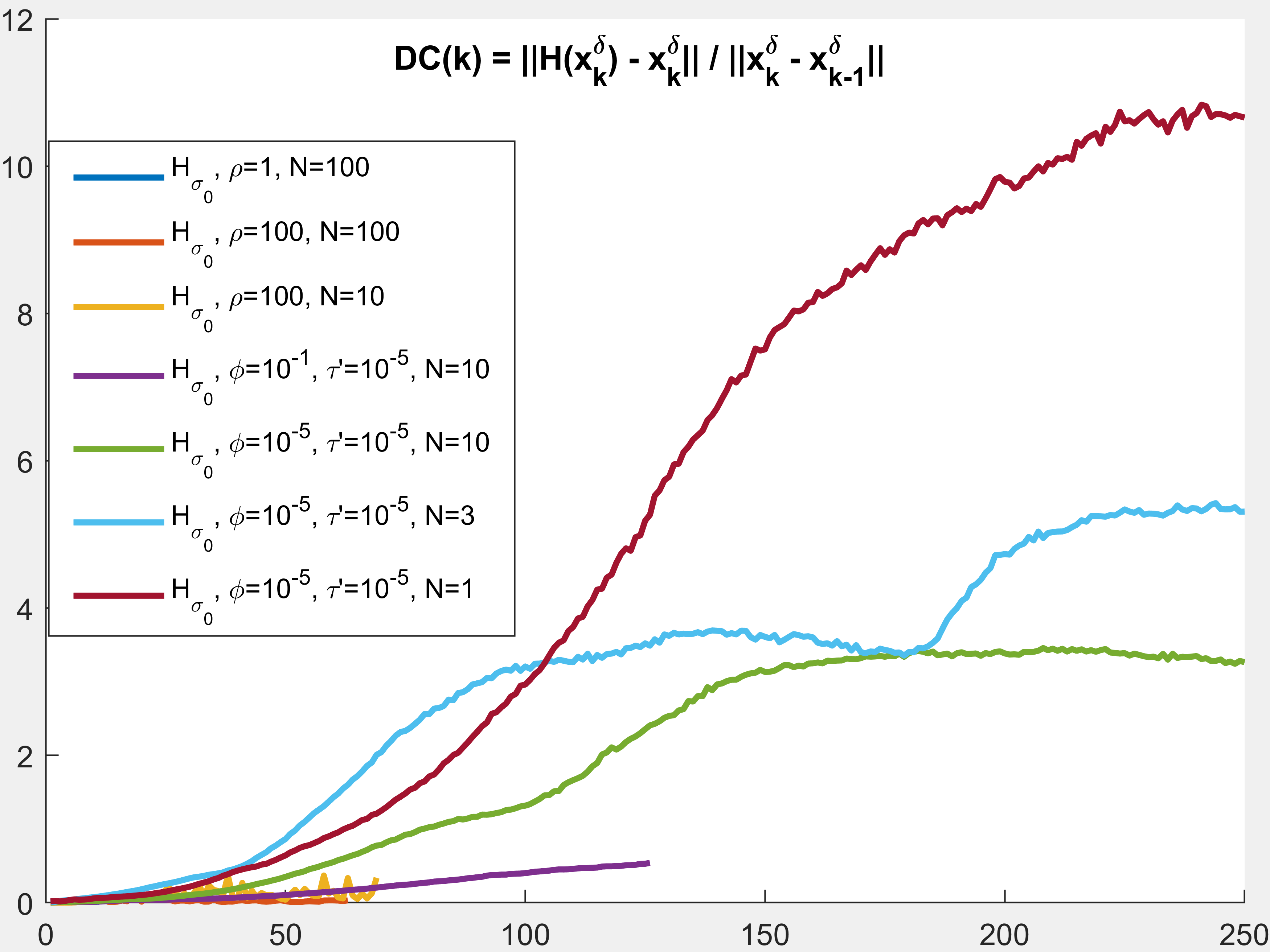}
        \caption{$DC(k)$ for $H_{\sigma_0}$ in ADMM-PnP}
    \end{subfigure}          
    \caption{$DC(k)$ for different $\mcal{OA}$ in ADMM-PnP, see Example \ref{Example ADMM-PnP}.} 
    \label{Figure denoising-to-consistency ADMM-PnP}
\end{figure}

In contrast to attenuating a strong denoiser, boosting a weaker denoiser is relatively much harder, since one can cannot simply parameterize a weak denoiser, naively, by any external parameter $\alpha > 1$, like in \eqref{Strong deno. alpha0}. In \cite{nayak2021PnP}, we provided a technique to boost a weaker denoiser when used in a FBS-PnP, as well as, ADMM-PnP settings. Here, we also provide certain insights to boost a weak denoiser in an ADMM-PnP setting, from a different angle. 
To have a better understanding of the reasons behind the boosting strategy, we would like to first dissect the net-change direction, at each step, in an ADMM-PnP algorithm. Note that, comparing the FBS-PnP algorithm to the ADMM-PnP algorithm, we have at every step $k$, fixing $u_{k-1}^\delta$,
\begin{align}
z_{k-1}^\delta \longmapsto \; x_k^\delta \; &= \; \argmin_{x} \; \mcal{D}(x) + \rho||x - (z_{k-1}^\delta - u_{k-1}^\delta)||_2^2, \; \longleftarrow \mbox{ data-consistency step} \label{ADMM-PnP data-consistency 1} \\
x_k^\delta \longmapsto \; z_k^\delta \; &= \; H_{\sigma_k}(x_k^\delta + u_{k-1}^\delta), \; \hspace{1cm} \longleftarrow \mbox{ data-denoising step}, \label{ADMM-PnP data-denoising 1}
\end{align}
and hence, the resulting direction, from $z_{k-1}^\delta$ to $z_k^\delta$, is given by
\begin{align}\label{ADMM-PnP direction 1}
&d_{k-1}^\delta := z_k^\delta - z_{k-1}^\delta \\ 
&= \underbrace{\left[ \argmin_{x} \; \mcal{D}(x) + \rho||x - (z_{k-1}^\delta - u_{k-1}^\delta)||_2^2 \right] - z_{k-1}^\delta}_{data-consistency} \; + \; \underbrace{\left[H_{\sigma_k}(x_k^\delta + u_{k-1}^\delta) - (x_k^\delta + u_{k-1}^\delta) \right]}_{data-denoising}. \; \notag
\end{align}
And, since one only estimates the minimizer of \eqref{ADMM-PnP data-consistency 1}, through certain number of iterative optimization steps, the resulting direction is in fact dependent on the optimization architecture ($\mcal{OA}$), which includes the number of iterations ($N$), the initial iterates ($y_0^k$), the  step-sizes $\tau_k'$, as well as, the error-iterate $u_{k-1}^\delta$ i.e.,
\begin{equation}\label{ADMM-PnP direction OA}
d_{k-1}^\delta = d_{k-1}^\delta\left( z_{k-1}^\delta, \; x_k^\delta(\mcal{OA}(N,y_0^k,\tau_k')), \; z_k^\delta(H_{\sigma_k}), \; u_{k-1}^\delta \right),
\end{equation}
in comparison, the resulting direction in a FBS-PnP algorithm is simply dependent on the the step-size $\tau_k$ and the gradient $\nabla_x \mcal{D}(z_{k-1}^\delta)$, i.e.,
\begin{equation}\label{FBS-PnP direction OA}
d_{k-1}^\delta = d_{k-1}^\delta\left( z_{k-1}^\delta, \; x_k^\delta(\tau_k, \nabla_x \mcal{D}(z_{k-1}^\delta)), \; z_k^\delta(H_{\sigma_k}) \right).
\end{equation}
Thus, one can see that for $u_{k-1}^\delta \equiv 0$, for all k, and $\mcal{OA}$ corresponding to a single step ($N=1$) of the descent direction, starting from $y_0^k = z_{k-1}^\delta$ with $\tau_k' = \tau_k$, i.e., $x_k^\delta(\mcal{OA}(N,y_0^k,\tau_k')) = z_{k-1}^\delta - \tau_k \nabla_x \mcal{D}(z_{k-1}^\delta)$, we retrieve back the FBS-PnP algorithm. Hence, for $H_{\sigma_k} = H_{\sigma_0}$, for all k, and appropriately modifying the initial iterates ($y_0^k$) corresponding to Fast FBS-PnP algorithm, i.e., with a momentum step \eqref{ADMM-PnP y_0^k momentum step}, we can significantly improve over the previously recovered ADMM-PnP solution. But then, one could have simply stuck with the Fast FBS-PnP algorithm, as we are not changing anything. In other words, we would like to investigate if there are other paths (descent flows) that can provide better results. Note that, the struggle faced by the weak denoiser $H_{\sigma_0}$, in this case, is that, it has to denoise $x_k^\delta + u_{k-1}^\delta$, where $u_{k-1}^\delta$ is the error term as defined in \eqref{ADMM-PnP noise upd.}, and hence, unable to produce an effective denoised iterate $z_k^\delta = H_{\sigma_0}(x_k^\delta + u_{k-1}^\delta)$. Now, instead of forcing $u_k^\delta \equiv 0$, for all k, we can have a scaled error update, given by
\begin{equation}
        u_{k+1}^\delta = u_k^\delta + \phi(x_{k+1}^\delta - z_{k+1}^\delta), \;\; \longleftarrow \text{ scaled noise update step}
        \label{ADMM-PnP noise scaled upd.}
\end{equation}
for $0 \leq \phi \leq 1$, which can also lead to efficient recoveries, see Table \ref{Table ADMM-PnP} and Figure \ref{Figure ADMM-PnP}. Therefore, one can observe that, the recovery in an ADMM-PnP algorithm (in fact, for any iterative regularization scheme, see \cite{nayak2021PnP}) is heavily dependent on the flow of the evolving iterations, even to an extent that, minimizing the expression in \eqref{ADMM-PnP data-consistency 1} with a different formulation:
\begin{equation}\label{ADMM-PnP data-consistency 2}
x_k^\delta = \argmin_{x} \; \frac{1}{\rho} \; \mcal{D}(x) + ||x - (z_{k-1}^\delta - u_{k-1}^\delta)||_2^2,
\end{equation}
will also yield a different solution, as we are not completely minimizing them. For example, compare the results in Table \ref{Table ADMM-PnP} for GD$(N,\phi,\tau',y_0^k)$, which solves \eqref{ADMM-PnP data-consistency 2} for 10 iterations, vs. CGLS$(N,\phi,\rho,y_0^k)$, which solves \eqref{ADMM-PnP data-consistency 1} for 10 iterations, for the same $\rho$ value. Some of the results corresponding to different $\mcal{OA}$ is presented in Table \ref{Table ADMM-PnP}, where CGLS$(N,\phi,\rho,y_0^k)$ stands for conjugate-gradient least-squares method with $N$ iterations, $\phi$ ($u_k^\delta$ scaling parameter, as in \eqref{ADMM-PnP noise scaled upd.}), $\rho$ (Lagrangian parameter) and $y_0^k$ is the initial point for the inner optimization process, which can be $x_k^\delta$, $z_k^\delta$ or $y_k^\delta$, where $y_k^\delta$ is a momentum step given by, $y_0^\delta = z_0^\delta$ and for $k \geq 1$
\begin{align} \label{ADMM-PnP y_0^k momentum step}
y_k^\delta &\longmapsto \; y_k^\delta \; = \; y_k^\delta \; + \; \alpha_k (y_k^\delta - y_{k-1}^\delta), \; &\longleftarrow \mbox{ momentum-step}
\end{align}
with $\alpha_k = \frac{t_{k-1} - 1}{t_k}$ and $t_k = \frac{ (1 + \sqrt{1 + 4t_{k-1}^2})}{2}$; and GD$(N,\phi, \tau', y_0^k)$ stands for the simple gradient descent method with a constant step-size ($\tau'$) and the gradient is defined as $ -\tau' \nabla_x \mcal{D}(x) - (x + (z_{k-1}^\delta - u_{k-1}^\delta))$. We can see, from Table \ref{Table ADMM-PnP}, that when using GD$(N,\phi, \tau',y_0^k)$, for $\phi = 10^{-5}$, $\tau' = 10^{-5}$, $y_0^k=y_k^\delta$ and $N=1$, we recover the best solution, even surpassing the Fast FBS-PnP solution. In contrast, the ADMM-PnP performance using $H_{\theta_0}$ is degrading with smaller $\phi$ values, since we are moving closer the FBS-PnP algorithm. Furthermore, even the iterative flow corresponding to GD$(1,10^{-5},10^{-5},y_k^\delta)$, for which we got the best result, may not be the best solution flow, that is, one might even recover better solutions through different $\mcal{OA}$, $\rho$ or $\phi$ values. Now, one may question the well-defineness of the recovered solution, since based on the same denoiser we are recovering wildly different solutions, where the answer to this question is explained in \cite{nayak2021PnP}. Note that, similar to Example \ref{Example FBS-PnP}, we can also plot the denoising-to-consistency ratio $DC(k)$ over the iterations, for the different denoisers and their $\mcal{OA}$, $\rho$ and $\phi$ values, some of which are shown in Figure \ref{Figure denoising-to-consistency ADMM-PnP}. One can see that, the $\mcal{OA}$, $\rho$ and $\phi$ values for which $DC(k)$ is small, yields a noisy solution, where as, the $\mcal{OA}$, $\rho$ and $\phi$ values associated with large $DC(k)$, results in a well-denoised reconstruction.

\begin{figure}[h!]
    \centering
    \begin{subfigure}{0.495\textwidth}
        \includegraphics[width=\textwidth]
        {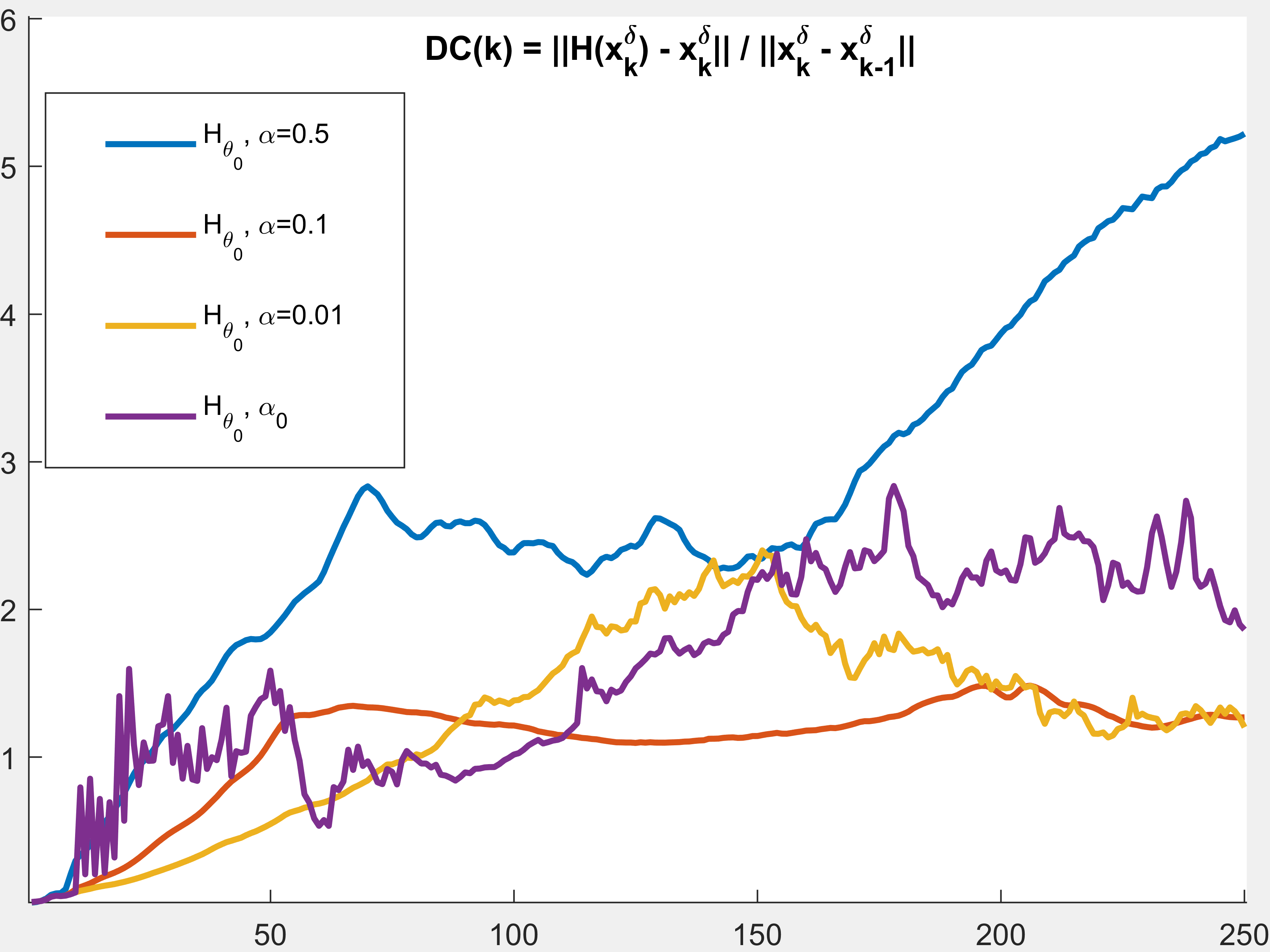}
        \caption{$DC(k)$ for $H_{\alpha\theta_0+\beta H_{\sigma_0}}$ in FBS-PnP}
    \end{subfigure}  
    \begin{subfigure}{0.495\textwidth}
        \includegraphics[width=\textwidth]
        {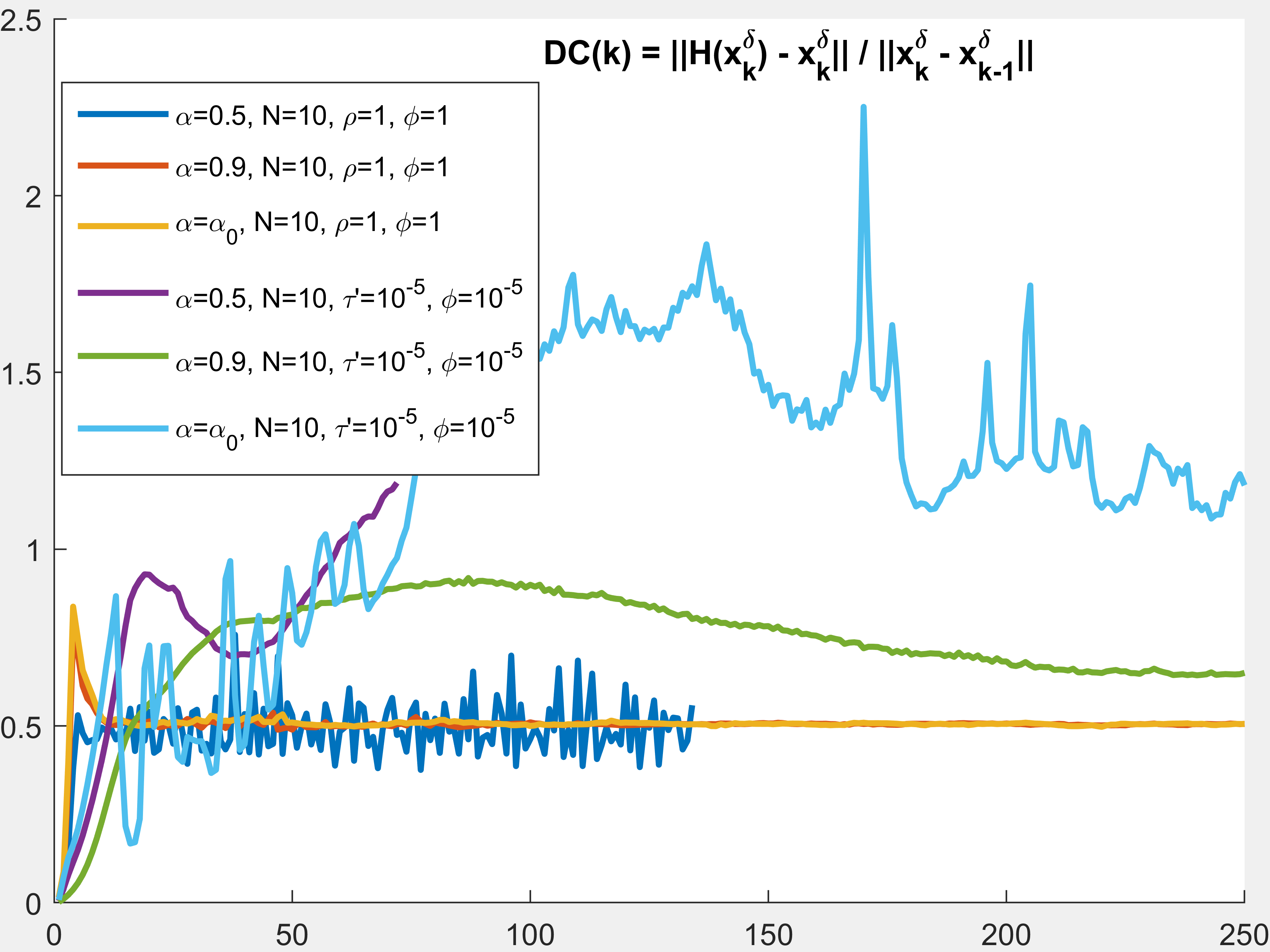}
        \caption{$DC(k)$ for $H_{\alpha\theta_0+\beta H_{\sigma_0}}$ in ADMM-PnP}
    \end{subfigure}           
    \caption{$DC(k)$ for $H_{\alpha\theta_0+\beta H_{\sigma_0}}$, such that $\alpha + \beta = 1$, in FBS-PnP vs. ADMM-PnP (with different $\mcal{OA}$ and $\phi$ values), see Example \ref{Example FBS-PnP BM3D-DnCNN} and Example \ref{Example ADMM-PnP DM3D+DnCNN}.} 
    \label{Figure denoising-to-consistency FBS-PnP + ADMM-PnP}
\end{figure}

\begin{figure}[h!]
    \centering
    \begin{subfigure}{0.495\textwidth}
        \includegraphics[width=\textwidth]{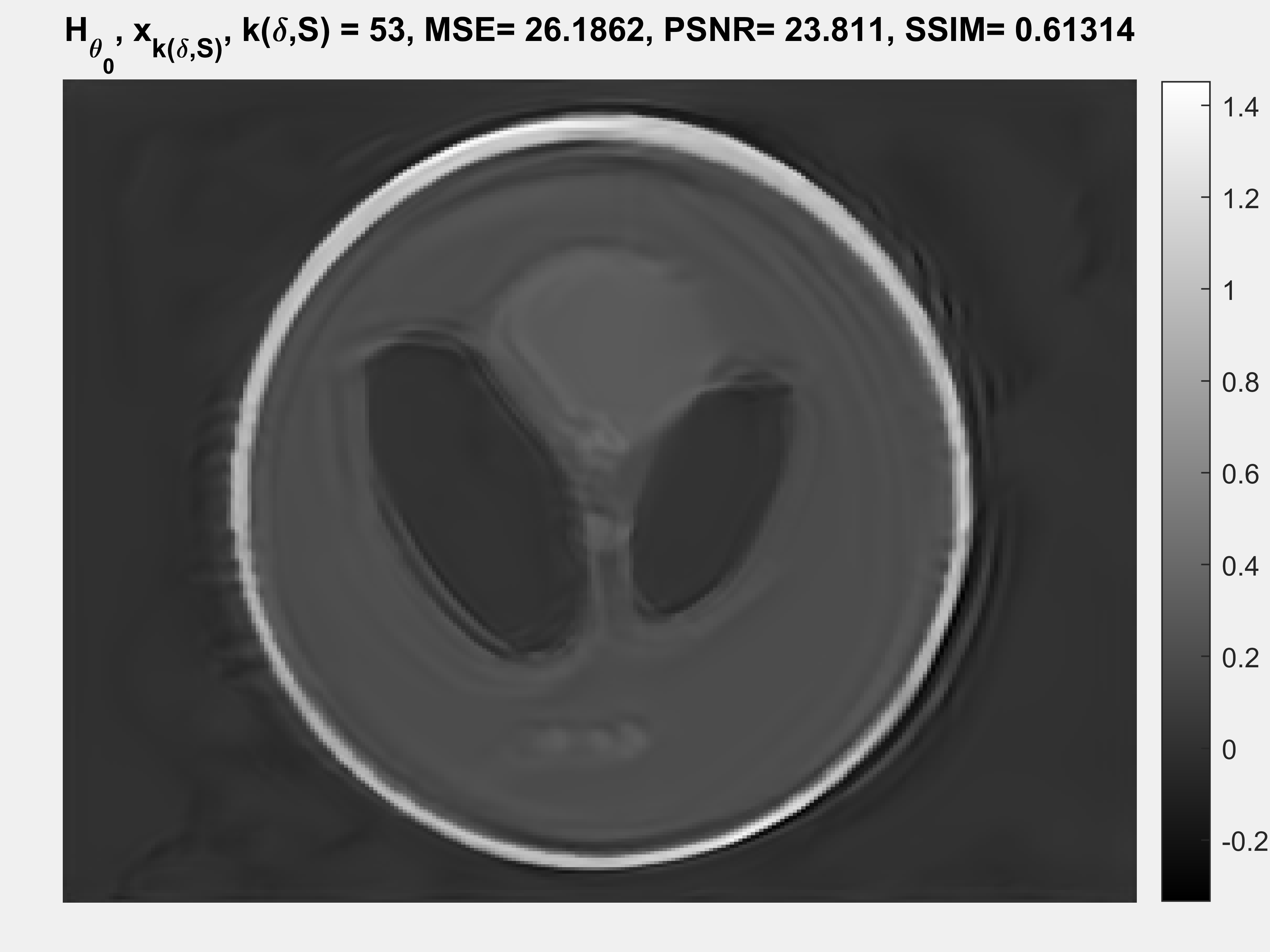}
        \caption{FBS-PnP, $\alpha=\frac{1}{2}$}
    \end{subfigure}       
    \begin{subfigure}{0.495\textwidth}
        \includegraphics[width=\textwidth]{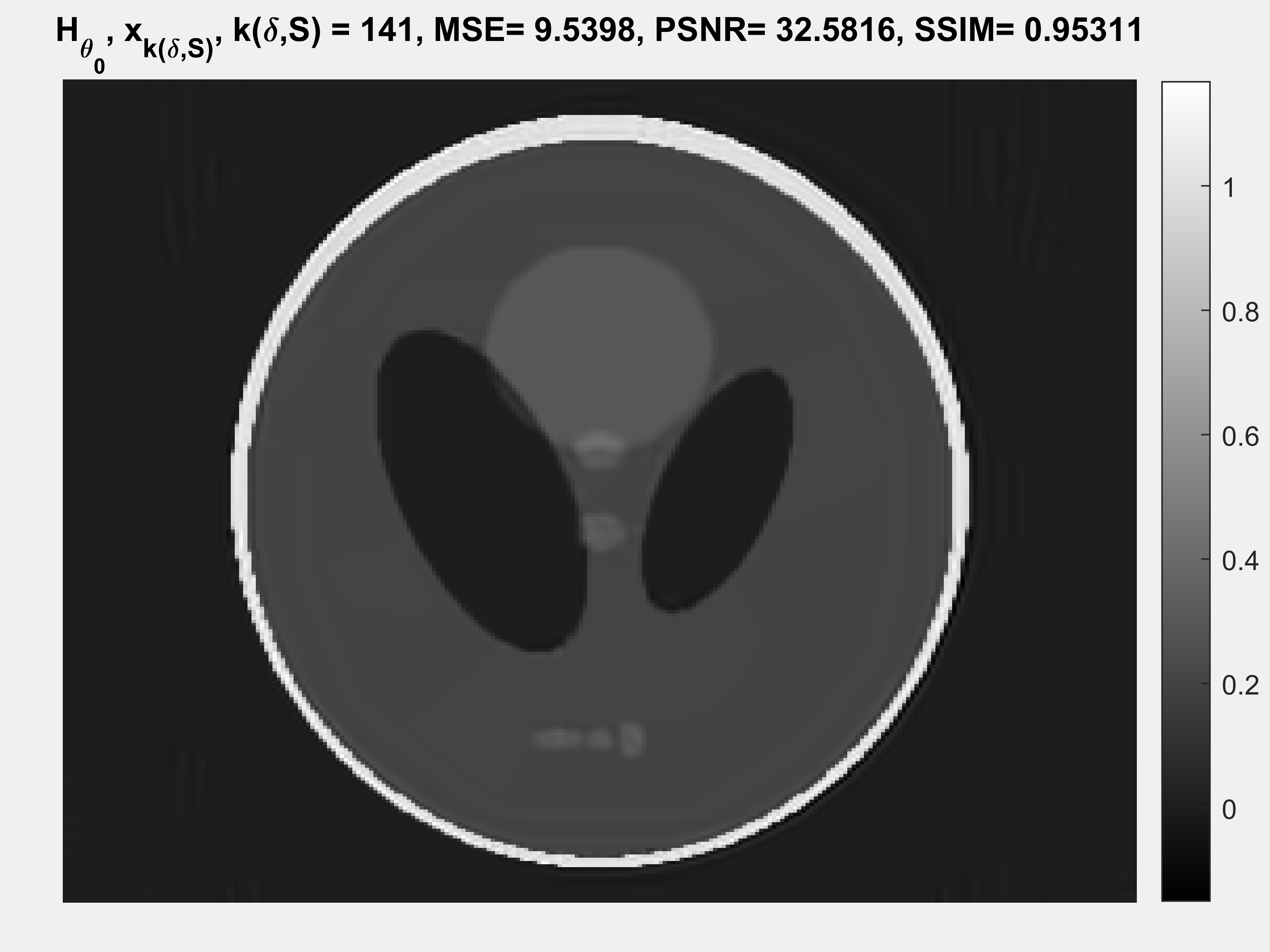}
        \caption{FBS-PnP, $\alpha = 0.01$}
    \end{subfigure}       
    \begin{subfigure}{0.495\textwidth}
        \includegraphics[width=\textwidth]{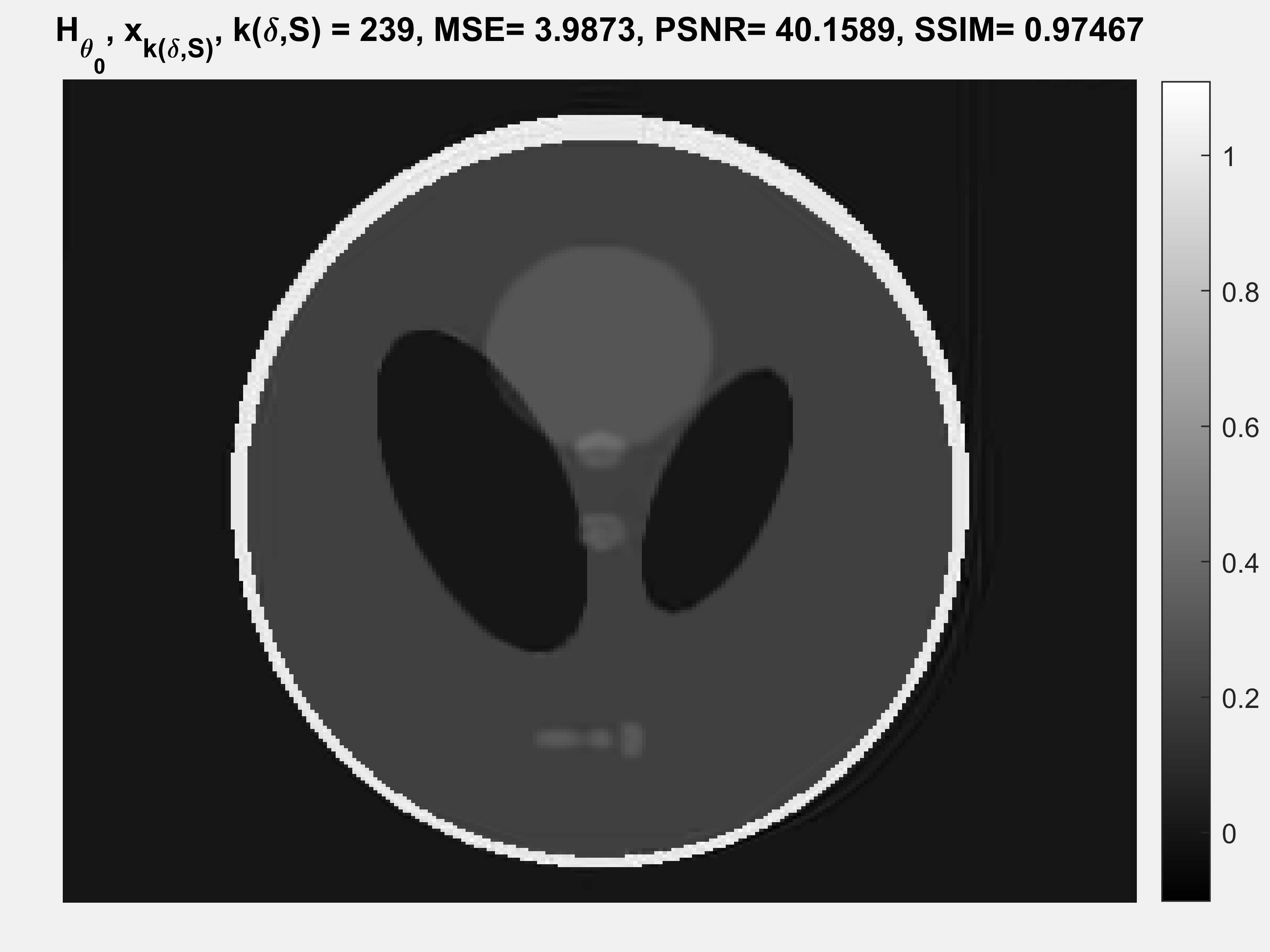}
        \caption{FBS-PnP, $\alpha=\alpha_0$}
    \end{subfigure} 
    \begin{subfigure}{0.495\textwidth}
        \includegraphics[width=\textwidth]{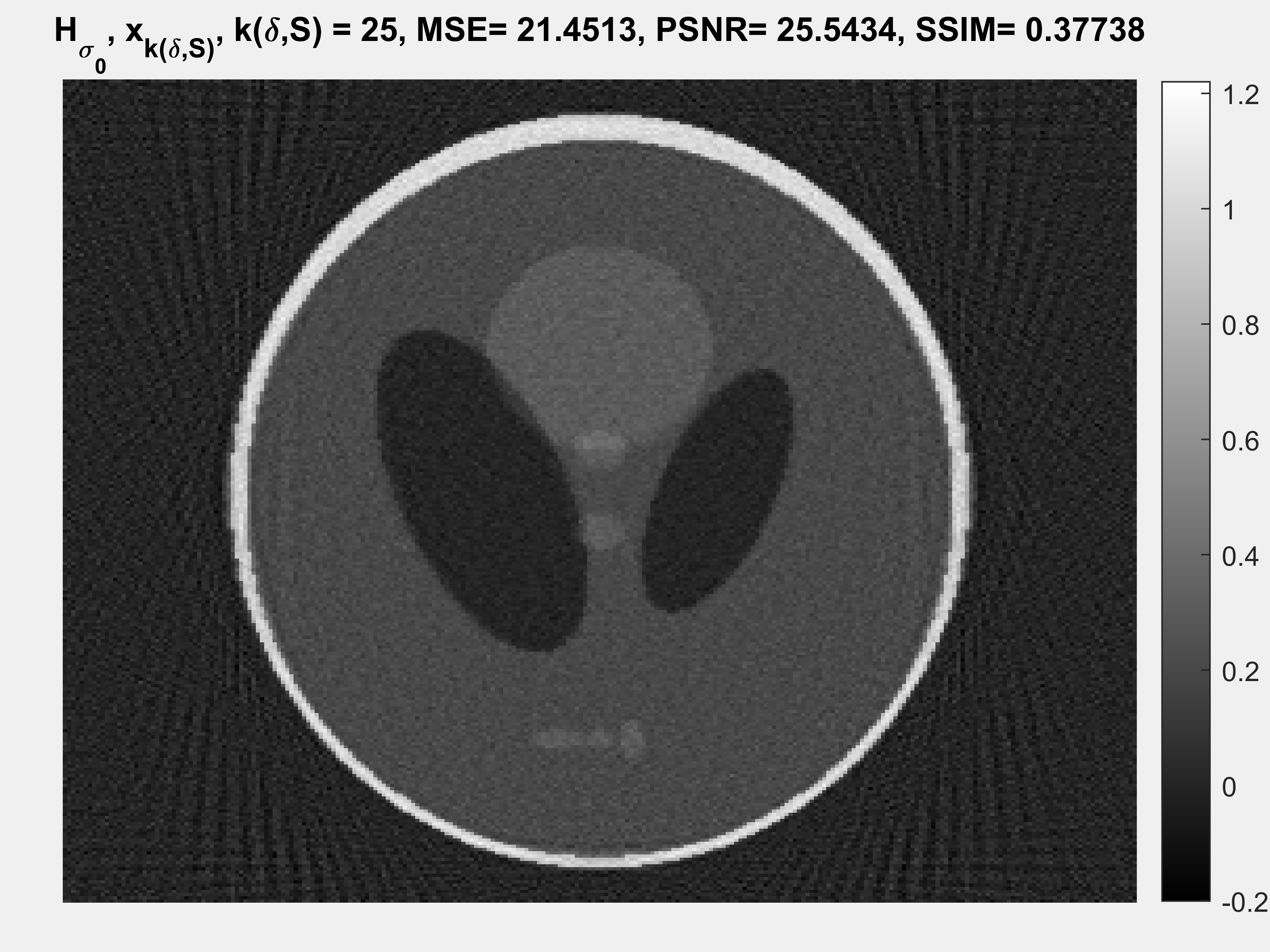}
        \caption{ADMM-PnP, $\alpha=\frac{1}{2}$, CGLS$(10,1,1,x_k^\delta)$}
    \end{subfigure}    
    \begin{subfigure}{0.495\textwidth}
        \includegraphics[width=\textwidth]{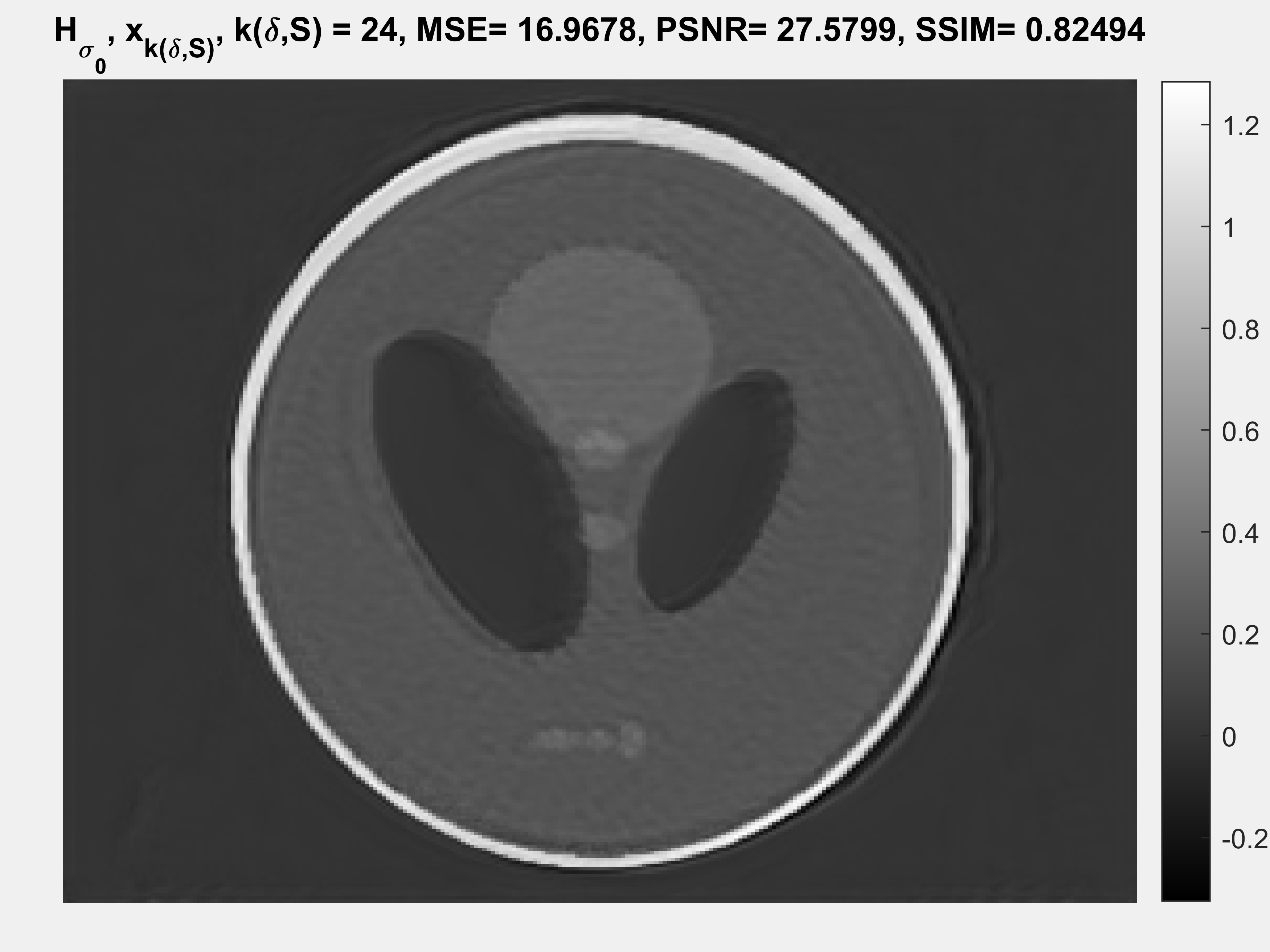}
        \caption{ADMM-PnP, $\alpha$ = $\frac{1}{2}$, GD$(10,10^{-5},10^{-5},y_0^k)$}
    \end{subfigure}
    \begin{subfigure}{0.495\textwidth}
        \includegraphics[width=\textwidth]{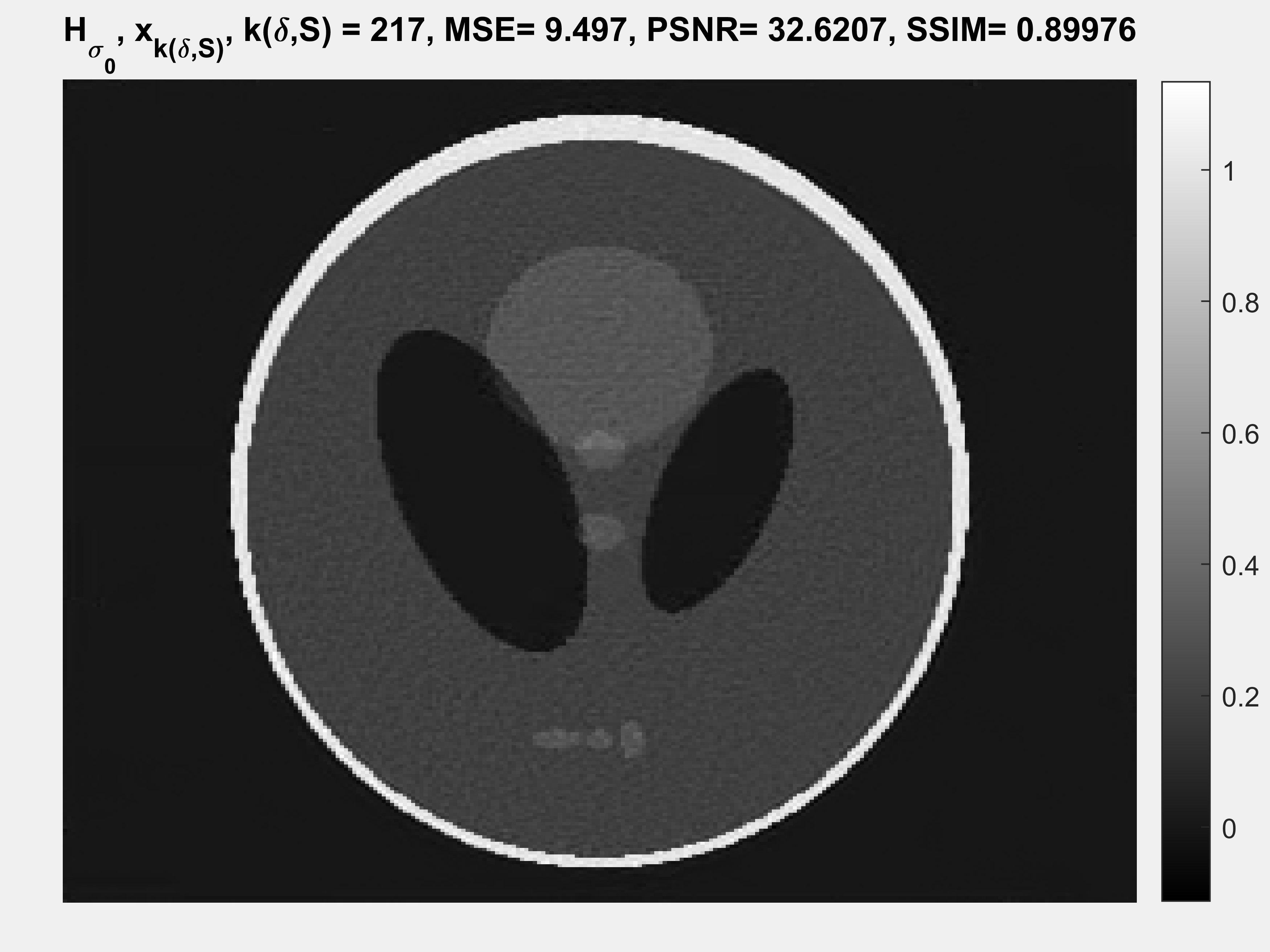}
        \caption{ADMM-PnP, $\alpha$ = $\alpha_0$, GD$(10,10^{-5},10^{-5},y_0^k)$}
    \end{subfigure}
    \begin{subfigure}{0.495\textwidth}
        \includegraphics[width=\textwidth]{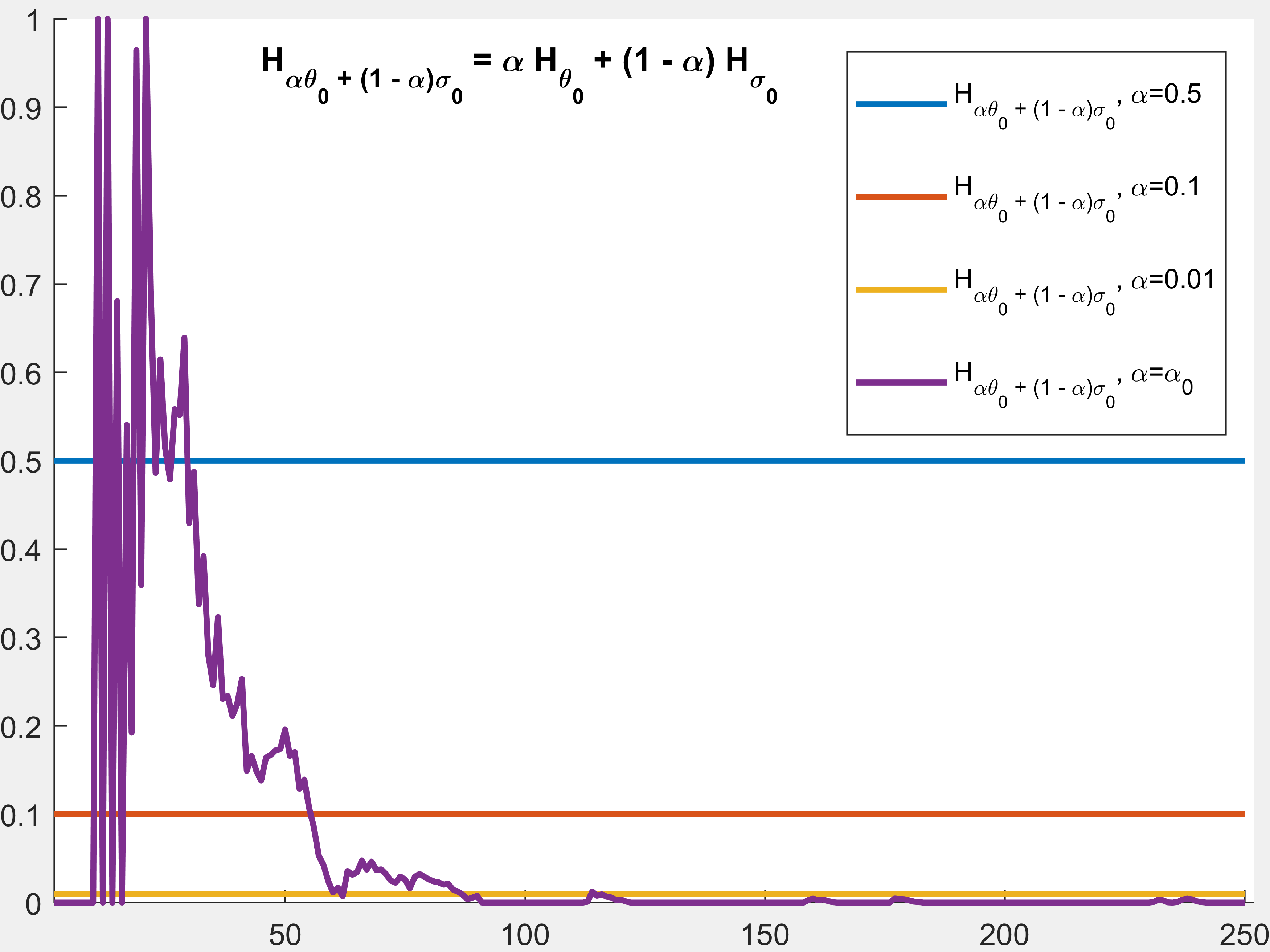}
        \caption{FBS-PnP, $\alpha_0(k)$ vs. $k$}
        \label{Fig FBS-PnP weights alpha}
    \end{subfigure}
    \begin{subfigure}{0.495\textwidth}
        \includegraphics[width=\textwidth]{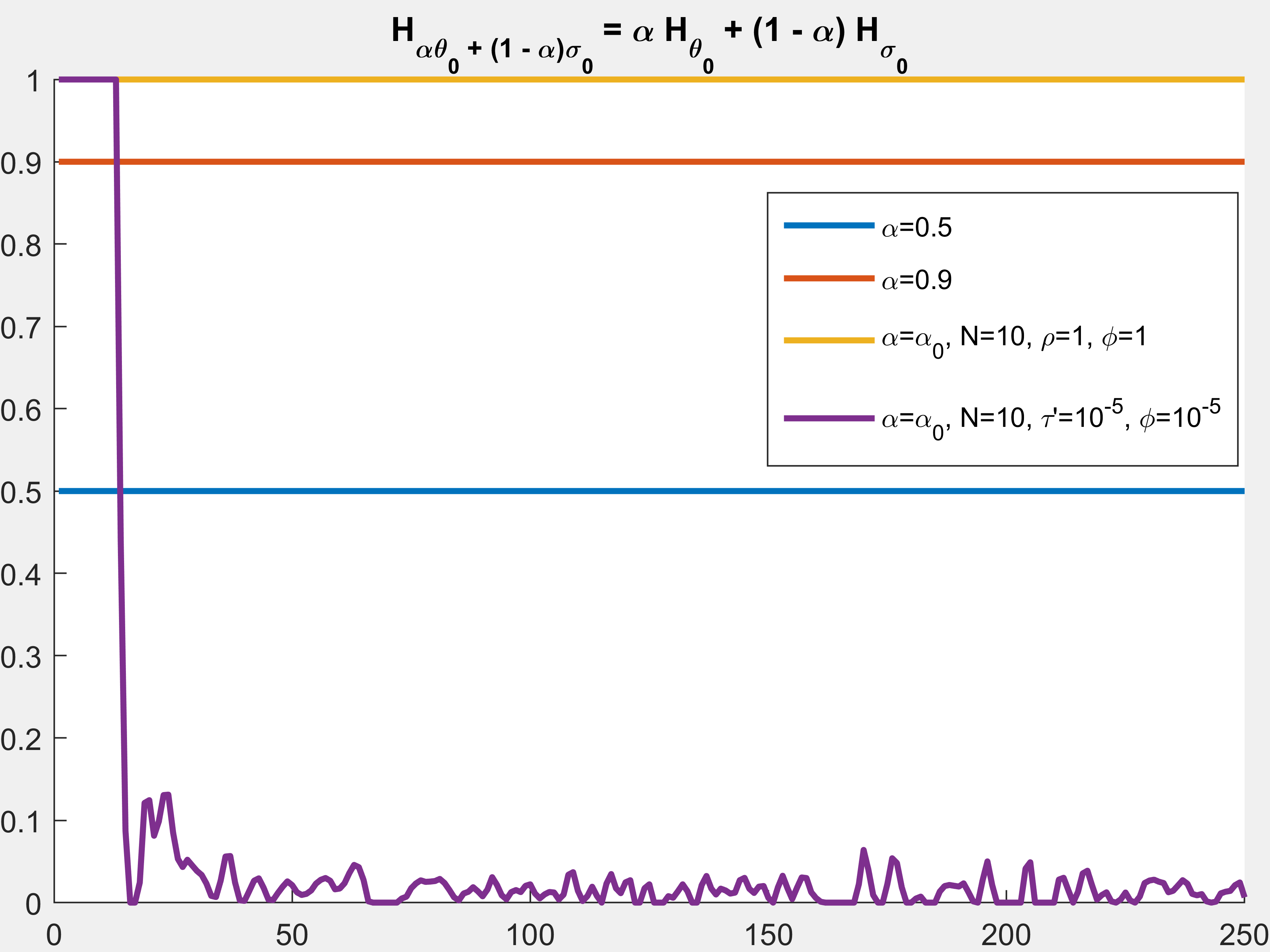}
        \caption{ADMM-PnP, $\alpha_0(k)$ vs. $k$}
        \label{Fig ADMM-PnP weights alpha}
    \end{subfigure}    
    \caption{FBS-PnP and ADMM-PnP with $H_{\alpha \theta_0} + (1-\alpha)\sigma_0$, see Example \ref{Example FBS-PnP BM3D-DnCNN} and Example \ref{Example ADMM-PnP DM3D+DnCNN}}
    \label{Figure FBS-PnP and ADMM-PnP combined denoiser} 
\end{figure}

\begin{example}\label{Example FBS-PnP BM3D-DnCNN}\textbf{[FBS-PnP using $H_{\theta_0}$ and $H_{\sigma_0}$ together]}
\mbox{ }

In this example, we combine the denoisers $H_{\theta_0}$ and $H_{\sigma_0}$ together, to produce the denoised iterates $z_k^\delta$ and examine their joined effects, i.e., we would like to take advantage of both these denoisers, the classical as well as the learned. Again, we keep the experimental setup of Example \ref{Example FBS-PnP} unchanged, but use the following weighted denoiser, for $\alpha \geq 0$ and $\beta \geq 0$,
\begin{equation}\label{BM3D+DnCNN weighted}
H_{\alpha \theta_0 + \beta \sigma_0} = \alpha H_{\theta_0} + \beta H_{\sigma_0}
\end{equation}
in the FBS-PnP algorithm. First, we try with assigning equal weights ($\alpha = \beta=\frac{1}{2}$), i.e., $H_{(\theta_0 + \sigma_0)/2} = (H_{\theta_0} + H_{\sigma_0})/2$, and the results are shown in Table \ref{Table FBS-PnP combined denoisers} and Figure \ref{Figure FBS-PnP and ADMM-PnP combined denoiser}. Although it's better than using only $H_{\theta_0}$, but no where comparable to the result obtained using $H_{\sigma_0}$, since the strong denoiser $H_{\theta_0}$ is dominating. Of course, now one can subdue the denoiser $H_{(\theta_0 + \sigma_0)/2}$, as done for $H_{\theta_0}$ in \eqref{Strong deno. alpha0}, but then, we won't be making much use of $H_{\sigma_0}$. The proper usage of both these denoisers is through weighing them differently in \eqref{BM3D+DnCNN weighted}. We instead use a simpler (normalized) version of the expression shown in \eqref{BM3D+DnCNN weighted}, by having $\beta = 1 - \alpha$, for $0\leq \alpha \leq 1$, i.e., 
\begin{equation}\label{BM3D+DnCNN alpha}
H_{\alpha H_{\theta_0} + (1-\alpha) H_{\sigma_0}} = \alpha H_{\theta_0} + (1-\alpha) H_{\sigma_0}.
\end{equation}
The recoveries corresponding to few $\alpha$ values are presented in Table \ref{Table FBS-PnP combined denoisers} and Figure \ref{Figure FBS-PnP and ADMM-PnP combined denoiser}. In addition, note that, the simpler expression in \eqref{BM3D+DnCNN alpha} further helps us to automate the selection of the $\alpha$-values, via a method suggested in \cite{nayak2021PnP}, that is, at every step $k$, choose the value of $\alpha$ such that $z_k^\delta(\alpha) = H_{\alpha H_{\theta_0} + (1-\alpha) H_{\sigma_0}}(x_k^\delta)$ best satisfies the selection criterion $\mcal{S}$, i.e.,
\begin{align}\label{Strong deno. alpha3}
\alpha_0(k) := &\argmin_{\alpha \in [0,1]} \;\;\; \mcal{S}(z_k^\delta(\alpha))\\
&\mbox{such that,} \;\; z_k^\delta(\alpha) := H_{\alpha H_{\theta_0} + (1-\alpha) H_{\sigma_0}}(x_k^\delta) = \alpha H_{\theta_0}(x_k^\delta) + (1-\alpha) H_{\sigma_0}(x_k^\delta). \notag
\end{align}
Note that, the minimization problem \eqref{Strong deno. alpha3} may not be strictly convex, i.e., there might not be a global minimizer $\alpha_0(k)$. Nevertheless, this is simply a sub-problem intended to find an appropriate $\alpha$ value between 0 and 1, depending on the selection criterion $\mcal{S}$, and hence, even if the best $\alpha_0$ is not obtained, any $\alpha \in [0, 1)$ will reduce the denoising strength, and empirical results show that \eqref{Strong deno. alpha3} works fine, see Figure \ref{Figure FBS-PnP and ADMM-PnP combined denoiser} and Table \ref{Table FBS-PnP combined denoisers}. 
Moreover, one doesn't have to compute $H_{\theta_0}(x_k^\delta)$ and $H_{\sigma_0}(x_k^\delta)$ repeatedly for different values of $\alpha$, when finding $\alpha_0$ in \eqref{Strong deno. alpha3}, as it can be expensive; one simply has to compute them once, for each iterations iteration, and use the results $\left(H_{\theta_0}(x_k^\delta), H_{\sigma_0}(x_k^\delta)\right)$ to find $\alpha_0(k)$. Note that, from Figure \ref{Fig FBS-PnP weights alpha}, the values of $\alpha_0(k)$ are high initially but then it's almost zero in the later iterations, indicating that $H_{\theta_0}$ is active in the initial iterations but then its contribution is pushed to zero (to avoid any instabilities/hallucinations arising from it), and the contribution of $H_{\sigma_0}$ starts dominating. This even leads to a recovery better than using only $H_{\sigma_0}$. Also, from the $DC(k,\alpha)$ ratio graph in Figure \ref{Figure denoising-to-consistency FBS-PnP + ADMM-PnP}, one can observe that, for equal weights ($\alpha = \beta$) the ratio $DC(k)$ keeps on increasing (as $H_{\theta_0}$ is dominant), where as, the graph does not blow up for $\alpha_0(k)$, generated from \eqref{Strong deno. alpha3}.
\end{example}

\begin{example}\label{Example ADMM-PnP DM3D+DnCNN}
In this example, keeping the settings of Example \ref{Example FBS-PnP} unchanged, we perform the ADMM-PnP algorithm using the weighted combination of both the denoisers $H_{\theta_0}$ and $H_{\sigma}$, i.e., $H_{\alpha H_{\theta_0} + (1-\alpha) H_{\sigma_0}}$, for $\alpha \in [0,1]$, as defined in \eqref{BM3D+DnCNN alpha}. Note that, here the optimization architecture ($\mcal{OA}$), for solving the data-consistency step \eqref{ADMM-PnP data-consistency 2}, and the value of $\phi$ also effect the recovery process. Thus, one can observe that, for $\mcal{OA}$ and $\phi$ values that promote larger data-consistency steps, such as CGLS$(N,\phi,\rho,y_0^k)$ for large $N$, $\phi$ values and small $\rho$ value (i.e., the noise levels in $x_k^\delta$ increase rapidly), if we choose $\alpha$ based on \eqref{Strong deno. alpha3}, then the weighted denoiser $H_{\alpha_0 H_{\theta_0} + (1-\alpha_0) H_{\sigma_0}}$ behaves similar to $H_{\theta_0}$ (i.e., $\alpha_0 \approx 1$), since $H_{\theta_0}$ can denoise $x_k^\delta$ more appropriately and $H_{\sigma_0}$ is inefficient, in this case. In contrast, for $\mcal{OA}$ and $\phi$ values that promote smaller data-consistency steps, such as GD$(N,\phi,\tau',y_0^k)$ for small $N$, $\tau'$ and $\phi$ values (i.e., the noise levels in $x_k^\delta$ increase steadily), if we choose $\alpha$ based on \eqref{Strong deno. alpha3}, then the stronger denoiser $H_{\theta_0}$ is dominant for the initial few iterations, when the noise level is high, but for later iterations, the contribution of $H_{\sigma_0}$ dominates $H_{\theta_0}$ (i.e., $\alpha_0(k) \approx 1$, for initial few $k$, and $\alpha_0(k) \approx 0$, for $k >> 1$), to avoid the instabilities/hallucinations created from $H_{\theta_0}$, in this case. This is also true for any fixed value of $\alpha$, i.e., if $\mcal{OA}$ promotes fast increment in the noise levels of $x_k^\delta$, then larger value of $\alpha$ (i.e., $H_{\theta_0}$ dominating $H_{\sigma_0}$) provides better result than smaller $\alpha$-values, where as, if $\mcal{OA}$ promotes slow increment in the noise levels of $x_k^\delta$, then smaller value of $\alpha$ (i.e., $H_{\sigma_0}$ dominating $H_{\theta_0}$) provides better result than smaller $\alpha$-values. These phenomena are presented in Table \ref{Table ADMM-PnP combined denoisers} and Figure \ref{Figure FBS-PnP and ADMM-PnP combined denoiser}.

\end{example}

\section{Conclusion and Future Research}
In this paper, we tried to present the instabilities/hallucinations arising in a PnP-algorithm when using a learned denoiser, which can be quite different from the artifacts/instabilities inherent to an inverse problem. We then provide some techniques to subdue these instabilities, produce stable reconstructions and improve the recoveries significantly. We also compare the behavior/dynamics of the FBS-PnP algorithm vs. the ADMM-PnP algorithm, and which method produce better results, depending on a given scenario. In fact, we showed that the ADMM-PnP algorithm is heavily dependent on the optimization architecture  ($\mcal{OA}$), involved in the data-consistency step, and the recoveries can greatly improve/degrade depending on $\mcal{OA}$ and the $\phi$ values. In addition, we also present a method to combine the classical denoiser and the learned denoiser, in a weighted manner, to produce results, which are much better than the individual reconstructions, i.e., one can take advantage of both these worlds.

In a future work, we would like to extend this idea to apply on image reconstruction methods based on deep-learning, i.e., instead of using a learned denoiser in the PnP-algorithm, the image reconstruction methods that involve deep-learning architecture directly in the reconstruction process, such as an unrolled neural network scheme for image reconstruction. We believe that, by incorporating ideas similar to what is developed in this paper, one can also subdue the instabilities arising in those methods, as is shown in \cite{Antun_Renna_Poon_Adcock_Hansen}.

\begin{table}[h!]
    \centering
    \begin{tabular}{|p{1.7cm}||p{1.1cm}|p{1.1cm}|p{1cm}|p{1cm}|p{1cm}|p{1cm}|p{1.9cm}|}
    \hline
    \multicolumn{8}{|c|}{Comparing denoisers $H_{\theta_0, \alpha}$ (Attenuated DnCNN) vs. $H_{\sigma_0}$ (BM3D)}\\
    \hline
    $H_{\theta_0,\alpha}$ & k($\delta,\mcal{S}_0$) & MSE & $\mcal{D}$-err.  & $\mcal{S}$-err. & PSNR & SSIM & Min.MSE \\
    \hline
	 $\alpha = 1$ & 47 & 0.3848 & 0.0667  & 0.0745 & 20.46 & 0.4493  & 0.3608 (40)\\ 
    \hline 
	 $\alpha = 0.1$ & 81 & 0.1796 & 0.0219  & 0.0236 & 27.08 & 0.8004  & 0.1763 (74)\\ 
    \hline          
	 $\alpha = 0.01$ & 140 & 0.1241 & 0.0096  & 0.0150 & 30.29 & 0.7899  & 0.1231 (154)\\ 
    \hline           
	 $\alpha = 0.001$ & 250 & 0.1151 & 0.0058  & 0.0153 & 30.95 & 0.6901  & 0.1151 (250)\\ 
    \hline        
	 $\alpha = 0.0001$ & 250 & 0.1800 & 0.0048  & 0.0202 & 27.06 & 0.4579  & 0.1800 (250)\\ 
    \hline     
    \hline
	 $H_{\sigma_0}$ & 250 & \textbf{0.0450} & 0.0101  & \textbf{0.0099} & \textbf{39.09} & \textbf{0.9340}  & 0.0450 (250)\\ 
    \hline     
    \end{tabular}
    \caption{FBS-PnP: learned vs. classical denoiser, Example \ref{Example FBS-PnP}.}
    \label{Table FBS-PnP}
\end{table}

\begin{table}[h!]
    \centering
    \begin{tabular}{|p{2cm}||p{1cm}|p{1cm}|p{1cm}|p{1cm}|p{1cm}|p{1cm}|p{1.9cm}|}
    \hline
    \multicolumn{8}{|c|}{Comparing denoisers $H_{\theta_0, \alpha}$ vs. $H_{\sigma_0}$ (BM3D), for various $\mcal{OA}(N,\phi,\rho, \tau',y_0^k)$}\\
    \hline
     CGLS $(N,\phi,\rho,y_0^k)$ & N = 100 & $\phi=1$ & $\rho=1$ & $y_0^k = x_k^\delta$ & & &\\
    \hline    
    Denoiser & $k(\delta,\mcal{S}_0)$ & MSE & $\mcal{D}$-err.  & $\mcal{S}$-err. & PSNR & SSIM & Min.MSE \\
    \hline    
    $H_{\theta_0}$ & 108 & 0.1200 & 0.0045 & 0.0171 & 30.59 & 0.5878 & 0.1168 (15) \\
    \hline
    $H_{\sigma_0}$ & 3 & 0.2184 & 0.0045 & 0.0243 & 25.38 & 0.3805 & 0.2177 (2) \\
    \hline    
    \hline
    Fixing $H_{\theta_0}$, & fix & $\phi$ & $y_0^k$ & but & $\rho$, & N & changing \\
    \hline    
    N=10/$\rho$=1 & 182 & 0.1203 & 0.0046 & 0.0171 & 30.57 & 0.5861 & 0.1201 (239)\\
    \hline      
    N=10/$\rho$=100 & 240 & 0.0934 & 0.0083 & 0.0129 & 32.77 & 0.8803 & 0.0933 (49)\\
    \hline     
    N=100/$\rho$=100 & 217 & 0.0934 & 0.0083 & 0.0129 & 32.77 & 0.8803 & 0.0933 (17)\\
    \hline       
    \hline
    Fixing $H_{\sigma_0}$, & fix & $\phi$ & $y_0^k$ & but & $\rho$, & N & changing \\
    \hline    
    N=10/$\rho$=1 & 20 & 0.2197 & 0.0044 & 0.0243 & 25.33 & 0.3759 & 0.2141 (11)\\
    \hline      
    N=10/$\rho$=100 & 31 & 0.2201 & 0.0044 & 0.0243 & 25.32 & 0.3744 & 0.2139 (14)\\
    \hline         
    N=100/$\rho$=100 & 15 & 0.2209 & 0.0044 & 0.0243 & 25.29 & 0.3721 & 0.2138 (6)\\
    \hline       
    \hline
     GD $(N,\phi,\tau',y_0^k)$ & fix & $H_{\theta_0}$ & N= 10 & $\tau'= 10^{-5}$ & $y_0^k = y_k^\delta$ & but & changing $\phi$\\
    \hline     
    $\phi= 10^{-1}$ & 22 & 0.1723 & 0.0197 & 0.0237 & 27.44 & 0.8026 & 0.1719 (21)\\
    \hline     
    $\phi= 10^{-3}$ & 21 & 0.1960 & 0.0252 & 0.0277 & 26.32 & 0.7596 & 0.1926 (19)\\
    \hline    
    $\phi= 10^{-5}$ & 21 & 0.1964 & 0.0254 & 0.0278 & 26.31 & 0.7583 & 0.1929 (19)\\
    \hline      
    \hline
     GD $(N,\phi,\tau',y_0^k)$ & fix & $H_{\sigma_0}$ & N= 10 & $\tau'= 10^{-5}$ & $y_0^k = y_k^\delta$ & but & changing $\phi$\\
    \hline     
    $\phi= 10^{-1}$ & 84 & 0.2219 & 0.0045 & 0.0248 & 25.24 & 0.3674 & 0.2166 (52)\\
    \hline     
    $\phi= 10^{-3}$ & 250 & 0.1497 & 0.0050 & 0.0162 & 28.67 & 0.5857 & 0.1473 (209)\\
    \hline    
    $\phi= 10^{-5}$ & 250 & 0.1405 & 0.0051 & 0.0158 & 29.21 & 0.6008 & 0.1400 (229)\\
    \hline   
    \hline
     CGLS $(N,\phi,\rho,y_0^k)$ & fix & $H_{\sigma_0}$ & N= 10 & $\rho = 10^{5}$ & $y_0^k = y_k^\delta$ & but & changing $\phi$\\
    \hline     
    $\phi= 10^{-3}$ & 250 & 0.2850 & 0.0384 & 0.0525 & 23.07 & 0.7222 & 0.2850 (250)\\
    \hline    
    $\phi= 10^{-5}$ & 250 & 0.2850 & 0.0384 & 0.0525 & 23.07 & 0.7222 & 0.2850 (250)\\
    \hline        
    \hline
     GD $(N,\phi,\tau',y_0^k)$ & fix & $H_{\sigma_0}$ & $\phi = 10^{-5}$ & $\tau'= 10^{-5}$ & $y_0^k = y_k^\delta$ & but & changing N\\
    \hline     
    N = 100 & 25 & 0.2161  & 0.0045 & 0.0238  & 25.47 & 0.3846 & 0.2132 (17) \\
    \hline     
    N = 10 & 250 & 0.1405 & 0.0051 & 0.0158 & 29.21 & 0.6008 & 0.1400 (229)\\
    \hline    
    N = 3 & 250 & 0.0509 & 0.0093 & 0.0102 & 38.03 & 0.8992 & 0.0509 (250) \\
    \hline             
    N = 1 & 250 & \textbf{0.0395} & 0.0102 & \textbf{0.0101} & \textbf{40.24} & \textbf{0.9640} & 0.0395 (250) \\
    \hline                 
    \end{tabular}
    \caption{ADMM-PnP: learned vs. classical denoiser, Example \ref{Example ADMM-PnP}.}
    \label{Table ADMM-PnP}
\end{table}

\begin{table}[h!]
    \centering
    \begin{tabular}{|p{1.7cm}||p{1.1cm}|p{1.1cm}|p{1cm}|p{1cm}|p{1cm}|p{1cm}|p{1.9cm}|}
    \hline
    \multicolumn{8}{|c|}{FBS-PnP: $x_{k(\delta,\mcal{S}_0)}$ for $H_{\alpha \theta_0 + (1-\alpha)H_{\sigma_0}} = \alpha H_{\theta_0} + (1-\alpha)H_{\sigma_0}$}\\
    \hline
    $\alpha$ & k($\delta,\mcal{S}_0$) & MSE & $\mcal{D}$-err.  & $\mcal{S}$-err. & PSNR & SSIM & Min.MSE \\
    \hline
	 $\alpha$=1 ($H_{\theta_0}$) & 47 & 0.3848 & 0.0667  & 0.0745 & 20.46 & 0.4493  & 0.3608 (40)\\ 
    \hline     
	 $\alpha = 0.5$ & 53 & 0.2618 & 0.0432  & 0.0476 & 23.81 & 0.6131  & 0.2609 (51)\\ 
    \hline 
	 $\alpha = 0.1$ & 84 & 0.1757 & 0.0229  & 0.0218 & 27.27 & 0.8605  & 0.1709 (76)\\ 
    \hline          
	 $\alpha = 0.01$ & 141 & 0.0954 & 0.0123  & 0.0127 & 32.58 & 0.9531  & 0.0953 (139)\\ 
    \hline         
	 $\alpha$=0 ($H_{\sigma_0}$) & 250 & {0.0450} & 0.0101  & {0.0099} & {39.09} & {0.9340}  & 0.0450 (250)\\
    \hline       
	 $\alpha = \alpha_0$ & 239 & {0.0398} & 0.0102  & \textbf{0.0099} & \textbf{40.16} & 40.16  & \textbf{0.0391}(250)\\ 
    \hline        
    \end{tabular}
    \caption{Combining classical and learned denoiser, see Example \ref{Example FBS-PnP BM3D-DnCNN}.}
    \label{Table FBS-PnP combined denoisers}
\end{table}

\begin{table}[h!]
    \centering
    \begin{tabular}{|p{2cm}||p{1cm}|p{1cm}|p{1cm}|p{1cm}|p{1cm}|p{1cm}|p{1.9cm}|}
    \hline
    \multicolumn{8}{|c|}{ADMM-PnP: $x_{k(\delta,\mcal{S}_0)}$ for $H_{\alpha \theta_0 + (1-\alpha)H_{\sigma_0}} = \alpha H_{\theta_0} + (1-\alpha)H_{\sigma_0}$}\\
    \hline
     CGLS $(N,\phi,\rho,y_0^k)$ & N = 10 & $\phi=1$ & $\rho=1$ & $y_0^k = x_k^\delta$ & & &\\
    \hline    
    $\alpha$ & $k(\delta,\mcal{S}_0)$ & MSE & $\mcal{D}$-err.  & $\mcal{S}$-err. & PSNR & SSIM & Min.MSE \\
    \hline    
    $\alpha=1$ ($H_{\theta_0}$) & 182 & 0.1203 & 0.0046 & 0.0171 & 30.57 & 0.5861 & 0.1201 (239)\\
    \hline      
    $\alpha=0$ ($H_{\sigma_0}$) & 20 & 0.2197 & 0.0044 & 0.0243 & 25.33 & 0.3759 & 0.2141 (11)\\
    \hline      
    $\alpha=0.5$ & 25 & 0.2145 & 0.0043 & 0.0234 & 25.54 & 0.3774 & 0.2112 (14) \\
    \hline    
    $\alpha=0.9$ & 92 & 0.1312 & 0.0043 & 0.0187 & 29.38 & 0.5257 & 0.1312 (250) \\
    \hline        
    $\alpha=\alpha_0$ & 179 & 0.1202 & 0.0049 & 0.0171 & 30.57 & 0.5862 & 0.1201 (246) \\
    \hline            
    \hline
     GD $(N,\phi,\tau',y_0^k)$ & N= 10 & $\phi = 10^{-5}$ & $\tau'= 10^{-5}$ & $y_0^k = y_k^\delta$ & & &\\
    \hline     
     $\alpha=1$ ($H_{\theta_0}$) & 21 & 0.1964 & 0.0254 & 0.0278 & 26.31 & 0.7583 & 0.1929 (19)\\
    \hline      
    $\alpha=0$ ($H_{\sigma_0}$) & 250 & 0.1405 & 0.0051 & 0.0158 & 29.21 & 0.6008 & 0.1400 (229)\\
    \hline   
    $\alpha=0.5$ & 24 & 0.1697 & 0.0197 & 0.0221 & 27.58 & 0.8249 & 0.1693 (23) \\
    \hline    
    $\alpha=0.9$ & 93 & 0.1307 & 0.0177 & 0.0154 & 29.84 & 0.8688 & 0.1284 (39) \\
    \hline        
    $\alpha=\alpha_0$ & 217 & {0.0949} & 0.0087 & \textbf{0.0127} & \textbf{32.62} & \textbf{0.8998} & \textbf{0.0946}(212) \\
    \hline                 
    \end{tabular}
    \caption{Combining classical and learned denoiser, see Example \ref{Example ADMM-PnP DM3D+DnCNN}.}
    \label{Table ADMM-PnP combined denoisers}
\end{table}

\bibliography{thesisref} 
\bibliographystyle{ieeetr}

\end{document}